%% file: paper.tex
\documentclass[conference]{IEEEtran}
\IEEEoverridecommandlockouts
\usepackage{cite}
\usepackage{amsmath,amssymb,amsfonts}
\usepackage{algorithmic}
\usepackage{graphicx}
\usepackage{textcomp}
\usepackage{xcolor}
\def\BibTeX{{\rm B\kern-.05em{\sc i\kern-.025em b}\kern-.08em
    T\kern-.1667em\lower.7ex\hbox{E}\kern-.125emX}}

\input{definitions}

\input{acronyms}

\begin{document}

\title{An Empirical Analysis of \\Measure-Valued Derivatives for Policy Gradients}

\author{\IEEEauthorblockN{Jo\~{a}o Carvalho$^{1}$, Davide Tateo$^{1}$, Fabio Muratore$^{1,2}$, Jan Peters$^{1,3}$}
\IEEEauthorblockA{\textit{Intelligent Autonomous Systems} \\
\textit{Technische Universit\"at Darmstadt}, Darmstadt, Germany \\
\{joao,davide,fabio\}@robot-learning.de, jan.peters@tu-darmstadt.de}

\thanks{$^{1}$~Technische Universit\"at Darmstadt, Darmstadt, Germany}
\thanks{$^{2}$~Honda Research Institute Europe, Offenbach am Main, Germany}
\thanks{$^{3}$~Max Planck Institute for Intelligent Systems, T\"{u}bingen, Germany}
\thanks{This project has received funding from the German Federal Ministry of Education and Research (BMBF) project 16SV798  (KoBo34), and the European Union’s Horizon 2020 research and innovation programme under grant agreement No. \#640554 (SKILLS4ROBOTS) and No. \#713010 (GOAL-Robots). Fabio Muratore gratefully acknowledges the financial support from Honda Research Institute Europe.}
}

{\maketitle}	

\begin{abstract}
Reinforcement learning methods for robotics are increasingly successful due to the constant development of better policy gradient techniques.
A \textit{precise} (low variance) and \textit{accurate} (low bias) gradient estimator is crucial to face increasingly complex tasks.
Traditional policy gradient algorithms use the likelihood-ratio trick, which is known to produce unbiased but high variance estimates.
More modern approaches exploit the reparametrization trick, which gives lower variance gradient estimates but requires differentiable value function approximators.
In this work, we study  a different type of stochastic gradient estimator: the Measure-Valued Derivative. This estimator is unbiased, has low variance, and can be used with differentiable and non-differentiable function approximators.
We empirically evaluate this estimator in the actor-critic policy gradient setting and show that it can reach comparable performance with methods based on the likelihood-ratio or reparametrization tricks, both in low and high-dimensional action spaces.
\end{abstract}

\section{Introduction}

Complex robotics tasks, such as locomotion and manipulation, can be formulated as \gls{rl} problems in continuous state and action spaces~\cite{kohl2004locomotion}.
The \gls{rl} objective is commonly an expectation dependent on a parameterized control policy, whose optimization is done by computing a gradient \wrt~the policy parameters to steadily move them in the gradient ascent direction~\cite{deisenroth2013} -- known as the policy gradient.
If the control policy is stochastic, this method can be viewed as computing the gradient of an expectation \wrt~a distribution containing the policy parameters, both in the actor-only version~\cite{williamsReinforce1992}, as well as in the actor-critic one~\cite{sutton1999PG}.

If the gradient of the expectation of a function \wrt~the distribution parameters cannot be found analytically, there are three main approaches to obtain an unbiased estimate: the likelihood-ratio, also called \gls{sf}; the Pathwise Derivative, commonly known as the \gls{reptrick}; and the \gls{mvd}~\cite{mohamed2019monte}.

In policy gradients, the gradient estimation problem has been solved extensively using the \gls{sf}, which is known to produce high-variance estimates, particularly if used in trajectory-based formulations, since its variance grows linearly with the number of steps in the trajectory~\cite{bhatt2019pgweak}.
Thus, many methods require a baseline for variance reduction~\cite{baxter2001GPOMDP}, and the optimal one needs to be computed for every single case~\cite{Peters:2006:PGRobotics}.
Frequently used baseline heuristics are the mean of the returns or the state value function.
REINFORCE is an example of a \gls{sf} based method~\cite{williamsReinforce1992}.

Generally, the \gls{reptrick}~\cite{kingma2014autoencoding} is a low variance gradient estimator that can be used in distributions of continuous and discrete random variables. For the latter, the Gumbel-softmax trick allows the backpropagation of gradients~\cite{JanGuPoo17}. Moreover, it requires the function we are optimizing to be differentiable, which excludes non-differentiable approximators, such as regression trees~\cite{Geurts2006ExtraTrees}. 
Although the \gls{reptrick} often produces low variance estimates, this property does not hold for all function classes, in particular those whose derivative has a high Lipschitz constant~\cite{glasserman2004monte}. An example of an algorithm using this estimator is the \gls{sac}~\cite{Haarnoja2018SAC}.

\gls{mvd} is another method to compute unbiased gradient estimates, which has close connections to \gls{fd}, and it is not used often in the \gls{ml} community.
Similarly to the \gls{sf} it applies to both discrete and continuous distributions, but it generally has low variance and avoids computing an extra baseline for variance reduction.
Contrary to the \gls{reptrick}, it is not restricted to differentiable functions, which makes it applicable to more function classes.
Few works have studied its application to \gls{rl}~\cite{krishnamurthy2011realtime, bhatt2019pgweak}. 

In this paper, we analyze the properties of \gls{mvd} in actor-critic policy gradient algorithms for on and off-policy \gls{rl}.
We compare the estimation errors produced by the three different estimators in \gls{lqr} problems, for which we know the true value function and policy gradient.
We show that replacing the \gls{reptrick} with \gls{mvd} in an off-policy deep \gls{rl} algorithm leads to similar performance results, even in complex environments with high-dimensional state and action spaces.
To argue that \glspl{mvd} can be generally used with broader classes of critic approximators, where the \gls{reptrick} is not available, we construct an example of an on-policy procedure that uses a $Q$-function fitted with Extra-Trees~\cite{Geurts2006ExtraTrees} -- a non-differentiable model.
The results suggest that this type of estimator can be used to compute policy gradients in continuous state and action spaces.

Supplementary material with experiment details and a link to a repository to reproduce the results from this paper is available at \href{https://www.ias.informatik.tu-darmstadt.de/uploads/Team/JoaoCarvalho/mvd\_rl-supp.pdf}{https://www.ias.informatik.tu-darmstadt.de/uploads/Team/JoaoCarvalho/mvd\_rl-supp.pdf}.

\section{Related Work}
\label{sec:relatedwork}
Policy gradient methods can be derived through the trajectory, also called actor-only, or actor-critic formulations~\cite{deisenroth2013}.
In the former, the expected return from a state-action pair is estimated with samples from the trajectory rollout, as done in REINFORCE~\cite{williamsReinforce1992} and GPOMDP~\cite{baxter2001GPOMDP}.
In stochastic environments this results in high variance gradient estimates and poor convergence.
Instead, an actor-critic agent performs updates by estimating the returns using a state-action value function approximator.
These methods have lower variance and show more stable learning.
Several actor-critic algorithms have been presented in the recent years, particularly using deep neural networks as function approximators, including \gls{trpo}~\cite{Schulman2015TRPO}, \gls{ddpg}~\cite{Lillicrap2016DDPG}, \gls{td3}~\cite{fujimoto2018td3} and \gls{sac}~\cite{Haarnoja2018SAC}. The connection between trajectory and actor-critic methods was established in the policy gradient theorem in its stochastic~\cite{sutton1999PG} and deterministic versions~\cite{Silver2014DPG}.

Most policy gradient methods optimize stochastic policies using the \gls{sf} or \gls{reptrick}, while \glspl{mvd} have not yet been extensively explored in the \gls{ml} and \gls{rl} communities. A modified version of the policy gradient theorem using \glspl{mvd}, also referred to as weak derivatives, was first introduced in~\cite{bhatt2019pgweak}.
The authors derive an unbiased estimator and provide an extensive theoretical analysis of its properties, including the variance and computational complexity. 
However, the practical implementation of the algorithm has to perform two rollouts starting from the same state to obtain an unbiased estimate of the state-action value function.
Applying this approach to a real-world scenario, where we cannot reset a system to an arbitrary continuous state, is impossible.
Furthermore, it makes use of Monte Carlo rollouts to estimate the return from a state-action pair, discarding all the collected data after an update, which leads to inefficient use of samples. 
Other works used \glspl{mvd} to solve Constrained Markov Decision Processes~\cite{krishnamurthy2003implementation, krishnamurthy2011realtime}, also assuming that the environment can be reset to any state.
To the best of our knowledge, there is no actor-critic algorithm based on \glspl{mvd} that replicates a \gls{rl} scenario.

\section{Background}
\label{sec:background}

\subsection{Reinforcement Learning and Policy Gradients}
\label{section:reinforcementlearning}

Let a \gls{mdp} be defined as a tuple $\mathcal{M} = (\statespace, \actionspace, \mathcal{R}, \mathcal{P}, \gamma, \mu_0)$, where $\statespace$ is a continuous state space $\state \in \statespace$, $\actionspace$ is a continuous action space $\action \in \actionspace$, $\mathcal{P}: \statespace \times \actionspace \times \statespace \to \R$ is a transition probability function, with $\mathcal{P}(\nextstate | \state, \action)$ the density of landing in state $\nextstate$ when taking action $\action$ in state $\state$, $\mathcal{R}: \statespace \times \actionspace \to \R$ is a reward function, $\gamma \in [0, 1) $ is a discount factor, and $\mu_0:\statespace \to \R$ the initial state distribution.
A policy $\policy$ is a mapping from states to actions. If deterministic, it defines what action to take in each state $\action = \pi(\state)$, while a stochastic one assigns a probability distribution over possible actions $\action \sim \policy(\cdot | \state)$.
The discounted state visitation under a policy $\policy$ is defined as $d^\policy_\gamma(\state) = \sum_{t=0}^{\infty} \gamma^t P(\state_t = \state | \state_0, \policy)$, where $\state_0$ is the initial state, and $P: \statespace \to \R$ the probability of being in state $\state$ at time step $t$ given the initial state and the policy.
The discounted state distribution is then given by $ \mu^\policy_\gamma(\state) = (1-\gamma) d^\policy_\gamma(\state)$.
The state-action value function -- $Q$-function -- is the discounted sum of rewards collected from a given state-action pair following the policy $\policy$, $Q^{\policy}(\state, \action) = \E{\policy,\Ptrans }{\sum_{t=0}^\infty \gamma^t r(\state_t, \action_t) | \state_0 = \state, \action_0 = \action}$, and the state value function is its expectation \wrt~the policy $V^{\policy}(\state) = \E{\action \sim \policy}{Q^{\policy}(\state, \action)}$. The advantage function is given as the difference between the two $A^{\policy}(\state, \action) = Q^{\pi}(\state, \action) - V^{\policy}(\state)$.
In general, the goal of a \gls{rl} agent is to maximize the expected sum of discounted rewards from any initial state
\begin{align}
	J(\policy) = \E{\state_0 \sim \mu_0}{V^{\policy}(\state_0)} = \E{\tau \sim \mu_0, \policy, \mathcal{P}}{\sum_{t=0}^{\infty} \gamma^t r_t},
	\label{eq:rlobjective}
\end{align} 
where $\tau$ is a state-action trajectory $(\state_0, \action_0, \state_1, \action_1, \ldots)$, determined by the policy and the environment dynamics.

In high-dimensional and continuous action spaces, instead of constructing value functions and retrieving optimal actions afterwards, a policy with parameters $\policyparams$ is updated iteratively with a gradient ascent step on \eqref{eq:rlobjective}, $\policyparams \leftarrow \policyparams + \alpha \gradpolicyparams J(\policyparametrized)$. The on-policy policy gradient theorem~\cite{sutton1999PG} establishes this gradient for stochastic policies as
\begin{align}
	\gradpolicyparams J(\policyparametrized) = \E{\state \sim \mu_{\gamma}^{\policy}}{\int \gradpolicyparams \policy(\action | \state; \policyparams) Q^\policy(\state, \action) \de \action},
	\label{eq:on-policy-grad-theorem}
\end{align}
which can be extended to the off-policy~\cite{Degris2012OffPAC} and deterministic policy~\cite{Silver2014DPG} settings. Note that the integral in \eqref{eq:on-policy-grad-theorem} is not an expectation. 

\subsection{Monte Carlo Gradient Estimators}
\label{section:mc-gradient-estimation}

The optimization goal of several problems in \gls{ml}, e.g. \gls{vi}, is often posed as an expectation of a function $f$ \wrt~a distribution $p$ parameterized by $\distributionparams$
\begin{align}
	J(\distributionparams) = \E{p(\xvec; \distributionparams)}{f(\xvec)} = \int p(\xvec; \distributionparams) f(\xvec) \de \xvec,
	\label{eq:stochasticobj}
\end{align}
where $f: \R^n \to \R$ is an arbitrary function of $\xvec\in\mathbb{R}^n$, $p:\R^n \times \R^m \to \R$ is the distribution of $\xvec$, and $\distributionparams \in \R^m$ are the parameters encoding the distribution, also known as distributional parameters. For instance, in amortized \gls{vi}~\cite{kingma2014autoencoding}, $f$ is the \gls{elbo} and $p$ the approximate posterior distribution. If $p(\xvec; \distributionparams)$ is a multivariate Gaussian distribution over $\vec{x}$, then $\distributionparams$ aggregates the mean and covariance. 

A gradient ascent algorithm is a common choice to find the parameters $\distributionparams$ that maximize \eqref{eq:stochasticobj}. For that we need to compute the gradient
\begin{equation}
    \graddistributionparams J(\distributionparams) = \int \graddistributionparams p(\xvec; \distributionparams) f(\xvec) \de \xvec,
	\label{eq:gradstochasticobj}
\end{equation}
where for clarity of explanation we assume interchangeability of differentiation and integration (see~\cite{mohamed2019monte} for cases where it does not apply).
Since the derivative of a distribution is in general not a distribution itself, the integral in \eqref{eq:gradstochasticobj} is not an expectation and thus not solvable directly by \gls{mc} sampling.

If $f$ also depends on $\distributionparams$, the product rule gives $\graddistributionparams J(\distributionparams) = \int \graddistributionparams p(\xvec; \distributionparams) f(\xvec; \distributionparams) \de \xvec + \int p(\xvec; \distributionparams) \graddistributionparams f(\xvec; \distributionparams) \de \xvec$. The second term in the sum is an expectation, and can thus be computed by sampling. Hence, to keep the notation simple, we assume $f$ does not depend on $\distributionparams$, as its extension is trivial.

In many problems we do not have access to the true function $f$, e.g. value functions in \gls{rl}, and need to use approximators $\hat{f}$, which introduce bias and variance. Therefore we are typically interested in finding unbiased estimators for \eqref{eq:gradstochasticobj}, to prevent the accumulation of the estimation errors already given by $\hat{f}$.

Let $\hat{\vec{g}}_i \in \R^m$ define one unbiased gradient estimate of \eqref{eq:gradstochasticobj}.
Then a \gls{mc} estimation of the gradient using $M$ samples is obtained as $
\graddistributionparams J(\distributionparams) \approx  1/M \sum_{i=1}^{M} \hat{\vec{g}}_i$. 
A straightforward variance reduction technique common to any method is to use more samples, as the variance reduces with $\bigO{M^{-1}}$. However, this increases the computational complexity.

Next, we summarize the three known ways to build an unbiased estimator for \eqref{eq:gradstochasticobj}.
See~\cite{mohamed2019monte} for a throughout analysis.
\medskip
\paragraph{\textbf{\glsfirst{reptrick}}~\cite{kingma2014autoencoding}}
A random variable $\vec{x}$ is reparametrizable if it can be obtained by constructing a deterministic path from a base random variable $\epsvec$ by applying a deterministic function $g$ with parameters $\distributionparams$, such that $\xvec = g(\epsvec; \distributionparams)$.
By the change of variables we have $\graddistributionparams J(\distributionparams) = \int  p(\epsvec) \graddistributionparams f(g(\epsvec; \distributionparams)) \de \epsvec $. A single gradient estimate is given as $\hat{\vec{g}}_i^{\textrm{Rep}} = \nabla_{\xvec} f(\xvec = g(\epsvec_i; \distributionparams)) \graddistributionparams g(\epsvec_i; \distributionparams) $, with $\epsvec_i \sim p(\epsvec)$. This estimator requires $f$ to be differentiable \wrt~$\vec{x}$, to be able to compute its derivative, and typically leads to low-variance estimates. However, it can have high variance if $f$ is rough~\cite{Schulman2015GradEstimation}.
\smallskip
\paragraph{\textbf{\glsfirst{sf}}~\cite{Glynn1987LikelihoodRatio}}
From the derivative of the logarithm identity we rewrite \eqref{eq:gradstochasticobj} as $\graddistributionparams J(\distributionparams) = \int p(\xvec; \distributionparams) \graddistributionparams \log p(\xvec; \distributionparams) f(\xvec) \de \xvec$. A single gradient estimate is given as $\hat{\vec{g}}_i^{\textrm{SF}} = \graddistributionparams \log p(\xvec_i; \distributionparams) f(\xvec_i)$, with $\xvec_i \sim p(\xvec; \distributionparams)$. Contrary to the \gls{reptrick}, it does not need the derivative of $f$, but just that we can query it. It is common for the estimates to have large variance, especially in a stochastic process, as it grows linearly with the number of steps~\cite{krishnamurthy2003implementation}. In practice, the variance is reduced by subtracting a baseline to $f$, keeping the estimator unbiased. Several policy gradient algorithms are based on this estimator~\cite{williamsReinforce1992, baxter2001GPOMDP, PetersNAC2008}.
\smallskip
\paragraph{\textbf{\glsfirst{mvd}}~\cite{Pflug89}}
Even though the derivative of a distribution is not in general a distribution, it is a difference between distributions up to a normalizing constant~\cite{Pflug89}.
The main idea behind \gls{mvd} is to write the derivative \wrt~a single distributional parameter $\distributionparamk \in \R$ as a difference between two distributions, i.e. $\graddistributionparamk p(\xvec; \distributionparams) = c_{\distributionparamk} \left( p^+_{\distributionparamk}(\xvec; \distributionparams) - p^-_{\distributionparamk}(\xvec; \distributionparams) \right)$, where $c_{\distributionparamk}$ is a normalizing constant that can depend on $\distributionparamk$, and $p^+_{\distributionparamk}$ and $p^-_{\distributionparamk}$ are two distributions referred to as the positive and negative components, respectively. 
The \gls{mvd} is described by the triplet $ \left(c_{\distributionparamk}, p^+_{\distributionparamk}(\xvec; \distributionparams), p^-_{\distributionparamk}(\xvec; \distributionparams) \right)$. 
The decomposition is not unique, and different configurations of the triplet can be obtained, which leads to estimators with different properties.
For common distributions, such as the Gaussian, Poisson, or Gamma, decompositions have already been analytically derived in~\cite{heidergott2003mvds}.
Table \ref{tab:distributions} shows a common one used for the mean and standard deviation of the Gaussian distribution.

\begin{table*}[ht]
\centering
\begin{tabular}{ l c c c c | l l l } 
\hline
\\
\multicolumn{5}{c}{\textbf{\gls{mvd} Gaussian distribution} $ \Gaussian{x ; \mu, \sigma}, \distributionparamk = \{\mu, \sigma\}$} & \multicolumn{3}{c}{\textbf{Univariate distributions}} \\
\multicolumn{1}{c}{$\frac{\partial}{\partial \distributionparamk} \mathcal{N}$} & \multicolumn{1}{c}{$c_{\distributionparamk}$} & \multicolumn{1}{c}{$p^+_{\distributionparamk}(x; \distributionparamk)$} & \multicolumn{1}{c}{$p^-_{\distributionparamk}(x; \distributionparamk)$} & & Gaussian &  $\Gaussian{x \in \R; \mu, \sigma}$ & $\frac{1}{\sqrt{2 \pi} \sigma} \exp{-\frac{1}{2}\left( \frac{x-\mu}{\sigma}  \right)^2}$ \\
\cline{1-4}
$\dpartialmu \mathcal{N}$  & $\frac{1}{\sigma\sqrt{2\pi}}$ & $\mathcal{W}(x - \mu; \sqrt{2}\sigma, 2)$ &  $\mathcal{W}(\mu - x ; \sqrt{2}\sigma, 2)$ &  &  Weibull & $\mathcal{W}\left(x \in \R^+; \lambda, k\right)$  & $\frac{k}{\lambda} \left( \frac{x}{\lambda} \right)^{k-1} \exp{-\left(\frac{x}{\lambda}\right)^k}$ \\
$\dpartialstd \mathcal{N}$ & $\frac{1}{\sigma}$ & $\mathcal{M}(x ; \mu, \sigma) $ & $\mathcal{N}(x ; \mu, \sigma)$ & &  Double-sided & $\mathcal{M}\left(x \in \R; \mu, \sigma\right)$ & $\frac{1}{\sqrt{2 \pi} \sigma^3} \left( x - \mu \right)^2 \exp{-\frac{1}{2}\left( \frac{x-\mu}{\sigma}  \right)^2}$  \\
& & & &  & Maxwell  & & \\
\\
\hline 
\\
\end{tabular}
\caption{\gls{mvd} decomposition for the derivative of the Gaussian distribution \wrt~to the mean and standard deviation (left) and the univariate distributions from the decomposition (right).}
\label{tab:distributions}
\end{table*}

Expanding \eqref{eq:gradstochasticobj} with the \gls{mvd} decomposition we obtain $\graddistributionparamk J(\distributionparams) = \int c_{\distributionparamk} \left(p^+_{\distributionparamk}(\xvec; \distributionparams) - p^-_{\distributionparamk}(\xvec; \distributionparams) \right) f(\xvec) \de \xvec$.
A single gradient estimate for $\distributionparamk$ is given by $\hat{g}_{k,i}^{\textrm{MVD}} = c_{\distributionparamk} \left(f(\xvec^+; \distributionparams) - f(\xvec^-; \distributionparams) \right)$, with $\xvec^+ \sim p^+_{\distributionparamk}(\xvec; \distributionparams)$ and $\xvec^- \sim p^-_{\distributionparamk}(\xvec; \distributionparams)$. I.e., the derivative is the difference between the function $f$ evaluated at different points sampled according to the decomposition, and scaled with the normalization constant.
For example, to compute the derivative of a one-dimensional Gaussian $\mathcal{N}(x; \mu, \sigma)$ \wrt~its standard deviation, we can sample the positive part from a Double-sided Maxwell and the negative one from a Gaussian. The gradient \wrt~all parameters is just the concatenation of the derivatives \wrt~all single parameters $\distributionparamk$.
Computing one gradient estimate for all parameters $\distributionparams$ is in the order of $\bigO{2|\distributionparams|}$ queries to $f$, in contrast to the \gls{sf} and \gls{reptrick} $\bigO{1}$ complexity.

Let $\hat{g}^{\textrm{MVD}}_k$ be the estimator of $ \graddistributionparamk \E{p(\vec{x};\distributionparams)}{f(\vec{x})}$, with variance given by~\cite{mohamed2019monte}
\begin{align*}
\V{p(\vec{x};\distributionparams)}{\hat{g}^{\textrm{MVD}}_k} & = \V{p^+_{\distributionparamk}(\vec{x};\distributionparams)}{f(\vec{x})} + \V{p^-_{\distributionparamk}(\vec{x};\distributionparams)}{f(\vec{x})} \\
& \; - 2 \cov{p^+_{\distributionparamk}(\vec{x};\distributionparams)p^-_{\distributionparamk}(\vec{x}';\distributionparams)} {f(\vec{x}), f(\vec{x}')},
\end{align*}
which depends on the chosen decomposition and how correlated are the function evaluations at the positive and negative samples.
A typical choice for variance reduction is to obtain a decomposition such that the positive and negative distributions are orthogonal~\cite{heidergott2000MVDfinitehorizon}, such as for the Gaussian mean in Table \ref{tab:distributions}.
Moreover, if $f(\xvec)$ and $f(\xvec ')$ are positively correlated, then the covariance term is positive and the variance is further reduced.
This last term is possibly reduced through \textit{coupling}, which works by constructing a variate from the negative component $\xvec'$ by applying a transformation to the sample from the positive component (or vice-versa), e.g. by using common random numbers, such that they share the same source of randomness.
A simple example is when computing the gradient \wrt~the mean of a Gaussian: first we sample a variate from the Weibull distribution $\mathcal{W}(\sqrt{2}, 2)$ to construct the positive component with $\mu+\sigma \mathcal{W}( \sqrt{2}, 2)$, and then reuse the same variate to also construct the negative component~\cite{rosca:2019:mvds}. Even though this does not guarantee that the function evaluations are positively correlated, it tends to work well in practice~\cite{mohamed2019monte}.

If $p(\xvec; \distributionparams)$ is a multivariate distribution with independent dimensions it factorizes as $p(\xvec; \distributionparams) = \prod_{i=1}^{n} p(x_i; \bm{\xi}_i)$, with $x_i \in \R$ and $\bm{\xi}_i$ a subset of $\distributionparams$. For instance, if $p(\xvec; \distributionparams)$ is a Gaussian with diagonal covariance, then $\bm{\xi}_i = \{\mu_i, \sigma_i\}$.
The derivative \wrt~one parameter $\distributionparamk \in \bm{\xi}_k$ is given by
\begin{align*}
	& \dpartialomegak p(\xvec; \distributionparams) = \prod_{i=1}^{k-1} p(x_i; \bm{\xi}_i) \dpartialomegak p(x_k; \bm{\xi}_k) \prod_{j=k+1}^{n} p(x_j; \bm{\xi}_j) \\
	= & c_{\distributionparamk} \left( p^+_{\distributionparamk}(x_k; \bm{\xi}_k) - p^-_{\distributionparamk}(x_k; \bm{\xi}_k) \right) \prod_{i=1, i \neq k}^{n} p(x_i; \bm{\xi}_i) \\
	= &  c_{\omega_k} \left( p^+_{\distributionparamk}(\xvec; \distributionparams) - p^-_{\distributionparamk}(\xvec; \distributionparams) \right),
\end{align*}
where $p^+_{\distributionparamk}(\xvec; \distributionparams)$ is the original multivariate distribution with the $k$-th component replaced by the positive part of the decomposition of the univariate marginal $p(x_k; \bm{\xi}_k)$ \wrt~$\omega_k$.
The negative component $p^-_{\distributionparamk}(\xvec; \distributionparams)$ is built similarly.
A practical algorithm to sample from $p^+_{\distributionparamk}(\xvec; \distributionparams)$ first samples from the initial multivariate distribution $p(\xvec; \distributionparams)$ and then replaces the $k$-th component with a sample from the marginal $p^+_{\distributionparamk}(x_k; \bm{\xi}_k)$.
E.g., for the derivative \wrt~the mean of the multivariate Gaussian with diagonal covariance, a sample from $p_{\distributionparamk}^+(\xvec; \distributionparams)$ includes sampling first from the original multivariate Gaussian and then replacing the $k$-entry with a sample from a Weibull.

\glspl{mvd} are closely related with \glsfirst{fd} methods, as both use perturbations per parameter dimension and take the difference between function evaluations to compute a gradient estimate.
To produce an estimator of \eqref{eq:gradstochasticobj} for each parameter, \gls{fd} methods construct two distributions by perturbing the distributional parameter $\distributionparamk$ by small amounts $\pm \Delta$, and then sample from the resulting modified distributions. Although the bias disappears for $\Delta \to 0$, the variance grows with $\bigO{\Delta^{-2}}$, making the estimator unreliable~\cite{Pflug1996OptimizationStochasticModels}. Instead, \glspl{mvd} can be seen as a principled way to choose the distributions to sample from.
Note that the \gls{fd} is applied to the distributional parameters in \eqref{eq:stochasticobj} and not to the deterministic function $f$, and thus different from computing gradients to optimize $f$ only.

\begin{figure}[b] 
	\vspace{-17pt}
	\centering
	\subfloat[Quadratic\label{fig:grad-test-functions-quadratic}]{%
		\includegraphics[width=0.16\textwidth,valign=t]{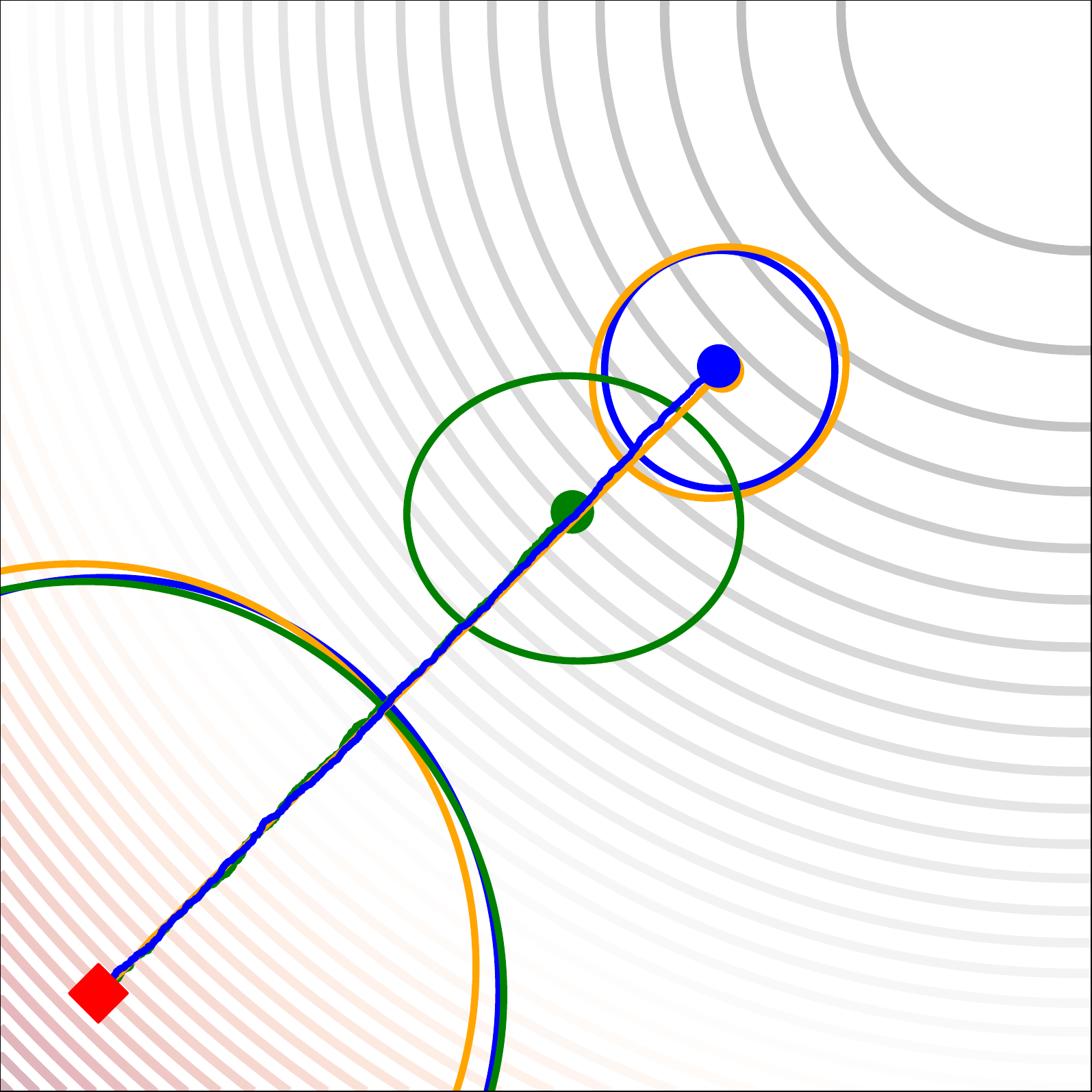}}
	\hfill
	\subfloat[Himmelblau]{%
		\includegraphics[width=0.16\textwidth,valign=t]{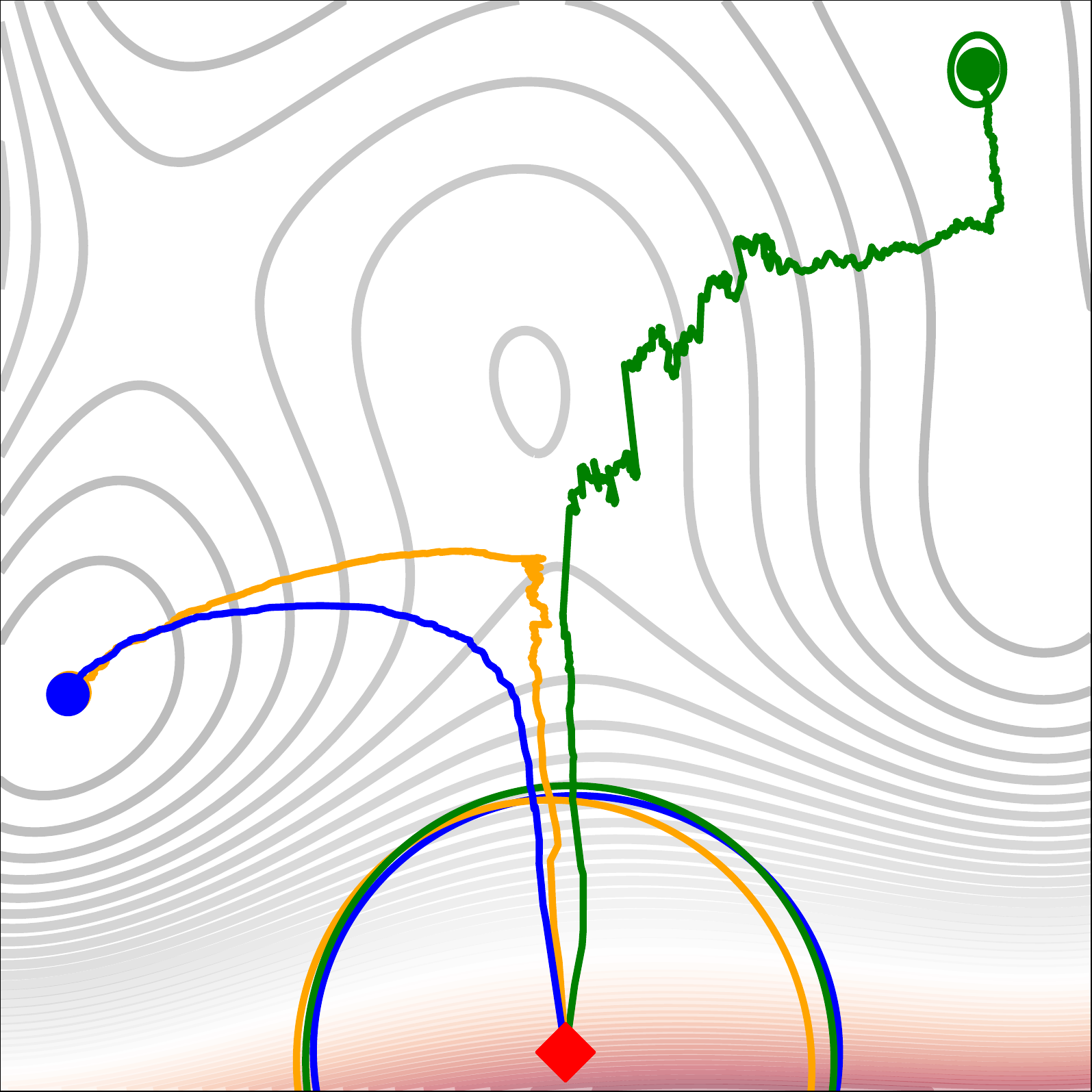}}
	\hfill
	\subfloat[Styblinski]{%
		\includegraphics[width=0.16\textwidth,valign=t]{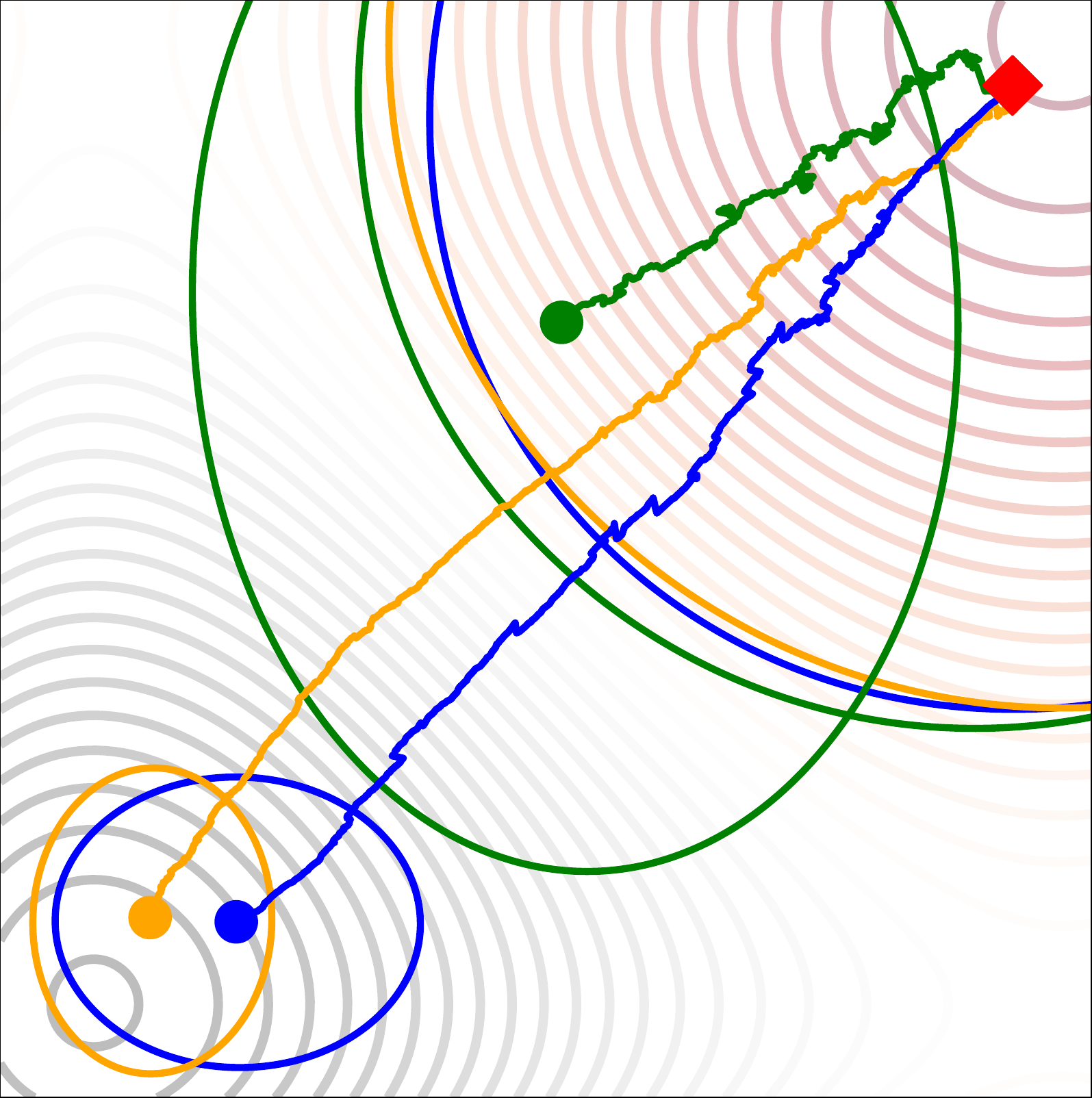}}
	\\
	\includegraphics[width=0.33\textwidth]{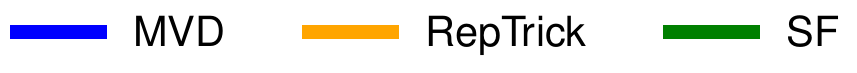}
	\caption{Sample run of the unbiased stochastic gradient estimators in optimization test functions for a fixed number of iterations. The lines show the means of the distributions, and the ellipses one standard deviation at the beginning (red diamond) and end (circle). Coloring: lower values in red, higher values in grey.}
	\label{fig:grad-test-functions} 
\end{figure}

\paragraph*{\textbf{Illustrative Example}}
Fig.~\ref{fig:grad-test-functions} shows the result of optimizing the expectation of common test functions under a 2-dimensional Gaussian distribution with diagonal covariance, using gradient ascent with the three referred methods.
The \gls{sf} uses the optimal baseline for black-box optimization based on the \gls{pgpe} algorithm~\cite{sehnke2010pgpe} to further reduce variance.
\gls{reptrick} and \gls{mvd} consistently move towards the global maximum in all test functions, while the \gls{sf} shows unstable behaviors. 
The comparison with the \gls{fd} method is not included since it is biased for finite $\Delta$, and for $\Delta \to 0$ it showed large variance.

\subsection{Parametrization of distributional parameters}
Distributional parameters $\distributionparams(\policyparams)$ can be parameterized further by other parameters $\policyparams$.
Consider for instance that the distribution $p$ represents a Gaussian stochastic policy with fixed covariance, and mean as the output of a neural network with state as input and parameters $\policyparams \in \R^M$, $p(\action | \state; \distributionparams(\policyparams)) = \Gaussian{\action | \state; \distributionparams=\{ \muvec(\state; \policyparams) \}, \covariance }$, with $\action \in \R^{D_a}$ and $\state \in \R^{D_s}$. Hence, $|\distributionparams| = D_a$.
Applying the chain rule, the gradient of \eqref{eq:stochasticobj} \wrt~$\policyparams$ becomes
\begin{align}
	\gradpolicyparams J\left( \distributionparams(\policyparams) \right) = \int \graddistributionparams p(\xvec; \distributionparams) f(\xvec) \de \xvec \gradpolicyparams \distributionparams(\policyparams),
	\label{eq:gradstochasticobj-chainrule}
\end{align}
which allows us to decompose the gradient in two parts -- the gradient \wrt~the distributional parameters $\distributionparams$ and the gradient \wrt~the parameters $\thetavec$. In the previous example we obtain $\gradpolicyparams p(\action | \state; \distributionparams(\policyparams)) = \gradmuvec \Gaussian{\action | \state; \muvec, \covariance} \gradpolicyparams \muvec(\state; \policyparams)$.
In practice, we can have a large neural network encoding the mean, but a tractable action space if $D_a \ll M$.
Despite the complexity of \glspl{mvd}, this fact still makes them tractable for policies with a large number of parameters, but considerable action space.

\begin{table*}[ht]
	\centering
	\begin{tabular}{ l l } 
		\hline
		\\
		\textbf{Policy gradient theorem} & $\graddistributionparamk J(\distributionparams) = \E{\state \sim \mu^{\pi}_{\gamma}}{\int \graddistributionparamk \policy(\action | \state; \distributionparams) Q^\policy(\state, \action) \de \action}$ \\
		\\
		\hline
		\\
		\textbf{Score-Function} & $\graddistributionparamk J(\distributionparams) = \E{\state \sim \mu^{\pi}_{\gamma}, \; \action \sim \policy(\cdot | \state; \distributionparams)}{\graddistributionparamk \log \policy(\action | \state; \distributionparams) Q^\policy(\state, \action)}$ \\
		\\
		\textbf{Reparametrization Trick} & $\graddistributionparamk J(\distributionparams) = \mathbb{E}_{\state \sim \mu^{\pi}_{\gamma}, \; \epsvec \sim p_{\epsvec}}\left[ \nabla_{\action} Q^\policy(\state, \action = h(\state, \epsvec; \distributionparams)) \graddistributionparamk h(\state, \epsvec; \distributionparams) \right] $ \\
		\\
		\textbf{Measure-Valued Derivative} &
		$\graddistributionparamk J(\distributionparams) = \mathbb{E}_{\state \sim \mu^{\pi}_{\gamma}} \left[ c_{\distributionparamk} \left( \E{\action \sim \policy_{\distributionparamk}^+(\cdot | \state; \distributionparams)}{Q^\policy(\state, \action)} -  \E{\action \sim \policy_{\distributionparamk}^-(\cdot | \state; \distributionparams)}{Q^\policy(\state, \action)} \right) \right]$
		\\
		\hline 
	\end{tabular}
	\caption{Policy gradient theorem with the different unbiased stochastic gradient estimators.}
	\label{tab:policygradient}
\end{table*}

\section{Actor-Critic policy gradient with MVDs}
\label{sec:stepbased_mvd}

The integral in \eqref{eq:on-policy-grad-theorem} is an application of \eqref{eq:gradstochasticobj} to \gls{rl}, as both are solving the same problem of computing an unbiased stochastic gradient estimate.
Hence, given the properties of \gls{mvd}, especially its low variance, we propose to analyze it in this setting.
In \gls{rl} it is common to only have access to an approximation of the true $Q$-function, which introduces extra bias and variance. Therefore, it is crucial to have a low variance estimator for the gradient of the expected return.

A straight forward application of \glspl{mvd} would be to perform black-box optimization to optimize the distributional parameters of an upper-level policy that models the distribution of a (typically deterministic) lower-level policy with parameters $\thetavec \in \R^M$~\cite{deisenroth2013}. In this case, $|\distributionparams| \geq \bigO{M}$, and the computational complexity discourages its application if $M$ is large. In contrast, the \gls{sf} complexity would be $\bigO{1}$.

\glspl{mvd} are also applicable in step-based actor-critic policy gradient methods with stochastic policies.
If an approximator for the critic is fitted from collected samples, policy updates are done without further interaction with the system, with queries to compute the gradient update done to the $Q$-function approximator and not estimated from the real environment.
This is in contrast with previous work in~\cite{bhatt2019pgweak}, where the policy gradient contribution for each state is computed by building an unbiased estimate of the $Q$-function for two different actions, by performing \gls{mc} rollouts.
We stress that this assumes we can perform two rollouts from the same state.
While it can be true when using a simulator, this is not generally applicable to \gls{rl}, where we assume the agent cannot reset to an arbitrary state, especially in real-world environments.
Furthermore, if the model of the environment is available, planning algorithms can be more efficient than \gls{rl} in solving a task.
Hence, applying \glspl{mvd} in actor-critic settings could be an interesting research direction.

Given a stochastic policy $\policy(\action | \state; \distributionparams = g(\state; \policyparams))$, where $\distributionparams$ are distributional parameters resulting from applying a function $g$ with parameters $\policyparams$ to the state $\state$, for instance neural networks that output the mean and covariance of a Gaussian distribution, the gradient \wrt~$\policyparams$ can be easily computed if $g$ is a continuous deterministic function, as in \eqref{eq:gradstochasticobj-chainrule}. From now on, we only consider the gradient \wrt~$\distributionparams$.

The policy gradient theorem~\cite{sutton1999PG} for a single parameter $\distributionparamk$ can thus be written with the three different estimators as shown in Table \ref{tab:policygradient}.
The \gls{mvd} formulation of the policy gradient shows that to compute the gradient of one distributional parameter we can sample from the discounted state distribution by interacting with the environment, and then evaluate the $Q$-function at actions sampled from the positive and negative components of the policy decomposition conditioned on the sampled states, $\policy_{\distributionparamk}^+(\cdot | \state; \distributionparams)$ and $\policy_{\distributionparamk}^-(\cdot | \state; \distributionparams)$, respectively.
Importantly, the $Q$-function estimate is the one from the policy $\policy$ and not from the ones resulting from the positive or negative decompositions.
We need a model for $Q$ that we can query without \gls{mc} sampling because we have to evaluate it for two actions starting from the same state. The function approximator for $Q$ can be a differentiable one, such as a neural network, as done in \gls{ddpg} or \gls{sac}, or a non-differentiable one, such as a regression tree.

Due to the high variance of the \gls{sf} estimator, the $Q$-function is replaced by the advantage function, which includes the value function as a baseline for variance reduction, while keeping the estimator unbiased. The benefit of \gls{mvd} over \gls{sf} is that they do not need to estimate an extra value function, which can be an extra source of bias and variance. Nevertheless, methods such as \gls{gae}~\cite{Schulmanetal_ICLR2016_gae} estimate the advantage function by only approximating the state value function, although introducing bias.

\section{Gradient Analysis in the LQR}
\label{sec:lqr_analysis}

\begin{figure*}[bh!]
	\noindent
	\centering
	\begin{minipage}[c]{0.333\textwidth}
		\subfloat[LQR  2 states 1 action]{%
			\includegraphics[width=1.0\linewidth,valign=t, ]{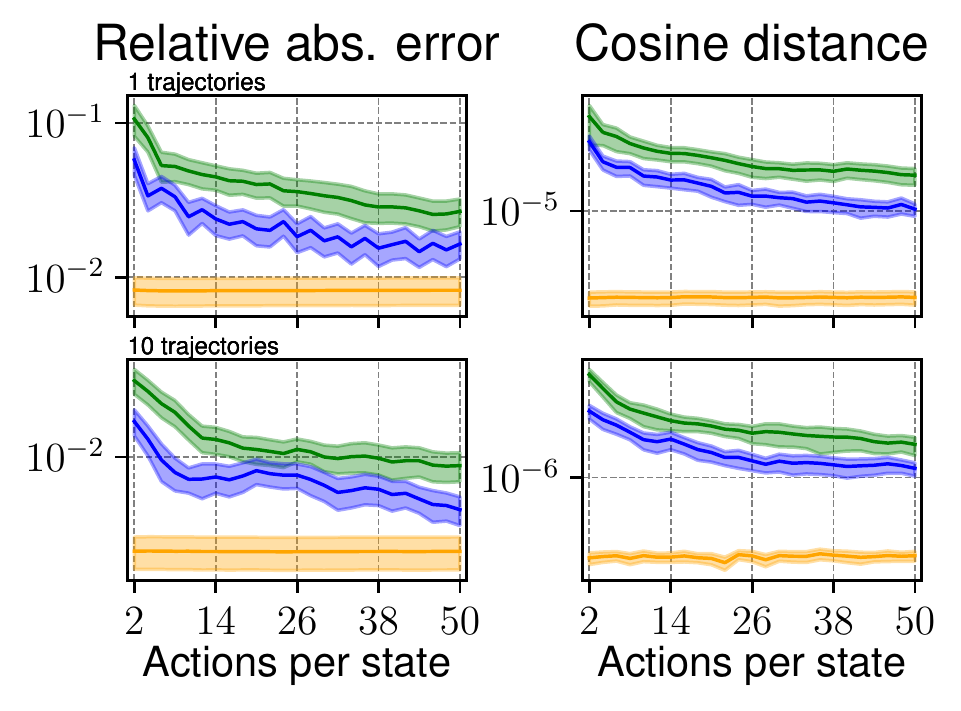} 
		}
	\end{minipage}%
	\begin{minipage}[c]{0.333\textwidth}
		\subfloat[LQR 2 states 2 actions]{%
			\includegraphics[width=1.0\linewidth,valign=t, ]{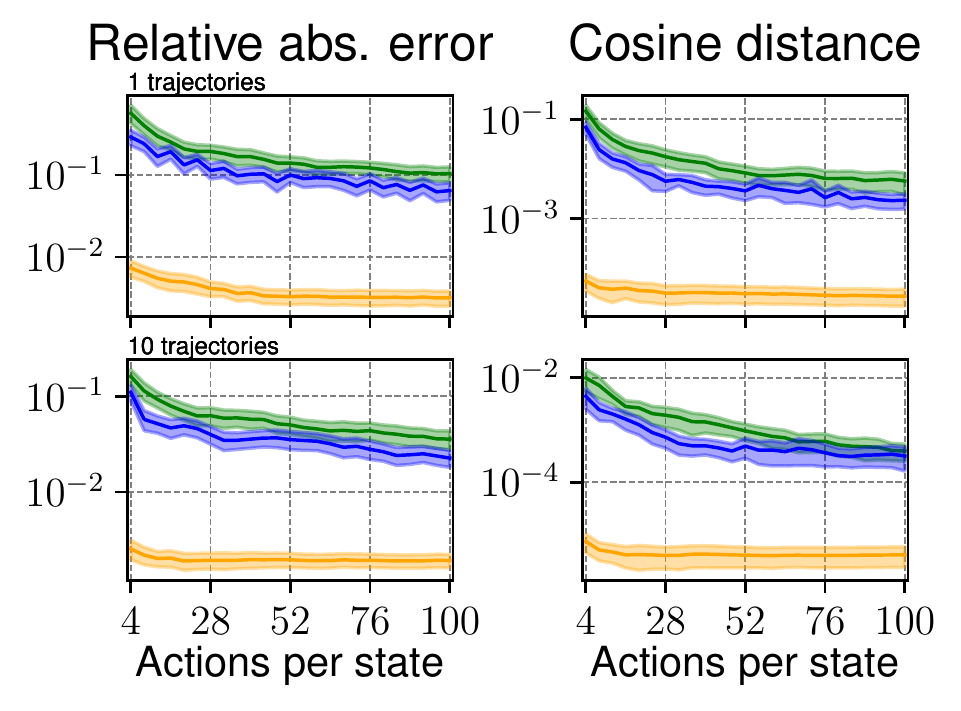} 
		}
	\end{minipage}%
	\begin{minipage}[c]{0.333\textwidth}
		\subfloat[LQR 4 states 4 actions]{%
			\includegraphics[width=1.0\linewidth,valign=t, ]{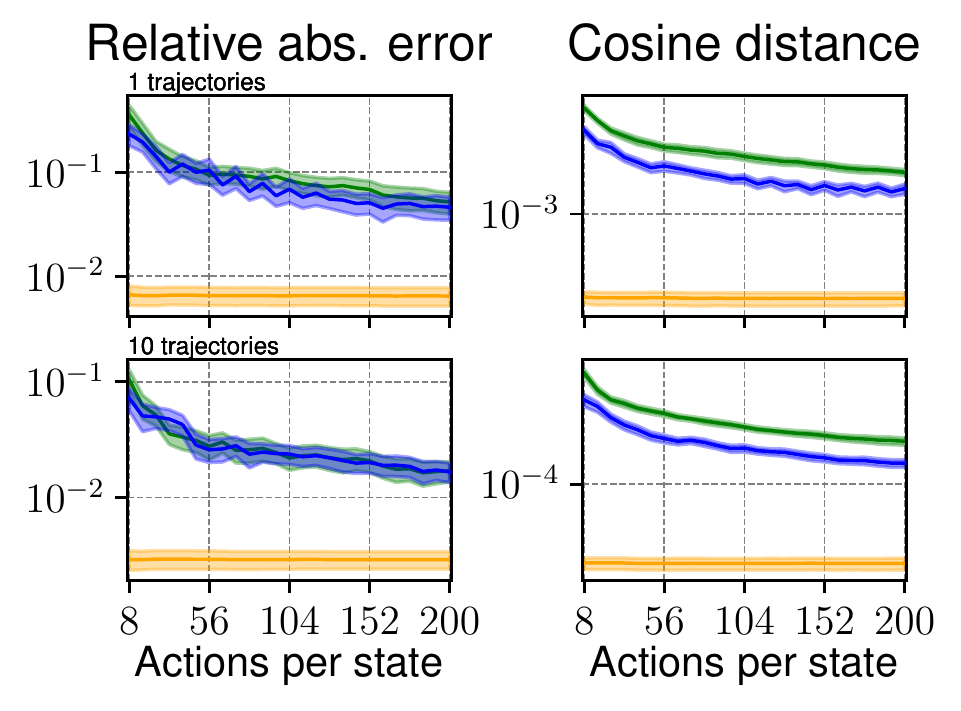} 
		}
	\end{minipage}%
	\\
	\includegraphics[width=0.3\linewidth,valign=t, ]{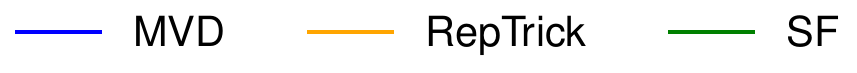} 
	\caption{Gradient errors in magnitude and direction in different \glspl{lqr}, per number of trajectories and sampled actions.  Depicted are the mean and the $95\%$ confidence interval of 25 random seeds.}
	\label{fig:lqr-errors} 
\end{figure*}

\begin{figure*}[bh!]
	\noindent
	\centering
	\begin{minipage}[c]{0.333\textwidth}
		\subfloat[LQR 2 states 1 actions]{%
			\includegraphics[width=1.0\linewidth,valign=t]{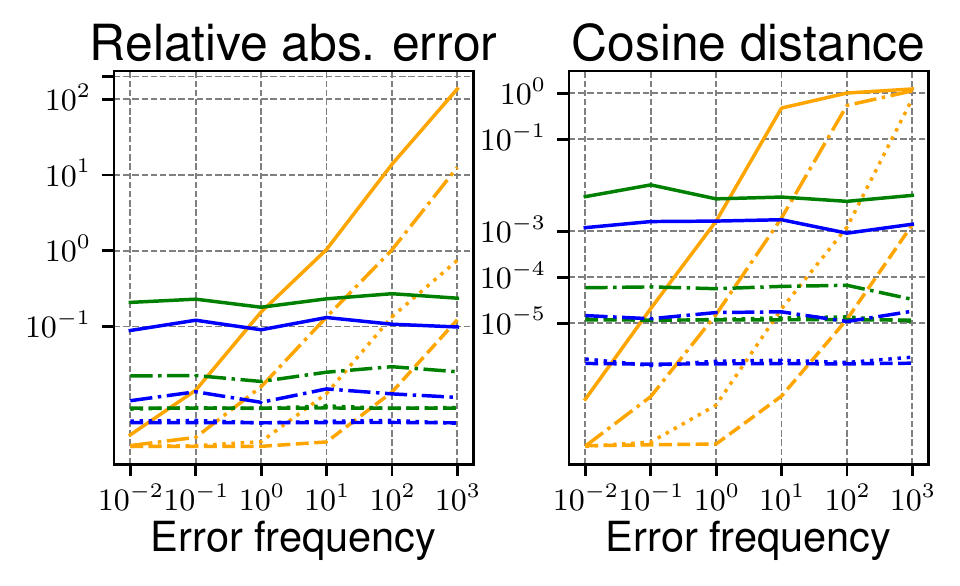} 
		}
	\end{minipage}%
	\begin{minipage}[c]{0.333\textwidth}
		\subfloat[LQR 2 states 2 actions]{%
			\includegraphics[width=1.0\linewidth,valign=t]{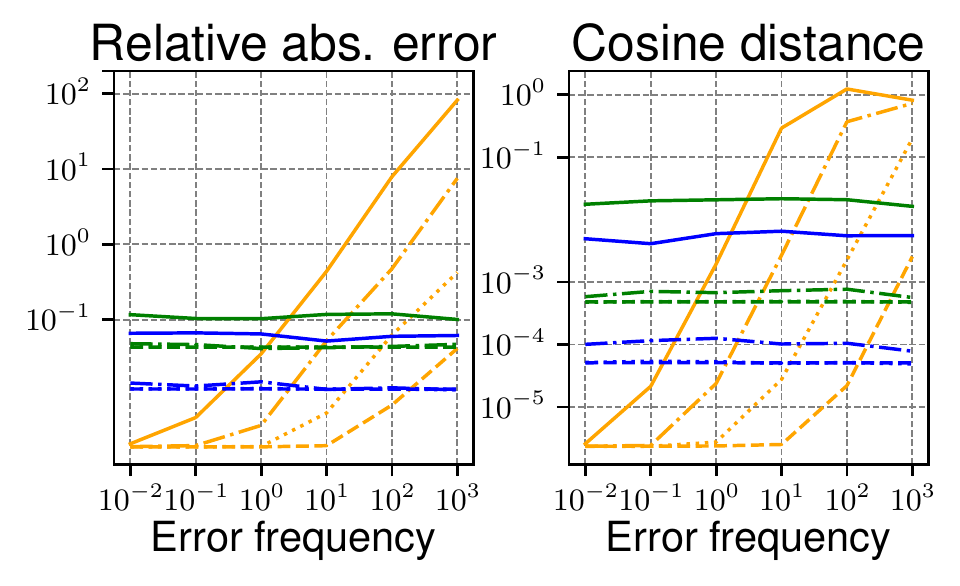} 
		}
	\end{minipage}%
	\begin{minipage}[c]{0.333\textwidth}
		\subfloat[LQR 4 states 4 actions]{%
			\includegraphics[width=1.0\linewidth,valign=t]{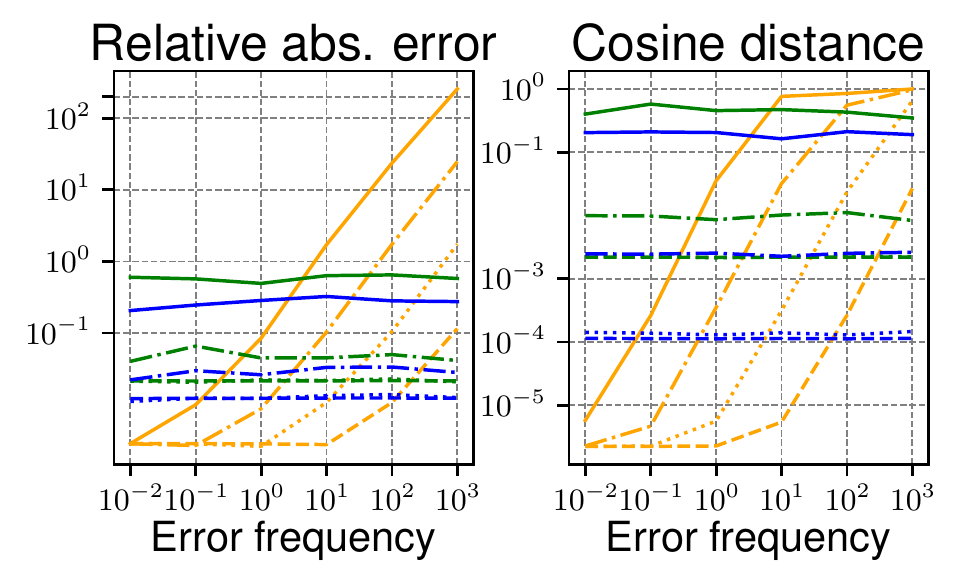} 
		}
	\end{minipage}%
	\\
	\includegraphics[width=0.7\linewidth,valign=t]{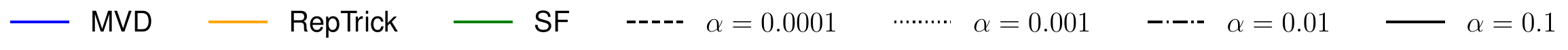} 
	\caption{Gradient errors in magnitude and direction in different \glspl{lqr}, with increasing error noise and frequency. The different linestyles correspond to different error amplitudes $\alpha$. The results are estimated using $10$ trajectories and $20 |\mathcal{\actionspace}|$ actions per state for each \gls{lqr}. For \gls{mvd} and \gls{sf} the amplitudes $0.0001$ and $0.001$ show identical results and thus the lines are superposed. Depicted is the mean over 25 random seeds.}
	\label{fig:lqr-noise} 
\end{figure*}

\begin{figure*}[bh!]
	\noindent
	\centering
	\begin{minipage}[c]{0.333\textwidth}
		\subfloat[LQR 2 states 1 actions]{%
			\includegraphics[width=1.0\linewidth,valign=t]{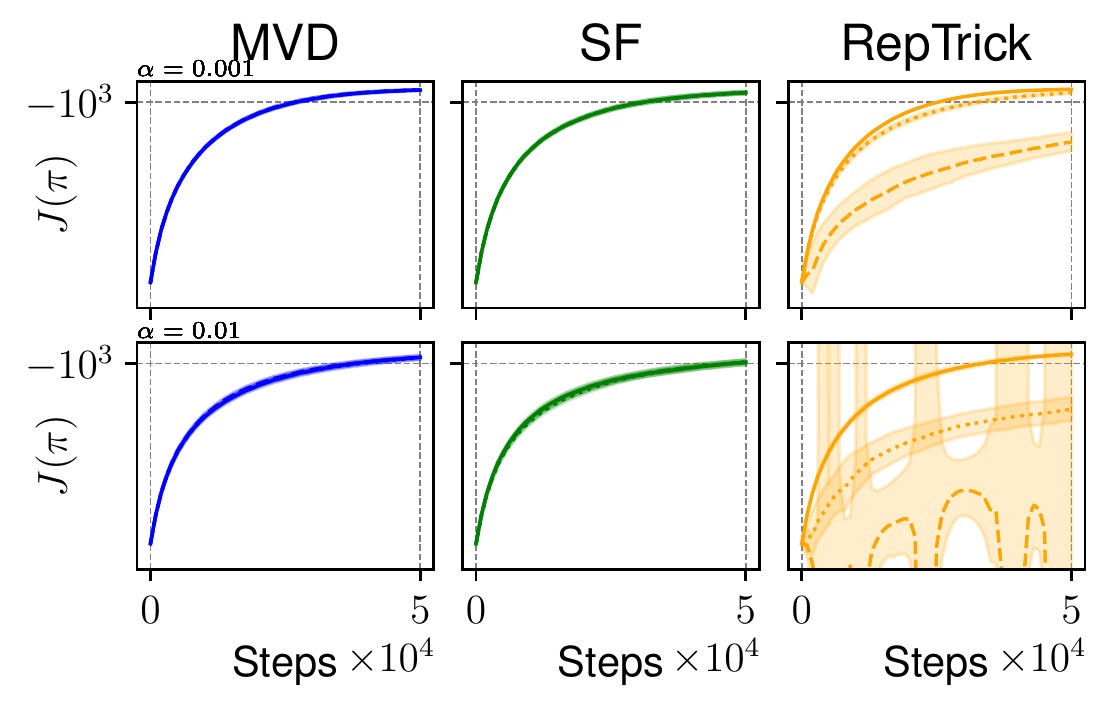} 
		}
	\end{minipage}%
	\begin{minipage}[c]{0.333\textwidth}
		\subfloat[LQR 2 states 2 actions]{%
			\includegraphics[width=1.0\linewidth,valign=t]{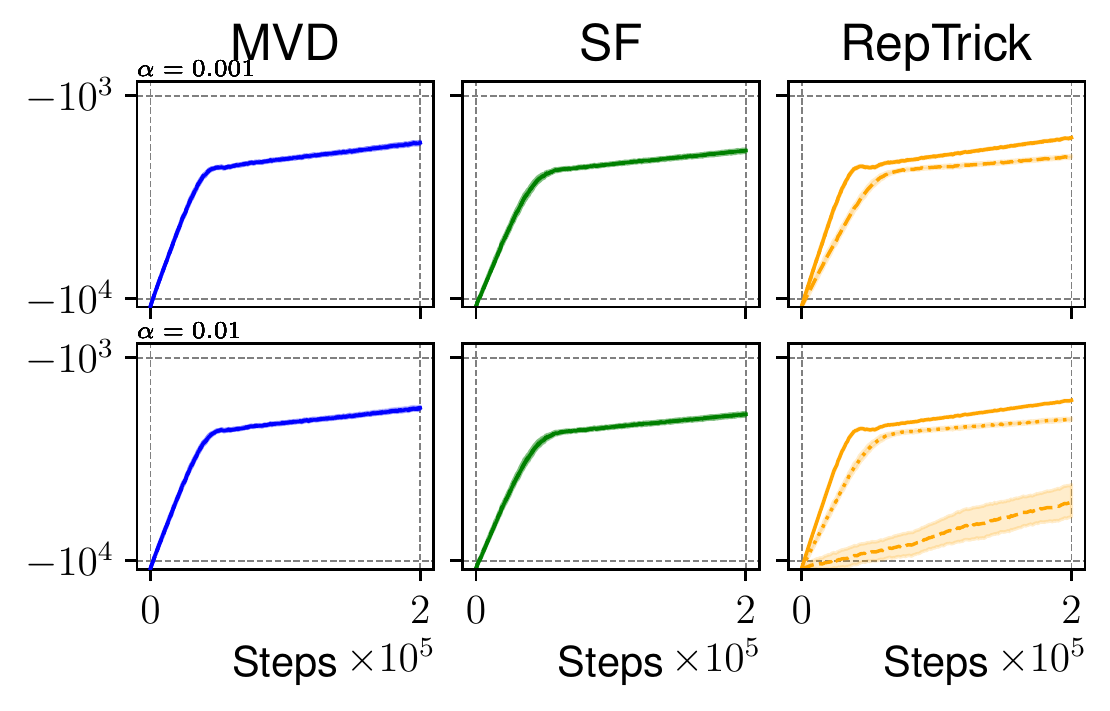} 
		}
	\end{minipage}%
	\begin{minipage}[c]{0.333\textwidth}
		\subfloat[LQR 4 states 4 actions]{%
			\includegraphics[width=1.0\linewidth,valign=t]{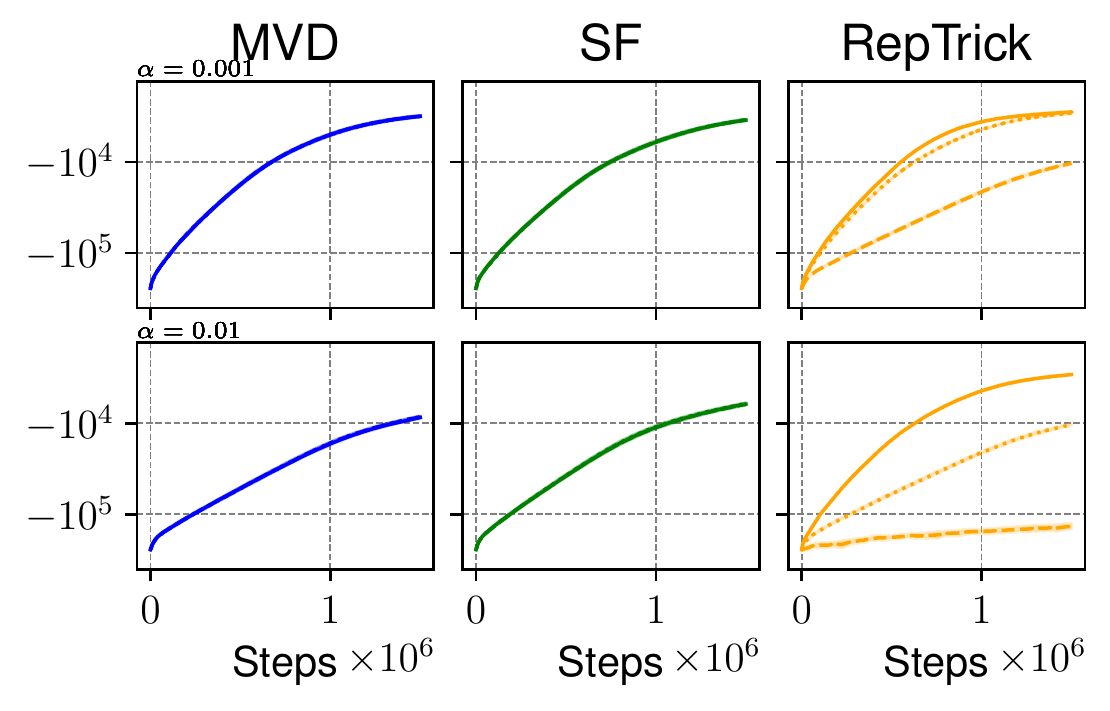} 
		}
	\end{minipage}%
	\\
	\includegraphics[width=0.65\linewidth,valign=t]{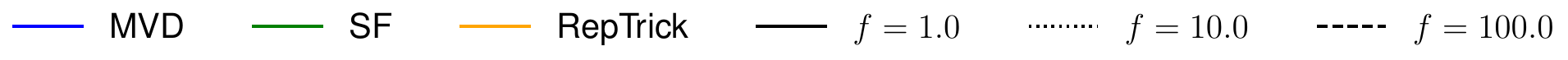} 
	\caption{Learning curves of expected return per collected transitions, for different error amplitudes and frequencies. The gradients are estimated using $1$ trajectory and $2|\mathcal{\actionspace}|$ actions per state for each \gls{lqr}. Depicted are the mean and the $95\%$ confidence interval of 25 random seeds. The lines for different frequencies and the confidence intervals appear superposed in some plots.}
	\label{fig:lqr-noise-training} 
\end{figure*}

The \gls{lqr} is a well-studied problem in control theory. Here we consider the discounted infinite-horizon discrete-time \gls{lqr} under a Gaussian stochastic policy $\action \sim \Gaussian{\cdot | -\mat{K}\state, \covariance} $, where $\mat{K}$ is a learnable feedback gain matrix, and $\covariance$ a fixed covariance. As the value function expression is known and can be computed by numerically solving the Algebraic Riccati Equation, as well as the gradient \eqref{eq:on-policy-grad-theorem} \wrt~$\mat{K}$, the \gls{lqr} is a good baseline for policy gradient algorithms.

In this analysis we construct \glspl{lqr}, with different state and action dimensions, with dynamics such that the uncontrolled systems are unstable, and then select a suboptimal initial gain $\mat{K}_{\mathrm{init}}$ such that the closed-loop is stable.
All environments have a fixed initial state.
We analyze the policy gradient with the estimators from Table~\ref{tab:policygradient}, computed at the initial gain matrix and state. In the \gls{sf} the $Q$-function is replaced by the advantage function for variance reduction.
The expectation of the on-policy state distribution is sampled directly from the environment, but the expectation over actions uses critics $Q$ and $V$ as the true value functions from the \gls{lqr}.
This is an idealized scenario to analyze the estimators' variances when there are no errors in the function approximator.

\begin{figure*}[!b] 
	\centering
	\includegraphics[width=0.95\textwidth,valign=t]{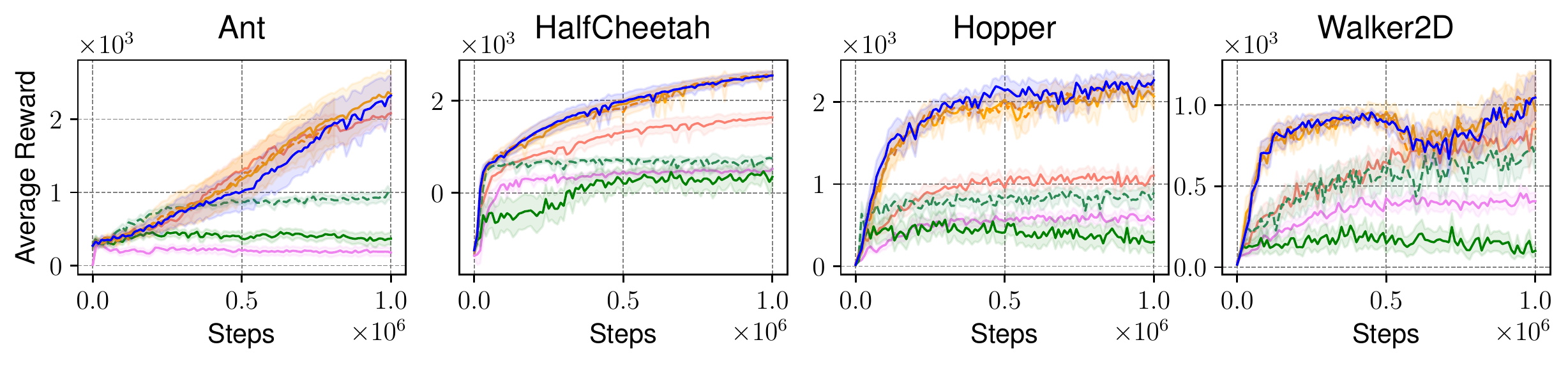}
	\\
	\includegraphics[width=1.0\textwidth]{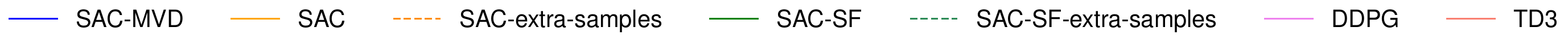}
	\caption{Policy evaluation results during training on different tasks in deep \gls{rl}. Depicted are the average reward per samples collected and the $95\%$ confidence interval of 25 random seeds.}
	\label{fig:step-based-experiments}
\end{figure*}

We compare two sources of error. The \textit{relative absolute error} relates the norms of the estimated $\hat{\vec{g}}$ and true gradient $\vec{g}$ as $ \left(| \|\hat{\vec{g}} \| - \| \vec{g} \| | \right) / \| \vec{g} \| $, i.e. $0$ indicates they have the same magnitude, and the \textit{cosine distance} as the error in the gradient direction as $1 - \hat{\vec{g}}^\transpose\vec{g}/(\|\hat{\vec{g}} \| \| \vec{g} \|) $, i.e. $0$ indicates collinearity. An ideal estimator has both measures close to zero.

The errors are analyzed along two dimensions -- the number of trajectories and the number of actions sampled to solve the action expectations.
Using \gls{mvd} the gradient \wrt~the mean needs two function evaluations per action dimension.
For a fair comparison, the \gls{reptrick} and \gls{sf} sample the same number of actions the \gls{mvd} needs for the gradient of all parameters. 
This way the computation complexity is not exactly the same, since for the \gls{reptrick} we still need to compute the gradient of $Q$ and for \gls{sf} the log-probability derivative, but gives a fair comparison than just using one \gls{mc} estimate.
The results are shown in Fig.~\ref{fig:lqr-errors}.
As expected, with the number of trajectories fixed, increasing the number of sampled actions decreases the estimators' errors.
The \gls{reptrick} achieves the best results in both magnitude and direction in all environments, and the \gls{mvd} is slightly better than the \gls{sf}.
In our experiments we observed this holds for higher dimensions as well.
This reveals that even though the \gls{sf} complexity is $\bigO{1}$, in practice we need roughly the same number of samples as \gls{mvd} to obtain a low error gradient estimate.
A more in-depth analysis of this fact is left for future work.
Knowing that the $Q$-function is quadratic in the action, the results are in line with the example from Fig.~\ref{fig:grad-test-functions-quadratic}, where all estimators perform well.

Next, we consider a scenario where the true $Q$-function is not available and has to be estimated.
Instead of fitting a state-action value function, we model the approximator with a local approximation error on top of the true value as $\hat{Q}(\state,\action) = Q(\state,\action) + \alpha Q(\state,\action)\cos(2\pi f \vec{p}^\transpose \vec{a} + \phi)$.
I.e., we use an additive sinusoidal error proportional to the true estimate, where $\alpha$ represents the fraction of the true $Q$ amplitude, $f$ is the error frequency, $\vec{p}$ is a random vector whose entries sum to 1, and $\phi\sim\mathcal{U}[0, 2\pi]$ a phase shift.
$\vec{p}$ and $\phi$ introduce randomness to remove correlation between action dimensions.
Fig.~\ref{fig:lqr-noise} depicts the gradient errors in magnitude and direction as a function of the error frequency and amplitude, which reveals two insights.
The error frequency does not affect the gradient estimation of \gls{sf} or \gls{mvd}, which are only affected by the amplitude, as can be seen by the horizontal blue and green lines. On the contrary the \gls{reptrick} is heavily affected by both variables.
This result matches the theory, since under a Gaussian distribution the variance of the \gls{reptrick} is upper-bounded by the Lipschitz constant of the derivative of $Q$~\cite{mohamed2019monte}. 
Fig.~\ref{fig:lqr-noise-training} shows the expected discounted return per steps taken in the environment with different amplitudes and frequencies of errors. As expected, the \gls{sf} and \gls{mvd} do not suffer from errors in the estimation and converge towards the optimal policy in the presence of high-frequency error terms. The \gls{reptrick} on the other hand either shows slower convergence or fails to converge due to the poor gradient estimates.

From these experiments we observe that even though the \gls{reptrick} provides the most precise and accurate gradient under a true value function, this does not always hold, especially in the case where there is an action correlated error in the approximation.

\section{MVDs in Off-Policy Policy Gradient \texorpdfstring{\\ for Deep Reinforcement Learning}{}}
\label{sec:mvd_deeprl}

In the following, we illustrate how \glspl{mvd} can be used in a deep \gls{rl} algorithm, by using the off-policy actor-critic method \gls{sac}. The optimization objective is similar to the off-policy gradient~\cite{Degris2012OffPAC} but with a soft $Q$-function, which includes an entropy regularization term. This additional term encourages exploration by preventing the policy from becoming too deterministic during learning.
The surrogate objective optimized by \gls{sac} is
\begin{align*}
    J_{\policy}(\distributionparams) = \E{\state \sim d^{\beta}, \action \sim \policy(\cdot | \state; \distributionparams)}{Q^{\policy}(\state, \action; \phivec) -\alpha \log \policy (\action | \state ; \distributionparams)  },
\end{align*}
where $d^{\beta}$ is an off-policy state distribution, the $Q$-function is a neural network parameterized by $\phivec$, $\alpha$ weighs the entropy regularization term, and $\distributionparams$ are the distributional parameters of $\policy$.
For simplicity, we omit the parameterization of $\distributionparams$ as its extension is done as in \eqref{eq:gradstochasticobj-chainrule}.

\gls{sac} estimates the gradient \wrt~$\distributionparams$ with the \gls{reptrick}.
Instead, we modify it by computing the gradient using \gls{mvd}. Defining $f(\state, \action; \phivec, \distributionparams) = Q^{\policy}(\state, \action; \phivec) -\alpha \log \policy (\action | \state ; \distributionparams)$, the resulting off-policy policy gradient \wrt~a parameter $\distributionparamk$ becomes
\begin{align*}
    & \graddistributionparamk J_{\policy}(\distributionparams) = \mathbb{E}_{\state \sim d^{\beta}} \left[ c_{\distributionparamk} \left( \mathbb{E}_{\action \sim \policy_{\distributionparamk}^+(\cdot | \state; \distributionparams)} \left[ f(\state, \action; \phivec, \distributionparams) \right] \right. \right. \\
    & \qquad\qquad\qquad\qquad\qquad\quad \left. \left. -  \mathbb{E}_{\action \sim \policy_{\distributionparamk}^-(\cdot | \state; \distributionparams)} \left[f(\state, \action; \phivec, \distributionparams) \right] \right) \right. \\
    & \qquad\qquad\qquad\qquad  \left.  - \mathbb{E}_{\action \sim \policy(\cdot | \state; \distributionparams)} \left[ \graddistributionparams \alpha \log \policy (\action | \state ; \distributionparams) \right] \right],
\end{align*}
where $\policy_{\distributionparamk}^+$ and $\policy_{\distributionparamk}^-$ are the positive and negative components of the policy decomposition. We denote this modification as \gls{sacmvd}. Unlike in \gls{sac}, we do not require the $Q$-function to be differentiable, allowing the use of any type of function approximators.

\subsection*{Experimental Results}

We benchmark the performance of \acrshort{sac} with the different gradient estimators in high-dimensional continuous control tasks, based on the PyBullet simulator~\cite{coumans2019bullet}. We compare the original \acrshort{sac} with: \acrshort{sacmvd} with one \gls{mc} sample; \acrshort{sacsf} - a version of \acrshort{sac} using the \acrshort{sf}; \acrshort{sacsfextrasamples} - same as \acrshort{sacsf} but with the same number of queries as \acrshort{sacmvd} per gradient estimate; \acrshort{sacextrasamples} - same as \acrshort{sac} but with the same number of gradient estimates as \acrshort{sacsfextrasamples}; \acrshort{ddpg}~\cite{Lillicrap2016DDPG} and \acrshort{td3}~\cite{fujimoto2018td3} - two state-of-the-art off-policy algorithms, which are used as baseline comparisons.
All variations of \gls{sac} use the same hyperparameters and neural network architectures for the value function and policy as in the original work~\cite{Haarnoja2018SAC}.

The average reward curves obtained during training are shown in Fig.~\ref{fig:step-based-experiments}.
The original version of \gls{sac} and \gls{sacmvd} show similar performance, and increasing the number of samples does not improve the results, as \acrshort{sacextrasamples} performs equally well.
This empirically shows that the \gls{mvd} estimator and the \gls{reptrick} are equally good to provide precise and accurate gradient estimates in these high-dimensional tasks.
\acrshort{sacsf} and \acrshort{sacsfextrasamples} fail to solve most of the tasks, since no baseline was used for variance reduction. This can be done with \gls{gae} or by estimating an extra value function. However, that can introduce more bias in the estimation. Therefore, we chose to not use a baseline for this experiment to compare directly all estimators in the base case.

The results of our experiments reveal that the \gls{reptrick} is not fundamental to replicate the performance of \gls{sac}, and suggest that the superior performances of this algorithm depend on other aspects, such as the entropy regularization, the state-dependent covariance, and the squashed Gaussian policy.
Additionally, it is worth remembering that \glspl{mvd} are applicable to function classes where the \gls{reptrick} is not, which allows to explore other function approximator classes and still use the benefits of \gls{sac}.

\section{MVDs in On-Policy Policy Gradient\texorpdfstring{\\ for Non-differentiable Approximators}{}}
\label{sec:mvd_nondiff_critic}

In this section we present the results of the on-policy gradient with \glspl{mvd} when using a non-differentiable $Q$-function approximator.
The goal is to show that when the \gls{reptrick} is not applicable, \gls{mvd}-based methods can obtain comparable or better results than \gls{sf} ones.

For the $Q$-function approximator we use Extra-Trees, a type of regression tree, due to their interpretability, low variance and bias properties, and better computational complexity compared to other tree methods~\cite{Geurts2006ExtraTrees}. Algorithm \ref{alg:tree-mvd-pg} shows a pseudo-code for the method we use -- \gls{treemvdpg}.
Fitting the $Q$-function in line \ref{treemvd:line:q} is done by applying Bellman Equation, $Q^{\policy}(\state, \action) = r(s,a) + \gamma \E{\nextstate, \action'}{Q^{\policy}(\nextstate, \action')}$, with bootstrapping for a fixed amount of iterations, where $\state$, $\action$ and $\nextstate$ are samples collected on-policy and from a replay buffer, as is common in recent \gls{rl} algorithms~\cite{mnih2015humanlevel}. The latter helps in the generalization out of the current on-policy samples.
The expectation over the next-action is solved with a single sample.
This procedure is an extension of \acrshort{expected-sarsa}~\cite{Sutton2018IntroRL}.

\begin{algorithm}[ht]
    \footnotesize
	\caption{\glsfirst{treemvdpg}}
	\label{alg:tree-mvd-pg}
	\SetAlgoLined
	\KwIn{Stochastic policy $\policy(\action | \distributionparams(\state; \policyparams))$; $E$ epochs; $R$ replay buffer samples; $P$ policy updates; $B$ batch size policy update; $N$ Monte Carlo samples; learning rate $\alpha$}
	\KwResult{Optimized policy parameters $\policyparams^*$}
	$\mathcal{B} \leftarrow $ start a replay buffer \\
	\For(\tcp*[f]{epochs}){$e=0$ \ldots $E-1$}{
	    $\mathcal{D}_{\policy} \leftarrow $ collect data on-policy in the environment \\
        $\mathcal{D}_{\mathcal{B}} \leftarrow $ get $R$ samples from $\mathcal{B}$  \\
	    $\mathcal{B} \leftarrow \mathcal{B} \cup \mathcal{D}_{\policy} $ augment the replay buffer \\
	    Fit $Q^{\policy}$ using Extra-Trees with $\mathcal{D}_{\policy} \cup \mathcal{D}_{\mathcal{B}}$ \label{treemvd:line:q} \\
		\For(\tcp*[f]{policy updates}){$p=0$ \ldots $P-1$}{
		    $\statespace \leftarrow $ Sample $B$ states from $D_{\policy}$ \\
		    \For{$\state \in \statespace$}{
		        \For(\tcp*[f]{dist params}){$k=0$ \ldots $K-1$}{
			        $\hat{g}_k = c_{\distributionparamk} \frac{1}{N} \left( \sum_{i=1}^N Q^{\policy}(\state, \action_i^+) - Q^{\policy}(\state, \action_i^-) \right)$\\
					$\action_i^+ \sim \policy_{\distributionparamk}^+(\cdot| \distributionparams(\state; \policyparams)),\; \action_i^- \sim \policy_{\distributionparamk}^-(\cdot| \distributionparams(\state; \policyparams)) $
		        }
		        $\hat{\vec{g}}_{\state}^{\textrm{MVD}} = \left[\hat{g}_0, \ldots, \hat{g}_{K-1} \right]^\transpose$ \\
		        $\hat{\vec{g}}_{\state} = \nabla_{\policyparams} \distributionparams (\state; \policyparams)  \hat{\vec{g}}_{\state}^{\textrm{MVD}} $ \tcp*[f]{grad estimate}  \\
		    }
		    $\hat{\vec{g}} = \textrm{average} \left( \left[ \hat{\vec{g}}_{1}, \ldots, \hat{\vec{g}}_{|\statespace|}   \right]  \right) $ \\
	        $\policyparams \leftarrow \policyparams + \alpha \hat{\vec{g}} $ \\
		}
	} 
	
\end{algorithm}

\subsection*{Experimental Results}

\begin{wrapfigure}{l}{0.25\textwidth}
	\setlength{\fboxsep}{0pt}
	\vspace{-16pt}
	\subfloat[Corridor]{
		\fbox{\includegraphics[scale=0.27]{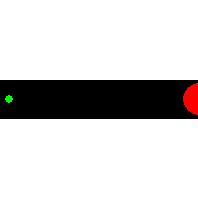}}
	}
	\subfloat[Room]{
		\fbox{\includegraphics[scale=0.27]{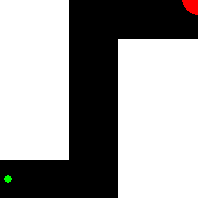}}
	}
	\\
	\caption{Corridor and Room environments.}
	\label{fig:corridor-room} 
	\vspace{-10pt}
\end{wrapfigure}

We evaluate this method in four environments with continuous state and action spaces.
The first two tasks are the Pendulum-v0 and LunarLanderContinuous-v2 from OpenAI Gym~\cite{brockman2016openai}, which are common baselines for continuous control.
The remaining two are the Corridor and Room, which are depicted in Fig.~\ref{fig:corridor-room}.
In both environments the agent starts at the green dot and the goal is to move towards the red area, where the episode terminates (or when a fixed amount of steps is reached).
The state is the $(x, y)$ position of the agent and the action the velocity vector, with bounded norm $\sqrt{2}$.
The reward function is the sum of a negative constant for being in the white area or hitting the wall, the negative euclidean distance between the current agent position and the goal, and a positive reward for reaching the red area. With these two environments we want to argue that due to the task structure, using a non-smooth approximator, such as a regression tree, is preferable to a smooth one, e.g. a neural network. 

\begin{figure}[ht]
    \noindent
    \centering
    \begin{minipage}[c]{0.45\textwidth}
        \subfloat[Pendulum Tree-MVD]{%
            \includegraphics[width=0.32\linewidth,valign=t, trim={1.5cm 1.8cm 0.8cm 0.8cm}, clip]{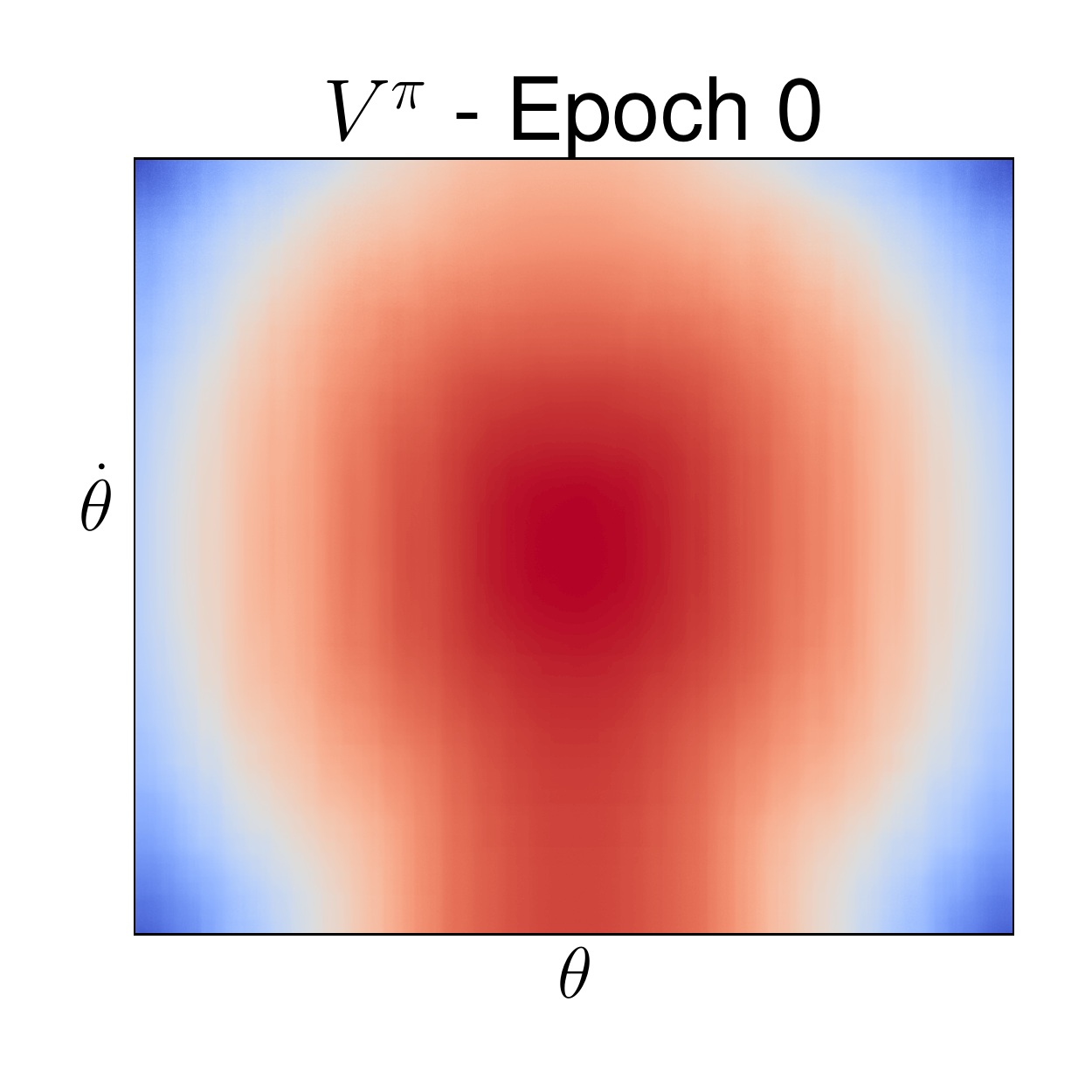} 
            \includegraphics[width=0.32\linewidth,valign=t, trim={1.5cm 1.8cm 0.8cm 0.8cm}, clip]{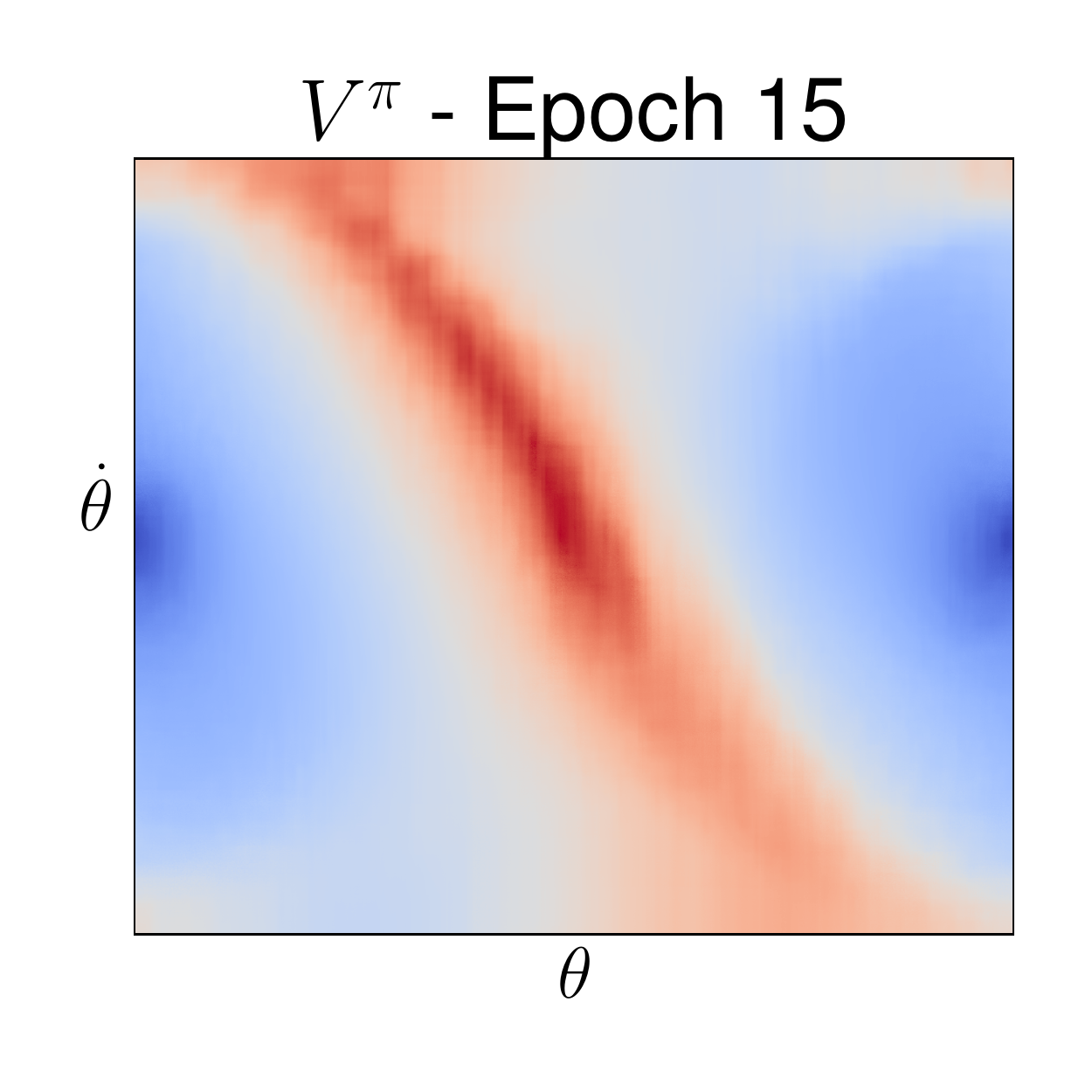}
            \includegraphics[width=0.32\linewidth,valign=t, trim={1.5cm 1.8cm 0.8cm 0.8cm}, clip]{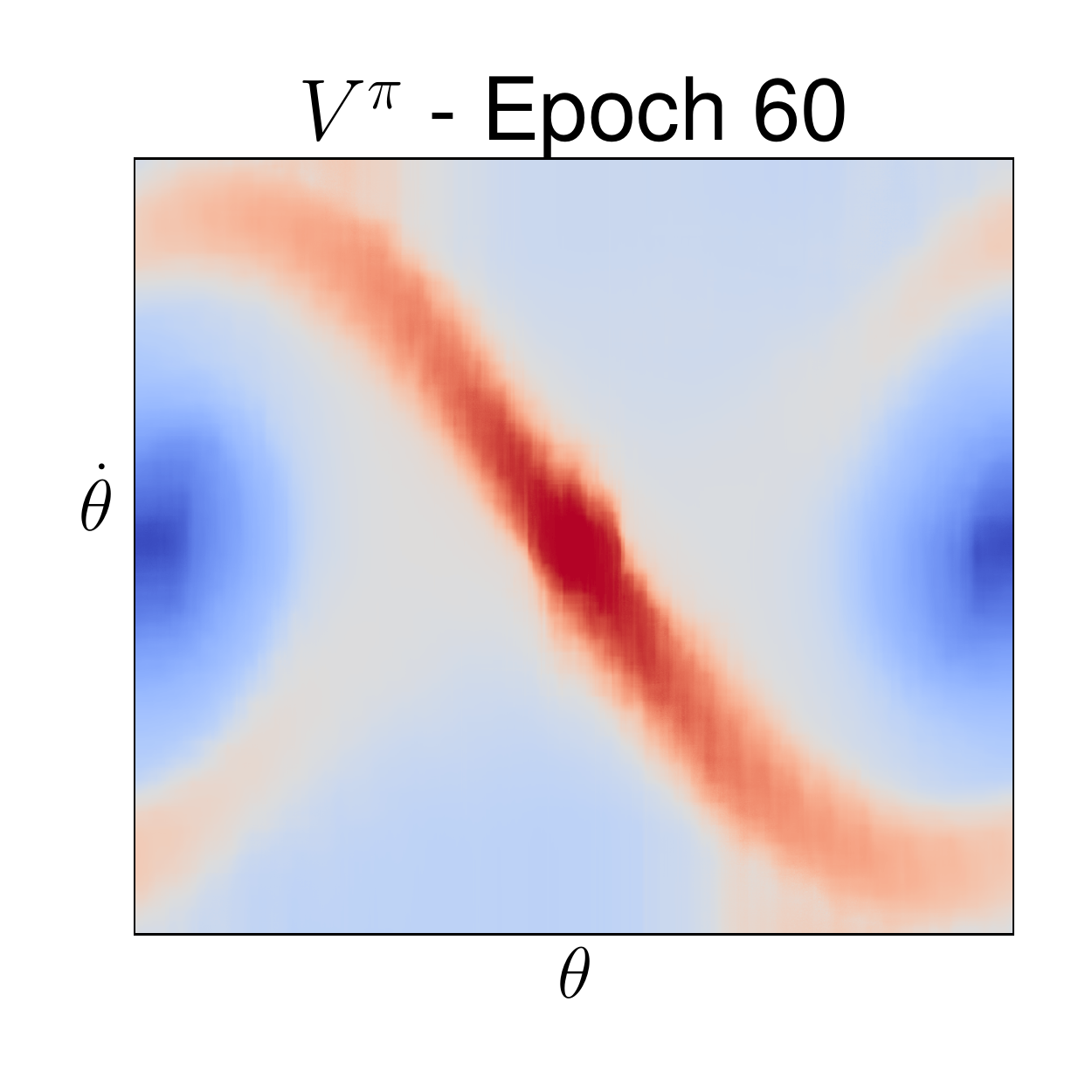}
        }
    \end{minipage}%
	\\
    \begin{minipage}[c]{0.45\textwidth}
        \subfloat[Pendulum PPO]{%
            \includegraphics[width=0.32\linewidth,valign=t, trim={1.5cm 1.8cm 0.8cm 1.85cm}, clip]{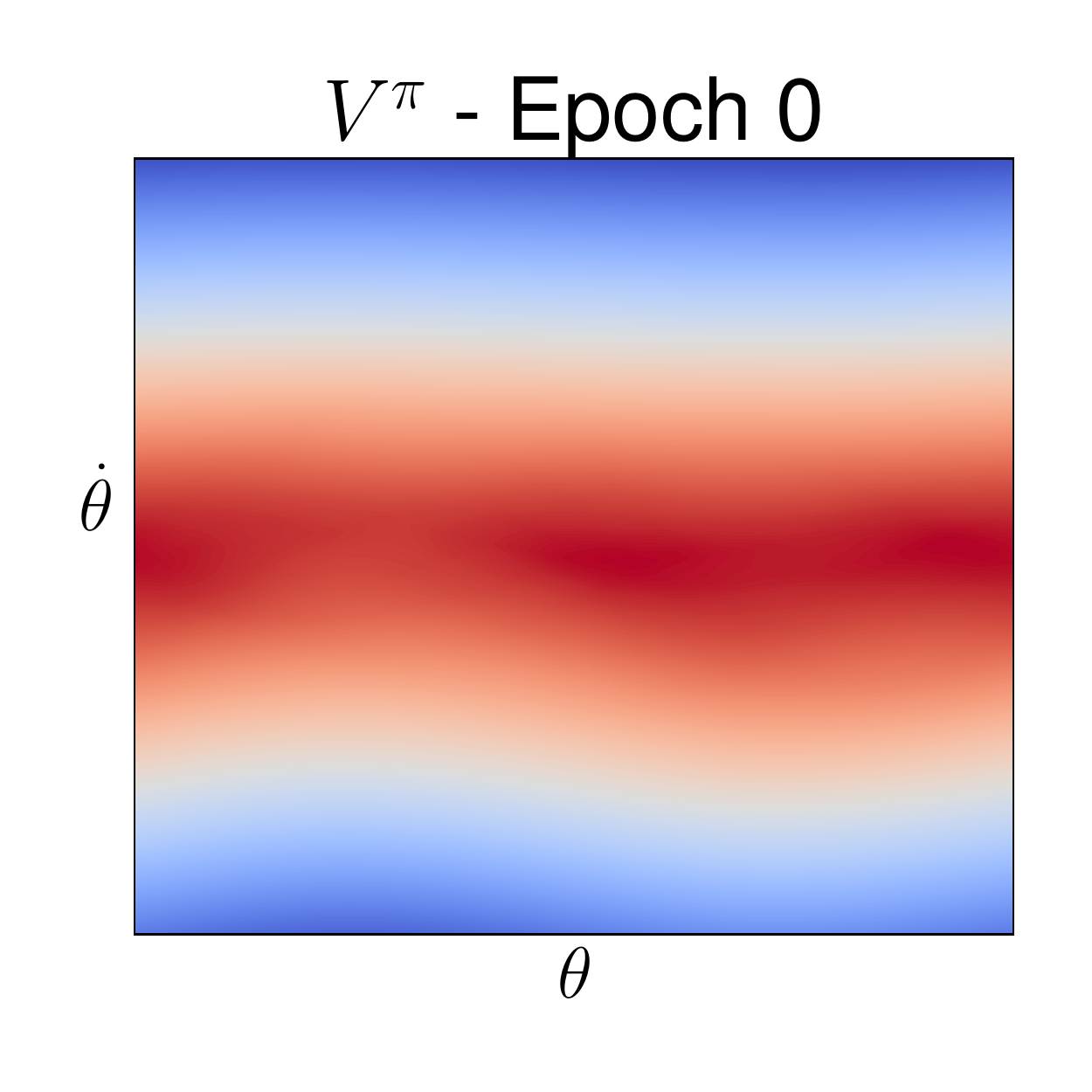} 
            \includegraphics[width=0.32\linewidth,valign=t, trim={1.5cm 1.8cm 0.8cm 1.85cm}, clip]{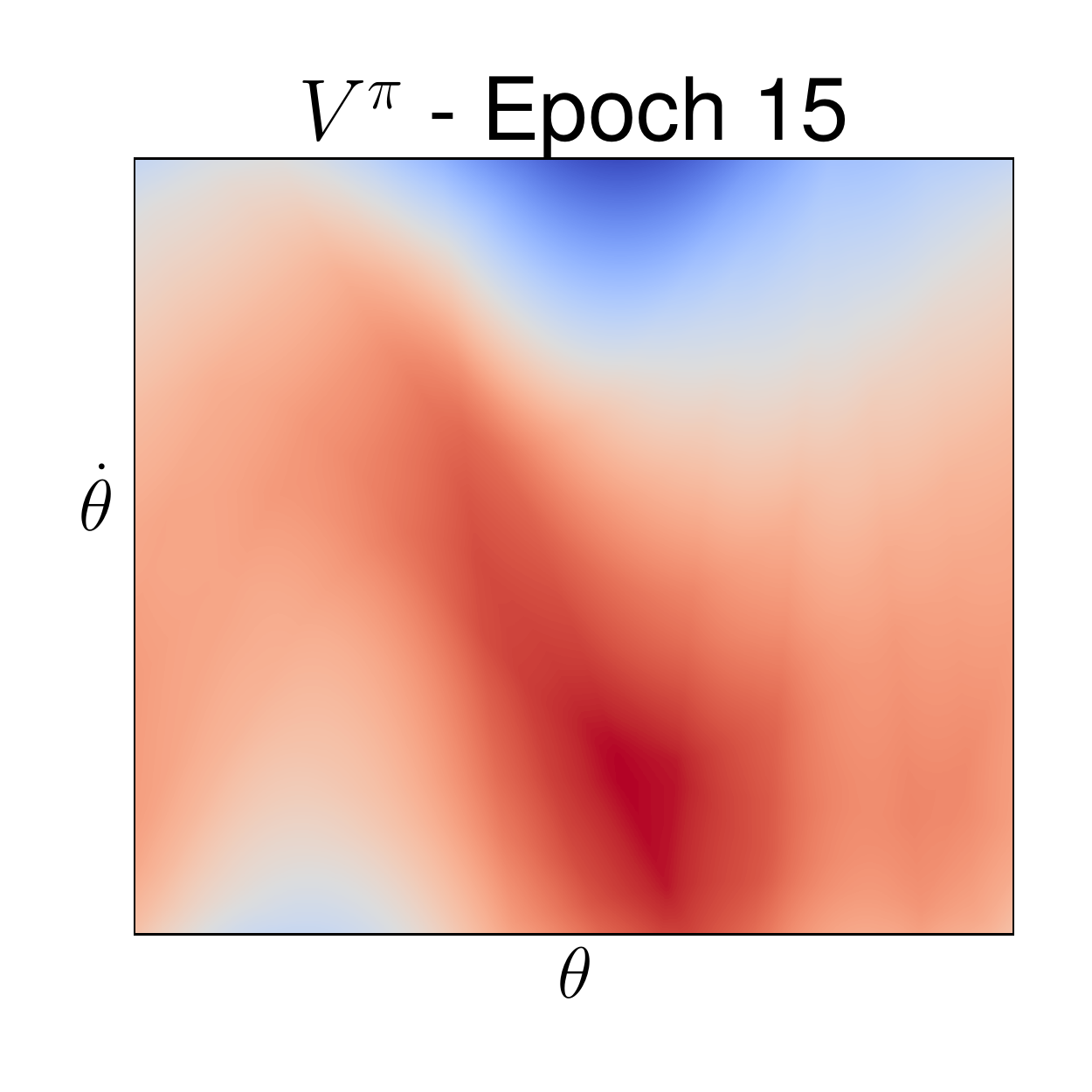}
            \includegraphics[width=0.32\linewidth,valign=t, trim={1.5cm 1.8cm 0.8cm 1.85cm}, clip]{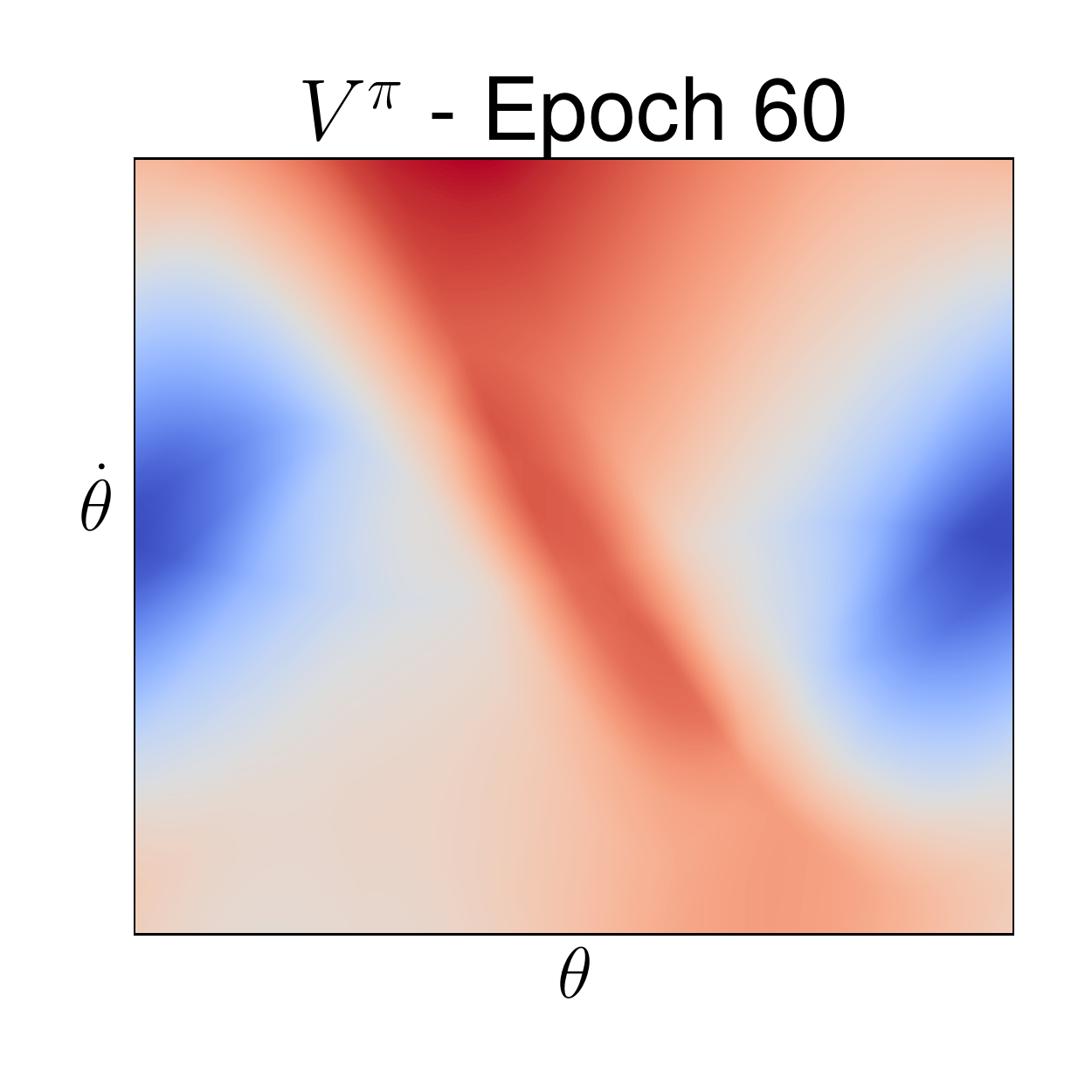}
        }
    \end{minipage}%
    \\
    \begin{minipage}[c]{0.45\textwidth}
        \subfloat[Corridor Tree-MVD]{%
            \includegraphics[width=0.32\linewidth,valign=t, trim={1.5cm 1.8cm 0.8cm 1.85cm}, clip]{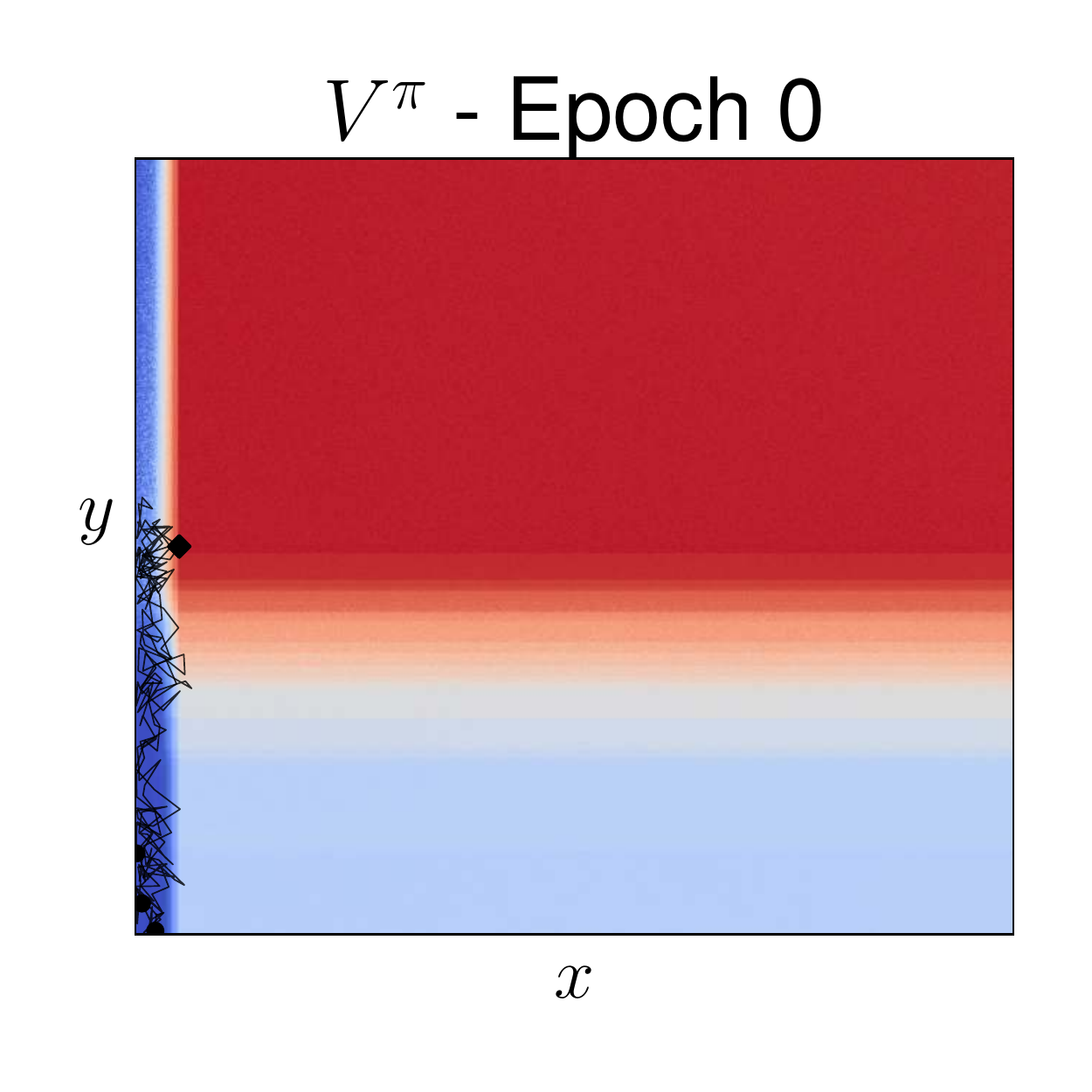} 
            \includegraphics[width=0.32\linewidth,valign=t, trim={1.5cm 1.8cm 0.8cm 1.85cm}, clip]{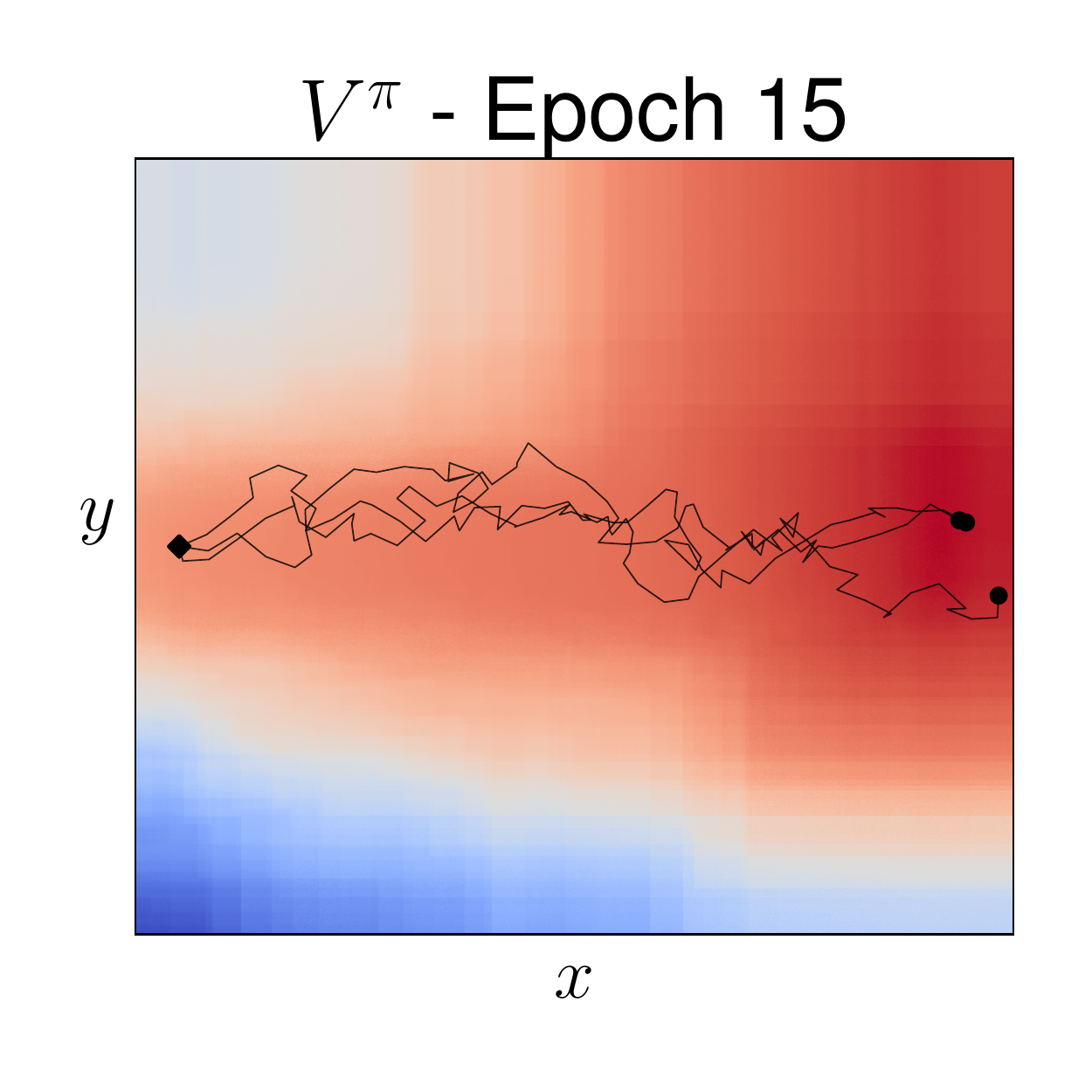}
            \includegraphics[width=0.32\linewidth,valign=t, trim={1.5cm 1.8cm 0.8cm 1.85cm}, clip]{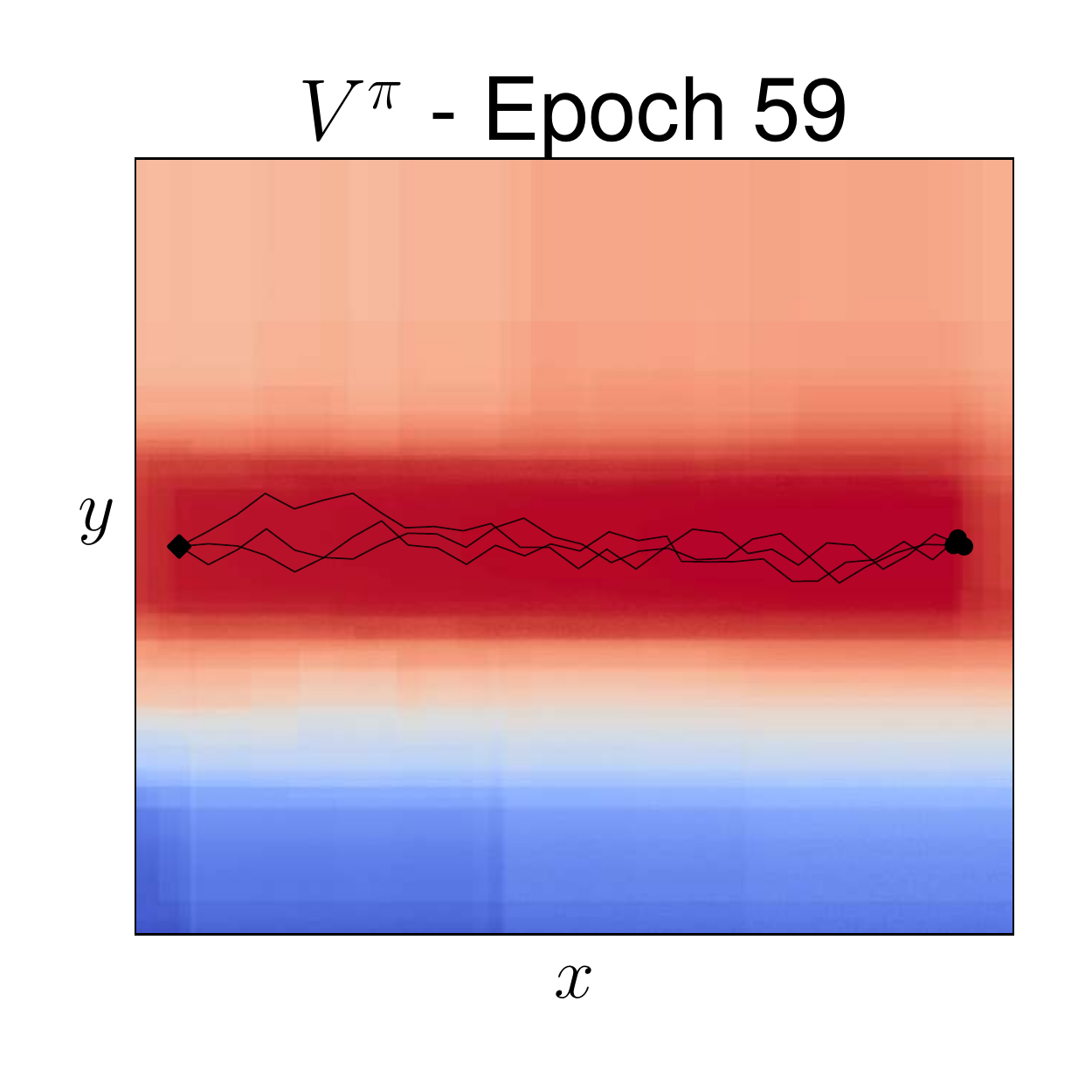}
        }
    \end{minipage}%
    \\
    \begin{minipage}[c]{0.45\textwidth}
        \subfloat[Corridor PPO]{%
            \includegraphics[width=0.32\linewidth,valign=t, trim={1.5cm 1.8cm 0.8cm 1.85cm}, clip]{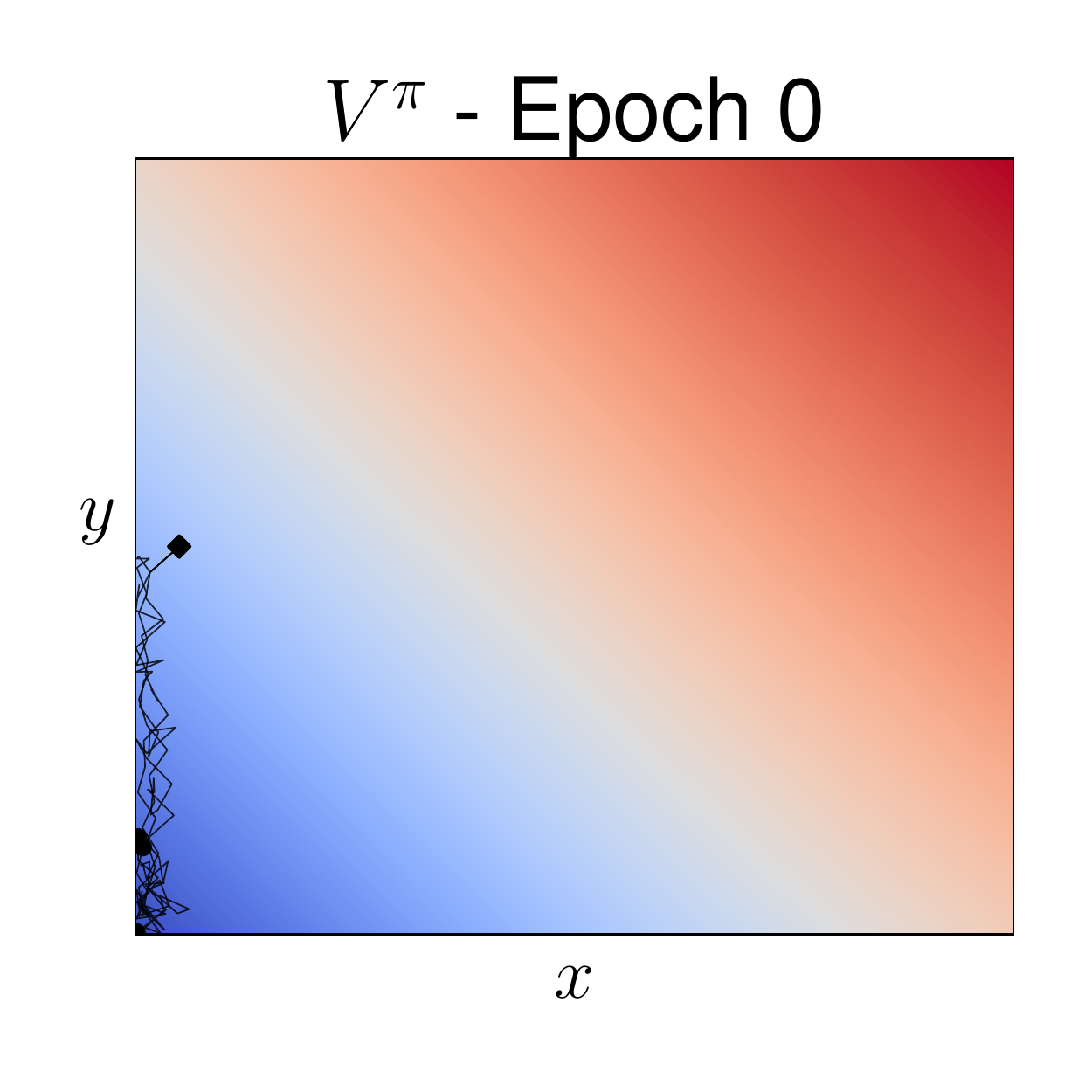} 
            \includegraphics[width=0.32\linewidth,valign=t, trim={1.5cm 1.8cm 0.8cm 1.85cm}, clip]{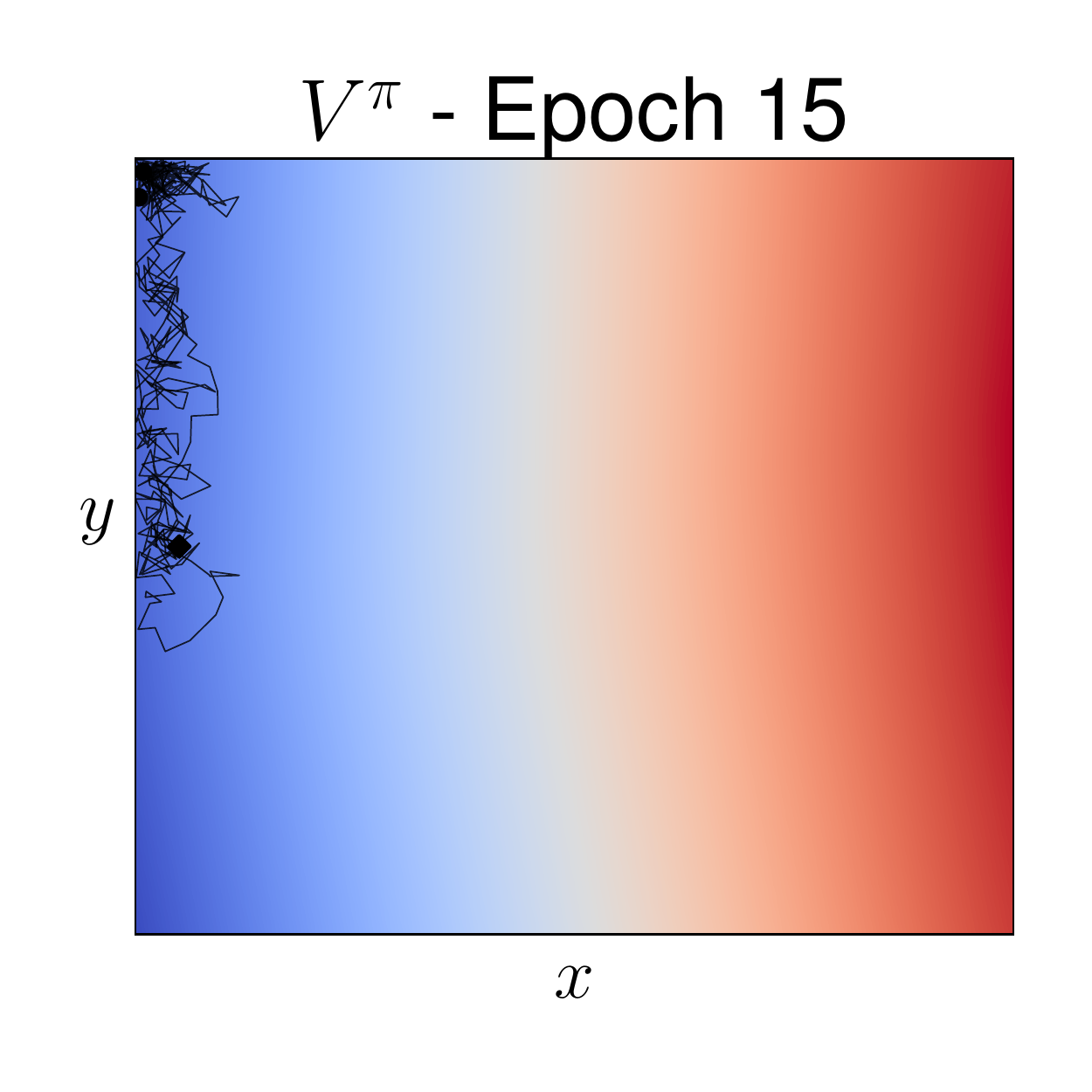}
            \includegraphics[width=0.32\linewidth,valign=t, trim={1.5cm 1.8cm 0.8cm 1.85cm}, clip]{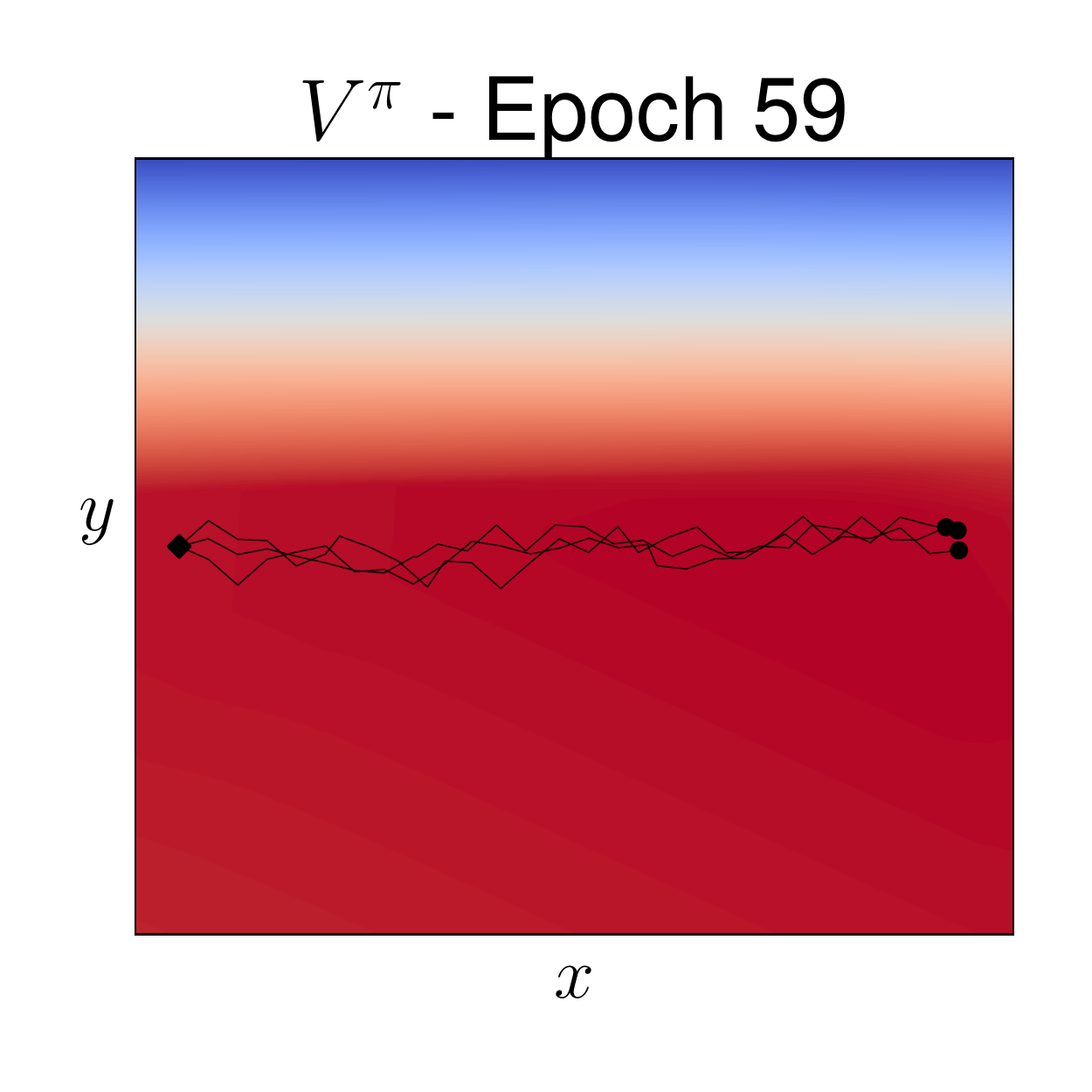}
        }
    \end{minipage}%
    \\
    \begin{minipage}[c]{0.45\textwidth}
        \subfloat[Room Tree-MVD]{%
            \includegraphics[width=0.32\linewidth,valign=t, trim={1.5cm 1.8cm 0.8cm 1.85cm}, clip]{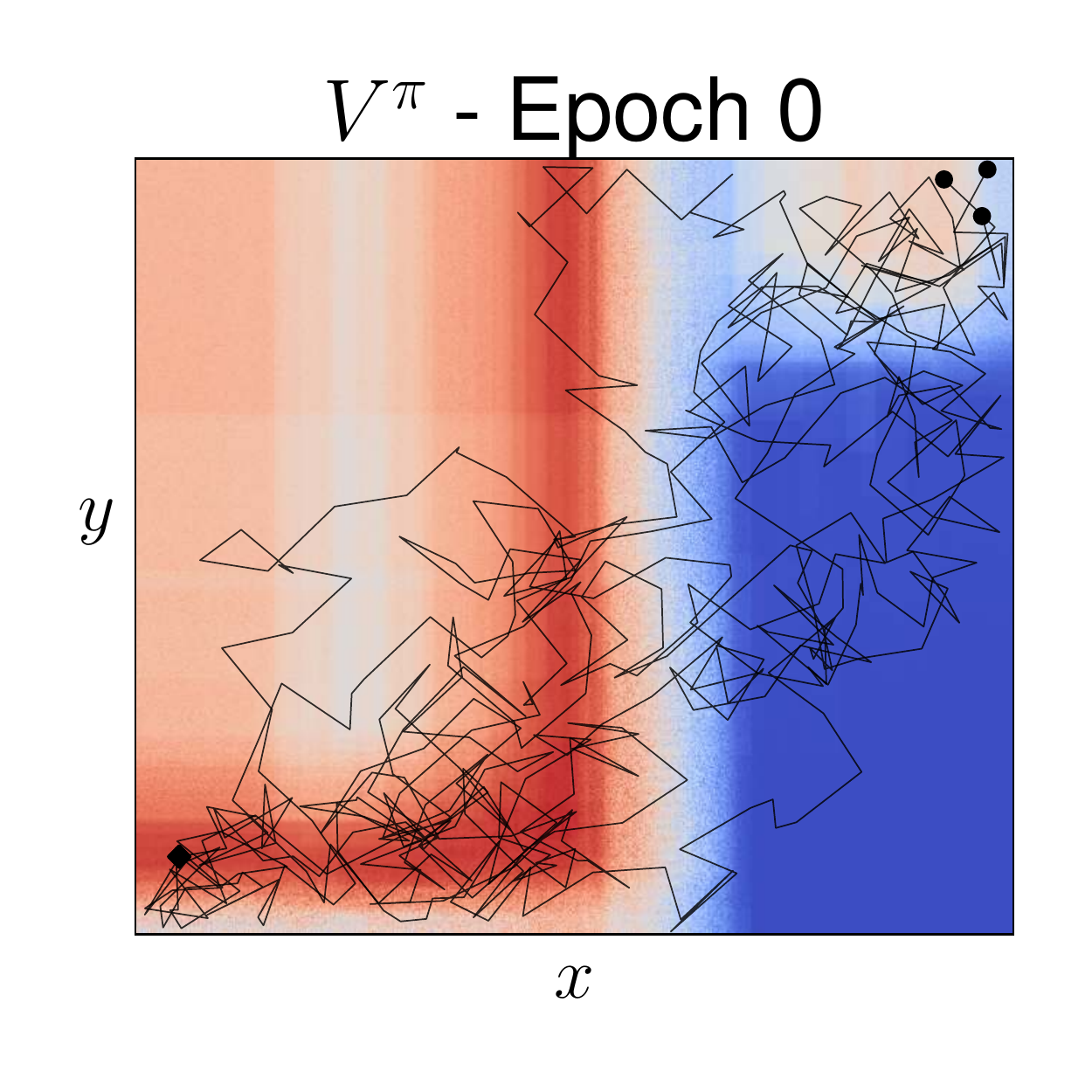} 
            \includegraphics[width=0.32\linewidth,valign=t, trim={1.5cm 1.8cm 0.8cm 1.85cm}, clip]{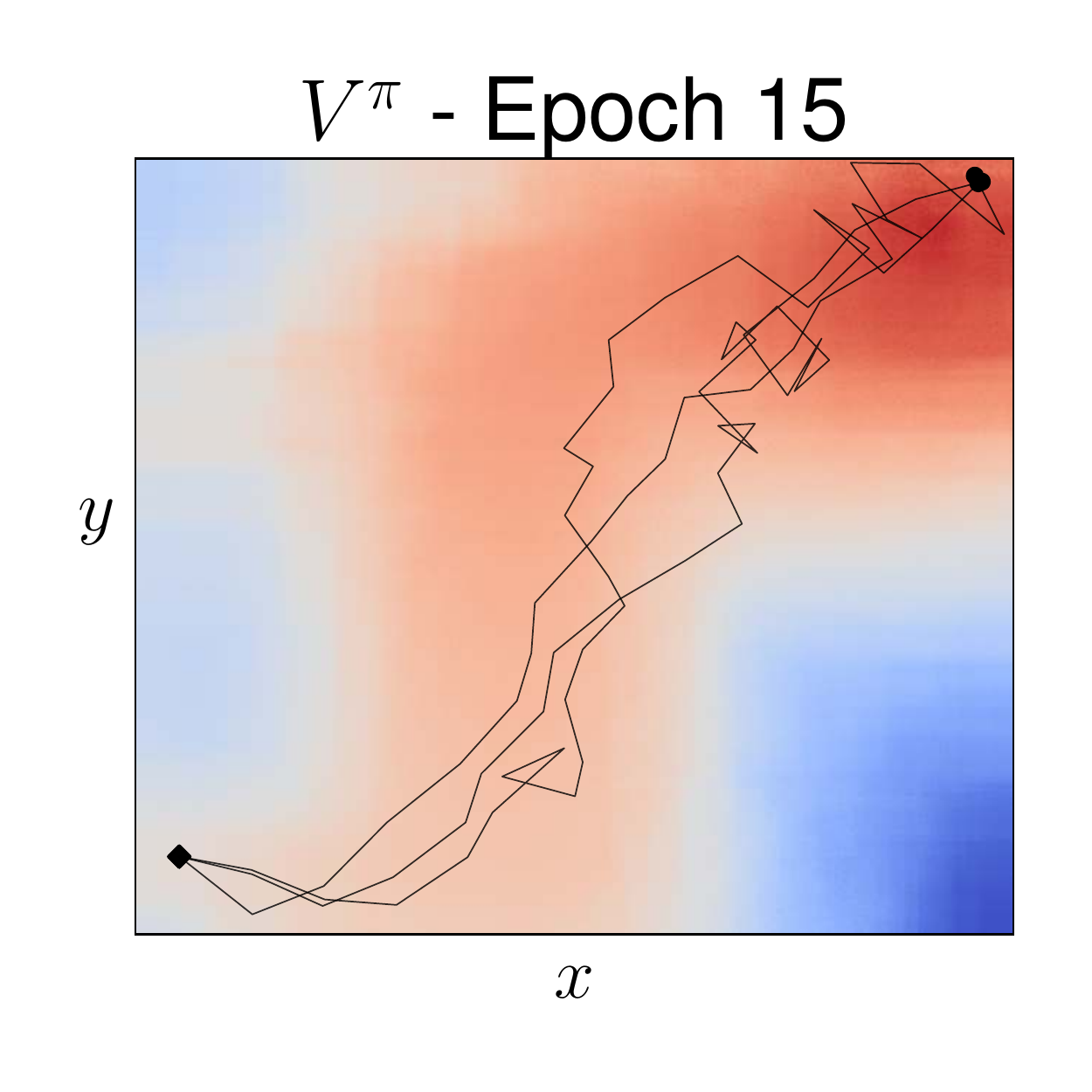}
            \includegraphics[width=0.32\linewidth,valign=t, trim={1.5cm 1.8cm 0.8cm 1.85cm}, clip]{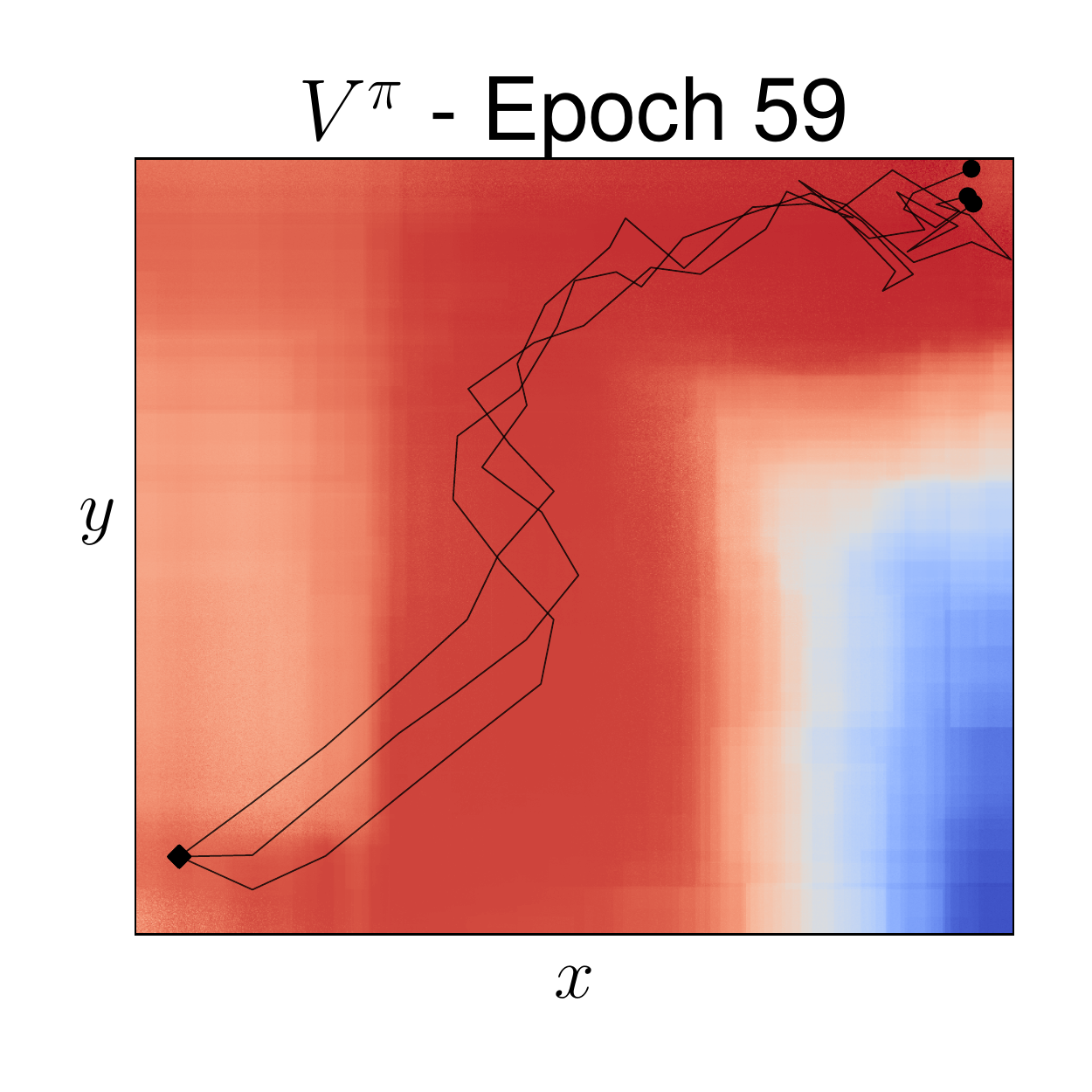}
        }
    \end{minipage}%
    \\
    \begin{minipage}[c]{0.45\textwidth}
        \subfloat[Room PPO]{%
            \includegraphics[width=0.32\linewidth,valign=t, trim={1.5cm 1.8cm 0.8cm 1.85cm}, clip]{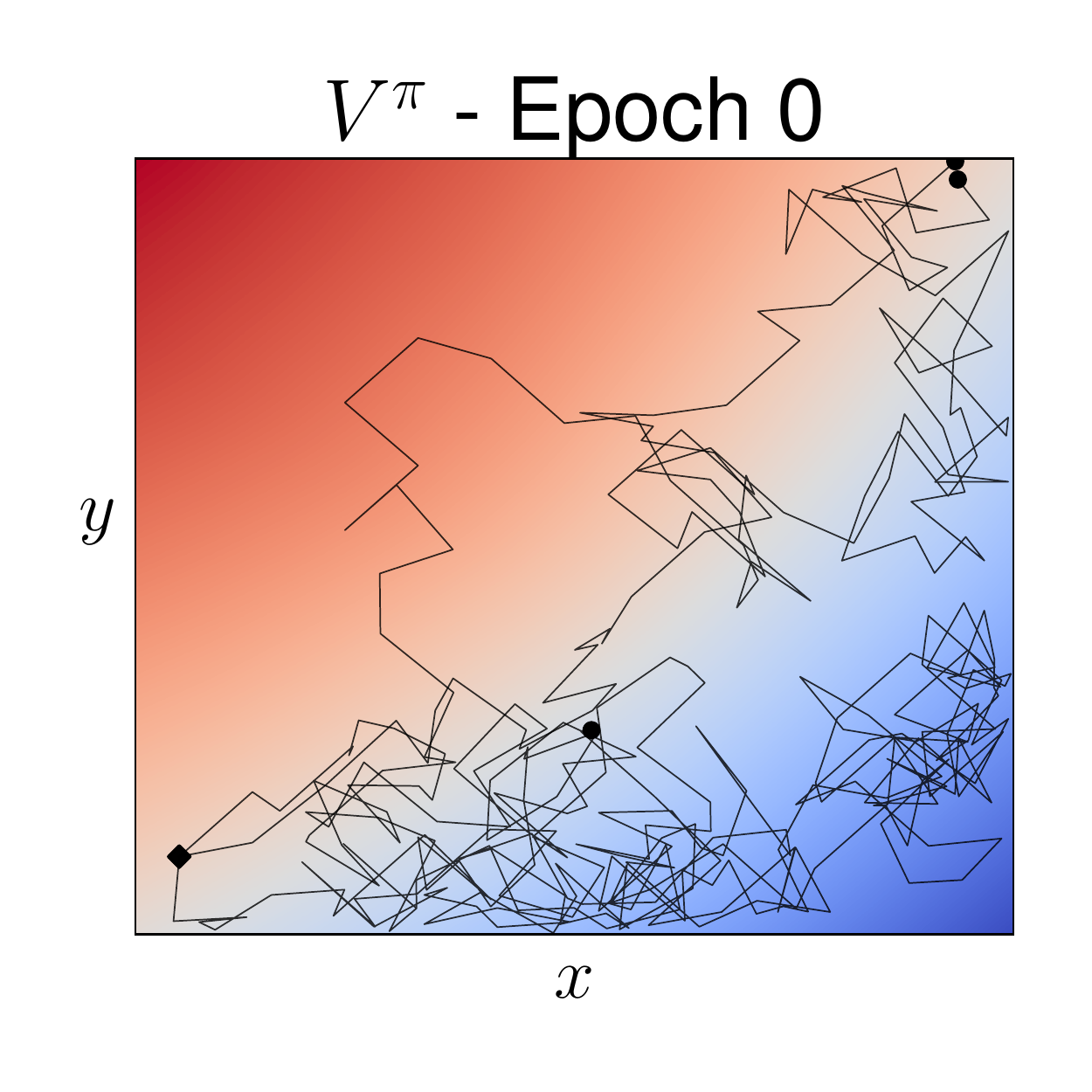} 
            \includegraphics[width=0.32\linewidth,valign=t, trim={1.5cm 1.8cm 0.8cm 1.85cm}, clip]{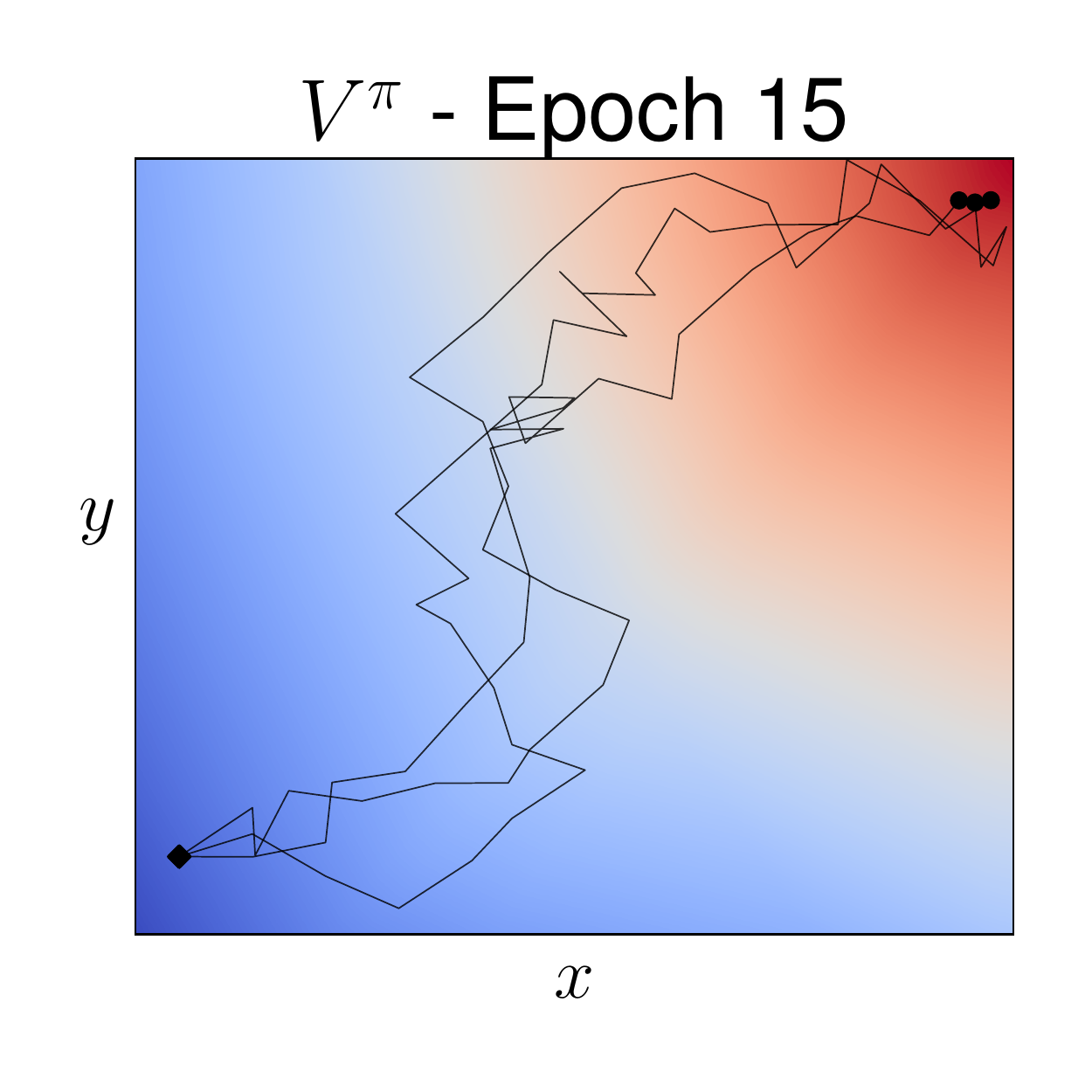}
            \includegraphics[width=0.32\linewidth,valign=t, trim={1.5cm 1.8cm 0.8cm 1.85cm}, clip]{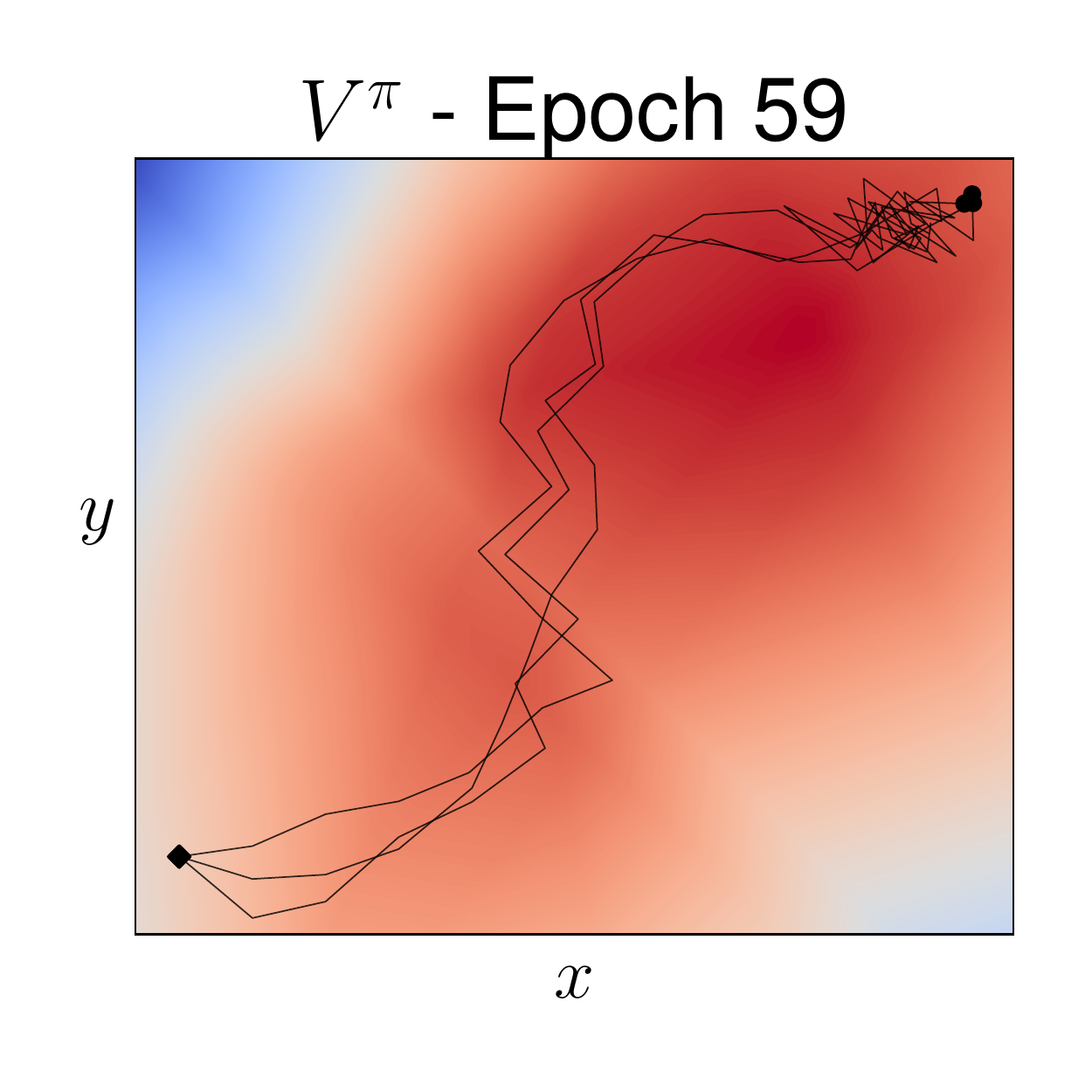}
        }
    \end{minipage}%
    \\
  \caption{Value functions for Tree-MVD and PPO at the end of different training epochs. In the Corridor and Room tasks the lines depict agent trajectories during evaluation, where the diamond is the starting state and the circle the ending state. Coloring: lower values in blue, higher values in red.}
  \label{fig:tree-value-functions} 
\end{figure}

\begin{figure}[ht] 
    \centering
    \includegraphics[width=0.48\textwidth,valign=t]{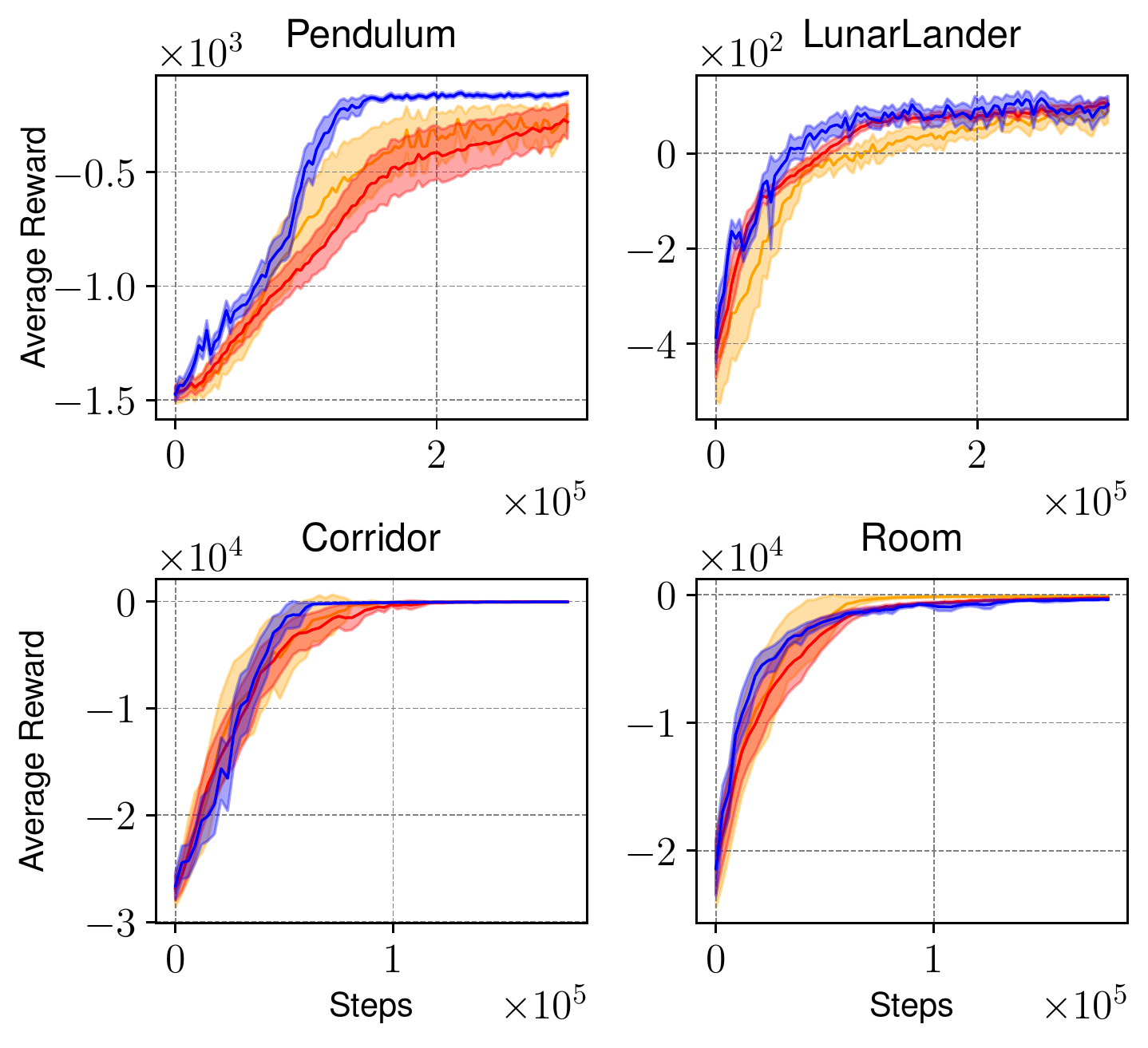}
    \\
    \includegraphics[width=0.3\textwidth]{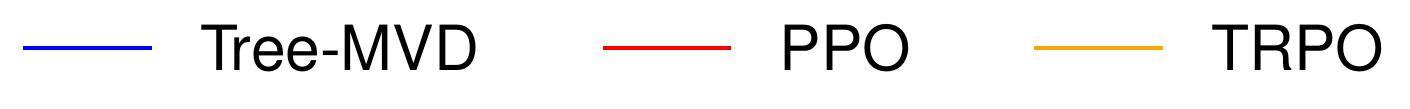}
    \caption{Policy evaluation results during training on different tasks with on-policy gradient algorithms. Depicted are the average reward per samples collected  and the $95\%$ confidence interval of 25 random seeds.}
    \label{fig:tree-experiments}
\end{figure}

As a baseline comparison we choose the on-policy algorithms \gls{ppo} and \gls{trpo}, which use a neural network to approximate the value function and \gls{gae} to estimate the advantage function.
For all tasks we use the same policy as in \gls{ppo}~\cite{schulman2017ppo} - a multivariate Gaussian distribution with diagonal covariance, where the mean is the output of a neural network and the log standard deviation a learnable state-independent parameter.
The best performing hyperparameters for both methods were chosen with a grid search.
Importantly, for \gls{treemvdpg} we use only one \gls{mc} sample to solve the action expectations in the policy gradient.

Fig.~\ref{fig:tree-experiments} shows the average reward during training. The plots suggest that the \gls{mvd}-based method performs comparable or better to the \gls{sf}-based \gls{ppo}.
Fig.~\ref{fig:tree-value-functions} presents the value functions learned by each algorithm at different training epochs in selected environments.
In the Pendulum task (1st row), the value function learned with \gls{treemvdpg} at epoch 15 is close to the optimal one at epoch 60.
This is in contrast to the one learned with \gls{ppo} for the same epoch (2nd row), which can explain the difference in performance in Fig.~\ref{fig:tree-experiments}.
In the Corridor and Room environments, we see the benefits of using regression trees instead of neural networks when the tasks exhibit a more structured representation.
Compare the value functions in the Corridor task for both algorithms (3rd and 4th rows).
Regression trees provide a better approximation of the optimal value function, are easier to interpret, and thus can partly explain the slight increase in performance over \gls{ppo} in Fig.~\ref{fig:tree-experiments}.
The value functions found by \gls{treemvdpg} better match the environments and reward descriptions from Fig.~\ref{fig:corridor-room}, and thus help accelerate learning.
A version of \gls{treemvdpg} with \gls{sf} showed unstable behaviors due to the absence of a value function estimator as baseline.

\section{Conclusion}
\label{sec:conclusion}

In this work we present \glspl{mvd} as a complement to the \gls{sf} and \gls{reptrick} estimators for actor-critic policy gradient algorithms.
We empirically show that methods based on this estimator are a viable alternative to the commonly used ones, and differently from~\cite{bhatt2019pgweak}, we avoid resetting the system to a specific state, which is impractical in real systems, showing how \glspl{mvd} are applicable to the general \gls{rl} framework.

Our experiments in step-based policy search highlight important facts about \glspl{mvd}.
In simple environments such as the \gls{lqr} and with an oracle $Q$-function, the \gls{mvd} performs better than the \gls{sf} and worse than the \gls{reptrick}.
However, in the presence of a local error, the \gls{mvd} and \gls{sf} estimates are not affected by the error frequency, but only by the amplitude.
On the contrary, the \gls{reptrick} is sensitive to both, which suggests that it should not be used if the $Q$-function approximator changes abruptly.
\gls{sf}-based methods need to estimate an advantage function for variance reduction, while \gls{mvd} only needs the $Q$-function approximation.
We can use non-differentiable critics, in particular regression trees, which in certain environments can lead to faster convergence and compete with \gls{sf}-based methods.
The \gls{reptrick} is not applicable to the latter case.
\glspl{mvd} can be used for high-dimensional action spaces in deep \gls{rl} algorithms, and obtain comparable results with the \gls{reptrick} using only one gradient estimate, which shows that the \gls{reptrick} is not a crucial part of the \gls{sac} algorithm. 

Which stochastic gradient estimator to use is problem dependent and an ongoing topic of research.
The \gls{reptrick} has shown in the recent years to achieve good results in \gls{vi}, e.g. in \glspl{vae}~\cite{kingma2014autoencoding}, and the \gls{sf} in policy gradient algorithms such as \gls{ppo}~\cite{schulman2017ppo}.
\glspl{mvd} are rarely used in \gls{ml}, however, they have been recently employed in approximate bayesian inference~\cite{rosca:2019:mvds}.

An interesting future research direction is to investigate how to reduce the computational complexity of \glspl{mvd} e.g., by computing only the derivatives along certain parameter dimensions and using a convex combination of all estimators. This kind of estimator is still unbiased~\cite{mohamed2019monte}.
Moreover, a theoretical analysis of bias and variance for \gls{mvd} is needed to improve our understanding of the estimators' properties and to determine in which kind of tasks one should be preferred over the others.

Given the empirical results presented in our work, we argue that the \gls{mvd} estimator is a useful tool for developing novel algorithms to solve challenging control problems in \glsdesc{rl} research.

\bibliography{references.bib}
\bibliographystyle{IEEEtran}

\newpage
\onecolumn

\clearpage

{\maketitle}

\section*{\centering \huge{Supplementary material}}
\setcounter{section}{0}

\section{Code to replicate the experiments}
\label{appendix:implementations}

The code is written in Python and makes use of PyTorch for automatic differentiation and the MushroomRL library for algorithm implementation and benchmarking \href{https://github.com/MushroomRL/mushroom-rl}{https://github.com/MushroomRL/mushroom-rl}. The code to reproduce the experiments is available at  \href{https://git.ias.informatik.tu-darmstadt.de/carvalho/mvd-rl}{https://git.ias.informatik.tu-darmstadt.de/carvalho/mvd-rl}.

\section{Optimization Test Functions}
\label{appendix:test-functions}

In Fig.~\ref{fig:grad-test-functions} we maximize $\E{p(\xvec; \distributionparams)}{f(\xvec)}$ via gradient ascent, where $f$ are three 2-dimensional test functions and $p(\xvec; \distributionparams)$ is a multivariate Gaussian distribution with diagonal covariance $p(\vec{x}; \distributionparams) = \Gaussian{\vec{x}; \distributionparams= \{ \muvec, \covariance \} }$, with $\xvec \in \R^2$ and $\distributionparams \in \R^4$.
The covariance is parameterized by the logarithm of standard deviation.
The learning rate is kept constant and equal to $5\times10^{-4}$ for all functions and estimators.
The \gls{mvd} uses one gradient estimate between parameter updates, while the \gls{sf} and \gls{reptrick} use the mean of eight estimates, which is the number of queries to $f$ done with \gls{mvd} for one estimate.
The \gls{sf} uses the optimal baseline for black-box optimization as computed by the \gls{pgpe} algorithm.

\medskip
\textbf{Quadratic function}

$f:\R^2 \to \R$

$f(\xvec)= -\xvec^\transpose \xvec $

Starting distribution: $\Gaussian{\xvec; \muvec=(-5, -5), \covariance=\diag{2^2, 2^2}}$

\medskip
\textbf{Himmelblau function}

$f:\R^2 \to \R$

$f(x,y)=-(x^{2}+y-11)^{2} - (x+y^{2}-7)^{2}$

Starting distribution: $\Gaussian{\xvec; \muvec=(0, -6), \covariance=\diag{2^2, 2^2}}$

\medskip
\textbf{Styblinski function}

$f:\R^2 \to \R$

$f(\xvec)= - \frac{1}{2} \sum_{i=1}^{2} \left( x_i^4 -16x_i^2 + 5x_i \right)$

Starting distribution: $\Gaussian{\xvec; \muvec=(0, 0), \covariance=\diag{2^2, 2^2}}$

\section{\texorpdfstring{\acrlong{lqr}}{}}
\label{appendix:lqr}

The discounted infinite-horizon discrete-time \gls{lqr} problem is defined as
\begin{align*}
\argmax_{\state_t, \action_t} J & = \sum_{t=0}^{\infty} -\gamma^t \left( \state_t^\transpose \mat{Q} \state_t + \action_t^{\transpose} \mat{R} \action_t \right) \\
\textrm{s.t. } & \state_{t+1} = \mat{A}\state_{t} + \mat{B}\action_{t}.
\end{align*}

The optimal control policy for this problem is a linear-in-the-state time-independent feedback controller $\action_t = - \mat{K}_{\mathrm{opt}}\state_t$. In the experiments we compute the gradient \wrt~$\mat{K}$ of a stochastic policy $\action \sim \mathcal{N}\left( \cdot |  -\mat{K}\state_t, \covariance \right) $, with fixed diagonal covariance $0.1^2$ in all action dimensions.

We build four environments with different dimensions of states and actions with ($|\statespace|, |\actionspace|$): $(2, 1)$, $(2, 2)$, $(4, 4)$ and $(6, 6)$. The matrices $\mat{A}$, $\mat{B}$, $\mat{Q}$ and $\mat{R}$ can be found in the accompanying code. The dynamics matrices $\mat{A}$ and $\mat{B}$ are non-diagonal, which makes the \gls{lqr} more difficult to solve.
The matrix $\mat{A}$ is chosen such that the system is unstable when $\action_t = 0$. The initial gain matrix $\mat{K}_{\mathrm{init}}$ is chosen by sampling from a Gaussian distribution with mean $\mat{K}_{\mathrm{opt}}$ and problem-dependent covariance, such that the closed-loop system is stable.
The initial state for each environment is $9$ for each dimension. For instance, in \glspl{lqr} $(2, 1)$ and $(2, 2)$ the initial state is $\state_0 = \left(9 ,9 \right)^\transpose$.
The discount factor is $\gamma=0.99$. To simulate infinite horizons, trajectory rollouts have $T=1000$ steps.
The policy gradients are computed from $\state_0$ and $\mat{K}_{\mathrm{init}}$. The discounted state distribution is obtained by sampling trajectories and multiplying each state at time $t$ with $\gamma^t$.

The experiments of Fig.~\ref{fig:lqr-errors} use the true $Q$ and $V$ functions computed with $\mat{K}_{\mathrm{init}}$.
Figures~\ref{fig:lqr-noise} and~\ref{fig:lqr-noise-training} simulate an error in the $Q$-function estimator with an added local sinusoidal term to the true state-action value function.
This approximator is modelled as $\hat{Q}(\state,\action) = Q(\state,\action) + \alpha Q(\state,\action)\cos(2\pi f \vec{p}^\transpose \vec{a} + \phi)$. $f$ is the error frequency, $\vec{p}$ is a random vector whose entries sum to 1, and $\phi\sim\mathcal{U}[0, 2\pi]$ a phase shift.
$\vec{p}$ and $\phi$ introduce randomness to remove correlation between action dimensions.  The factor $\alpha$ represents an error proportional to the true value, and the cosine term models adding a (high) frequency error component. These frequency components can appear in function approximators, especially if they overfit to the data. This is a simplified error model that uses only one frequency component, but it is useful to understand the sensitivity of the gradient estimators to local errors in function approximators.

\begin{figure*}[ht!]
	\noindent
	\centering
	\begin{minipage}[c]{0.45\textwidth}
		\subfloat[LQR 2 states 1 actions]{%
			\includegraphics[width=1.0\linewidth,valign=t]{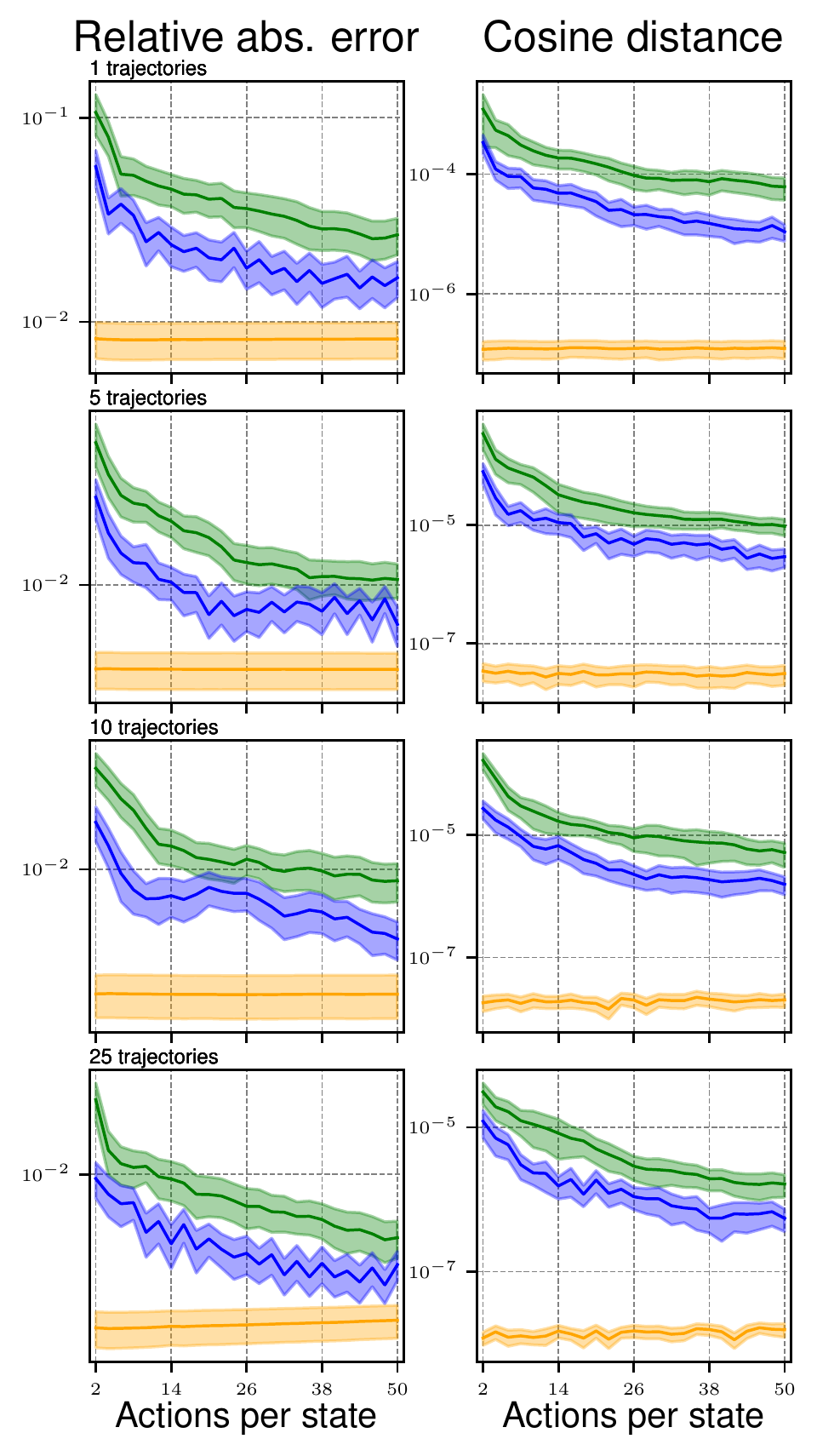} 
		}
	\end{minipage}%
	\hfill
	\begin{minipage}[c]{0.45\textwidth}
		\subfloat[LQR 2 states 2 actions]{%
			\includegraphics[width=1.0\linewidth,valign=t]{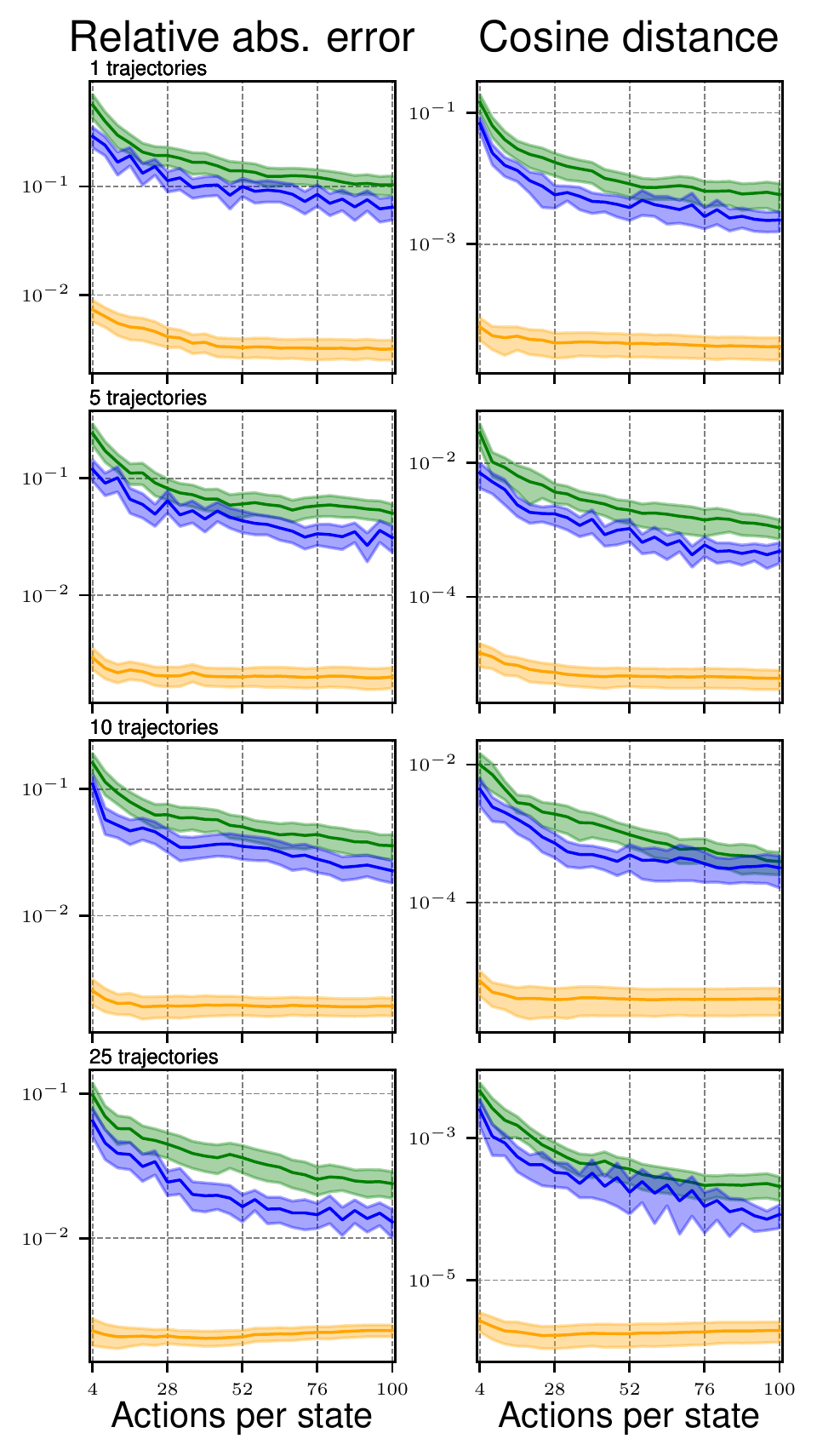} 
		}
	\end{minipage}%
	\\
	\includegraphics[width=0.32\linewidth,valign=t]{figures/legend_lqr.pdf} 
	\caption{Gradient errors in magnitude and direction in the \glspl{lqr} (2 states, 1 action) and (2 states, 2 actions), per number of trajectories and sampled actions. Depicted are the mean and the $95\%$ confidence interval of 25 random seeds.}
	\label{fig:lqr-appendix-1} 
\end{figure*}

\begin{figure*}[ht]
	\noindent
	\centering
	\begin{minipage}[c]{0.45\textwidth}
		\subfloat[LQR 4 states 4 actions]{%
			\includegraphics[width=1.0\linewidth,valign=t]{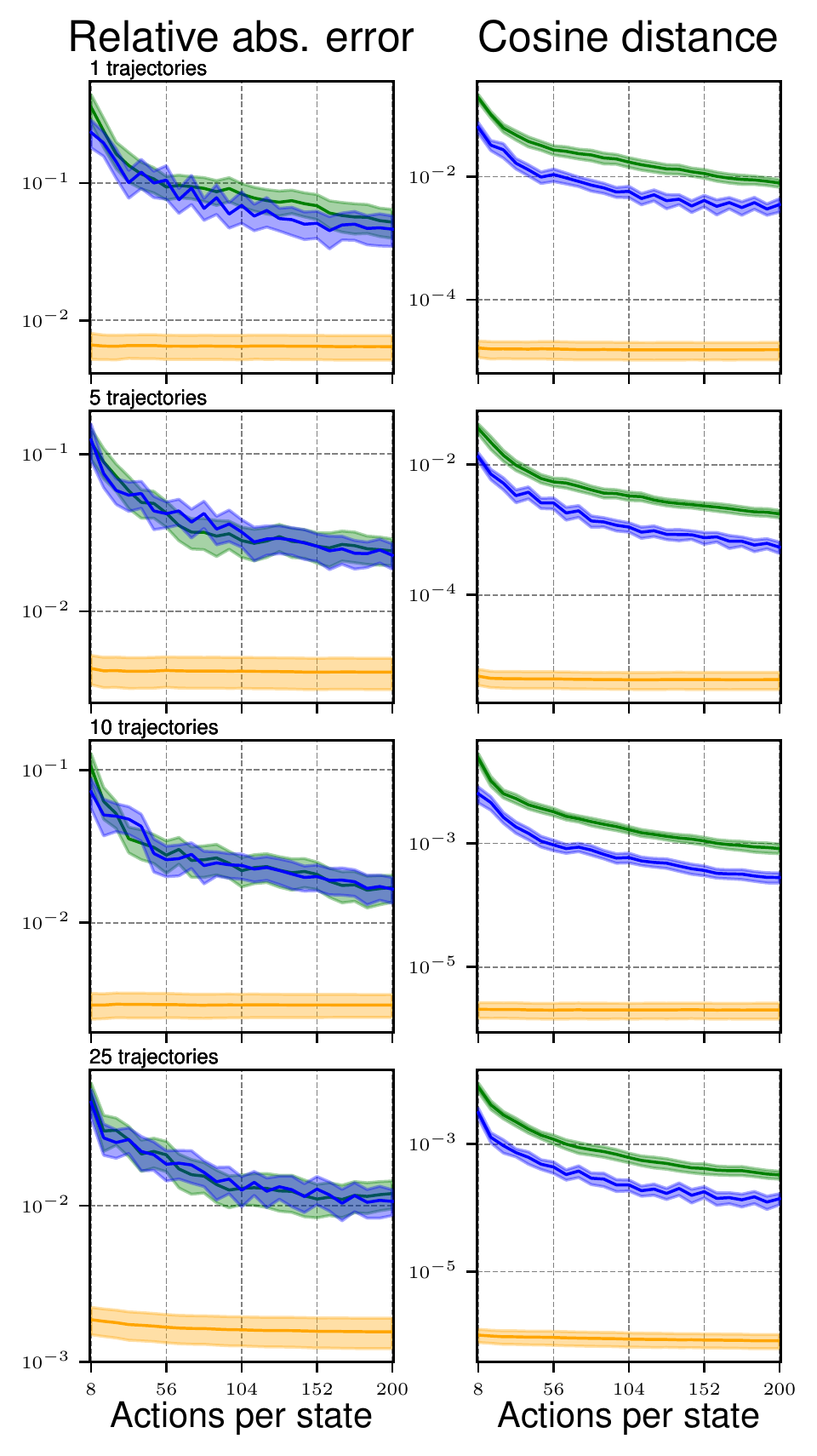} 
		}
	\end{minipage}%
	\hfill
	\begin{minipage}[c]{0.45\textwidth}
		\subfloat[LQR 6 states 6 actions]{%
			\includegraphics[width=1.0\linewidth,valign=t]{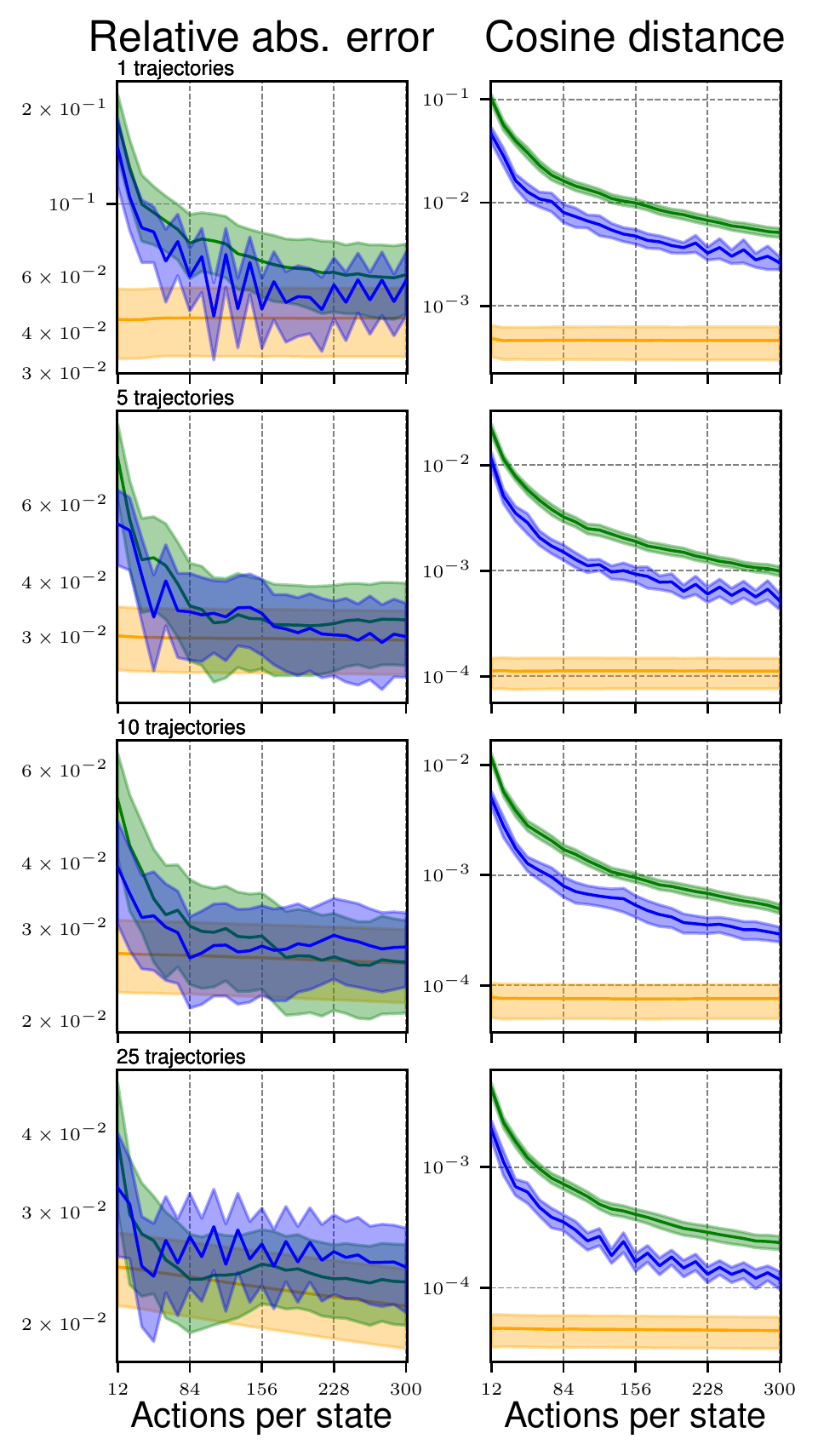} 
		}
	\end{minipage}%
	\\
	\includegraphics[width=0.4\linewidth,valign=t]{figures/legend_lqr.pdf} 
	\caption{Gradient errors in magnitude and direction in the \glspl{lqr}  (4 states, 4 actions) and (6 states, 6 actions), per number of trajectories and sampled actions. Depicted are the mean and the $95\%$ confidence interval of 25 random seeds.}
	\label{fig:lqr-appendix-2} 
\end{figure*}

Figures~\ref{fig:lqr-appendix-1} and~\ref{fig:lqr-appendix-2} complement the results of Fig.~\ref{fig:lqr-errors} with more trajectories, and a $6$-dimensional \gls{lqr}.

\begin{figure*}[ht!]
	\noindent
	\centering
	\begin{minipage}[c]{0.5\textwidth}
		\subfloat[LQR 2 states 1 actions]{%
			\includegraphics[width=1.0\linewidth,valign=t]{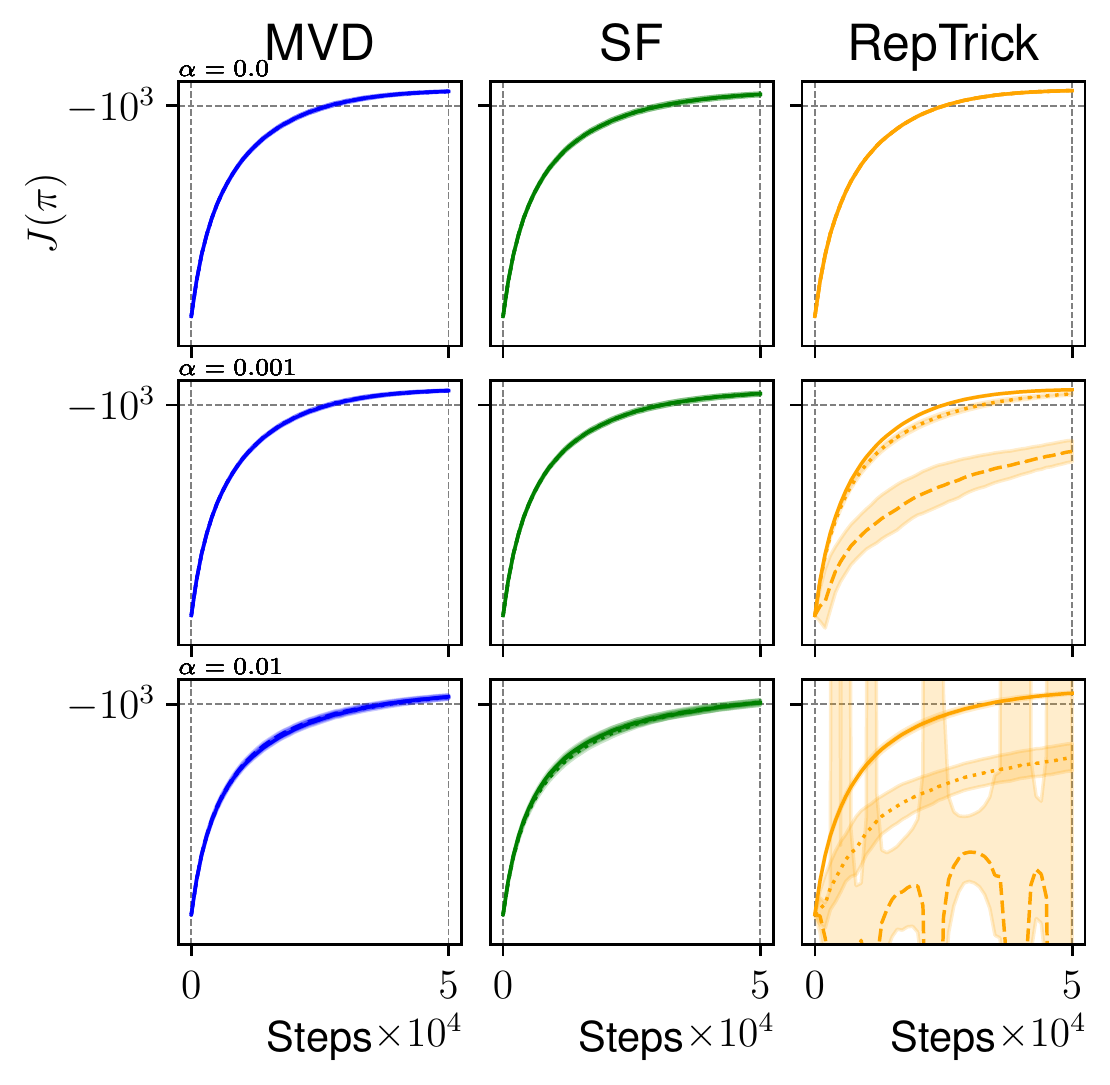} 
		}
	\end{minipage}%
	\hfill
	\begin{minipage}[c]{0.5\textwidth}
		\subfloat[LQR 2 states 2 actions]{%
			\includegraphics[width=1.0\linewidth,valign=t]{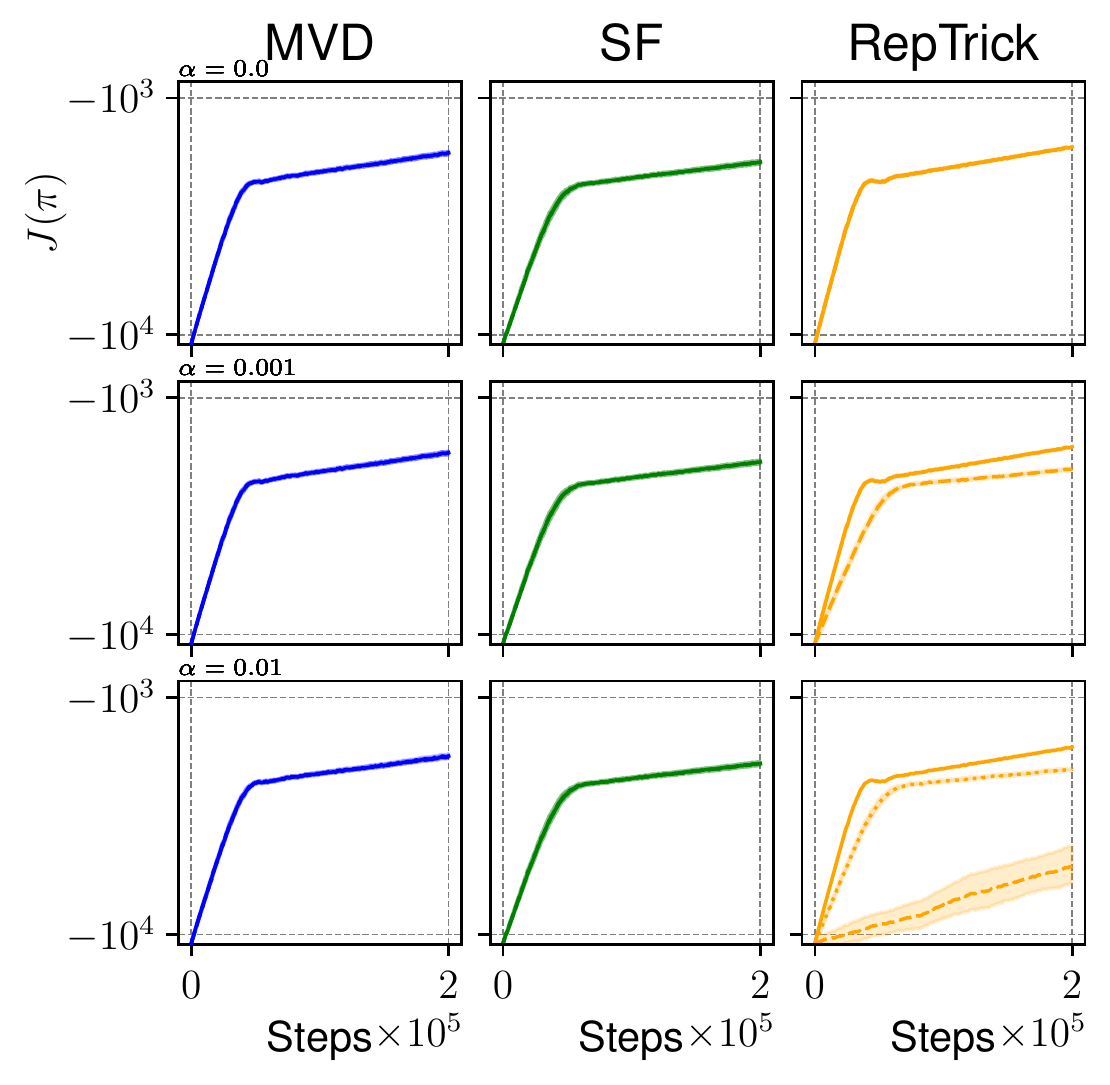} 
		}
	\end{minipage}%
	\\
	\hfill
	\begin{minipage}[c]{0.5\textwidth}
		\subfloat[LQR 4 states 4 actions]{%
			\includegraphics[width=1.0\linewidth,valign=t]{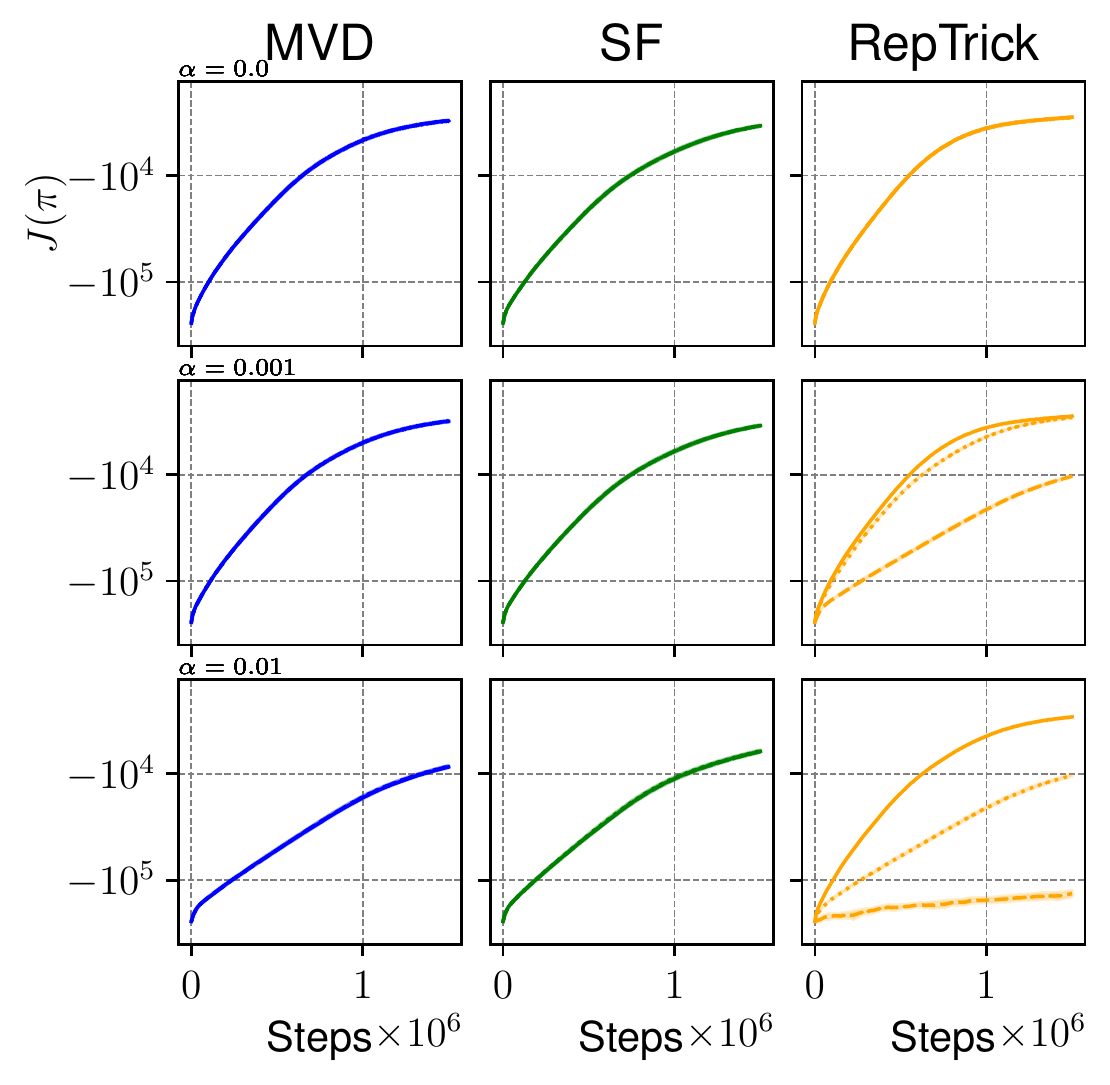} 
		}
	\end{minipage}%
	\hfill
	\begin{minipage}[c]{0.5\textwidth}
		\subfloat[LQR 6 states 6 actions]{%
			\includegraphics[width=1.0\linewidth,valign=t]{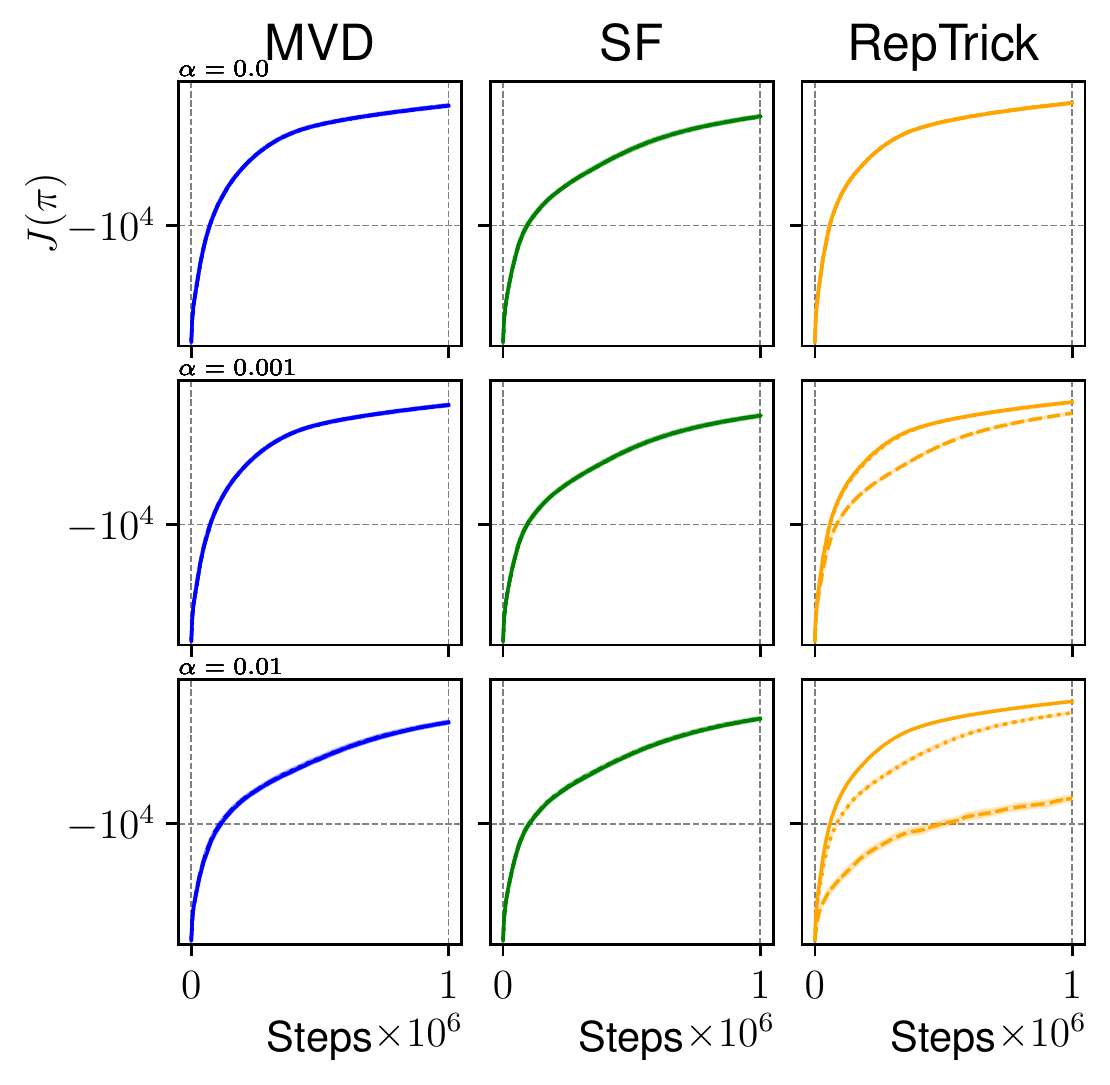} 
		}
	\end{minipage}%
	\\
	\includegraphics[width=1.0\linewidth,valign=t]{figures/legend_lqr_training.pdf} 
	\caption{Learning curves of the LQR tasks with an error in the $Q$-function approximator. Noise amplitudes (0.001, 0.01, 0.1, 1.0, 10.0), bottom to top.}
	\label{fig:lqr-noise-training-appendix} 
\end{figure*}

In Fig.~\ref{fig:lqr-noise-training} the initial policy is the same as in Fig.~\ref{fig:lqr-errors}. The policy update uses the Adam optimizer and the following learning rates  ($|\statespace|, |\actionspace|$): $(2, 1): 5\times10^{-2}$; $(2, 2): 1\times10^{-2}$; $(4, 4): 3\times10^{-3}$; $(6, 6): 5\times10^{-3}$.
Fig.~\ref{fig:lqr-noise-training-appendix} complements the results without added noise ($\alpha=0$), and a $6$-dimensional \gls{lqr}.

\clearpage
\section{Off-Policy Experiments}
\label{appendix:off-policy}

\begin{figure*}[ht] 
    \centering
    \includegraphics[width=1.0\textwidth,valign=t]{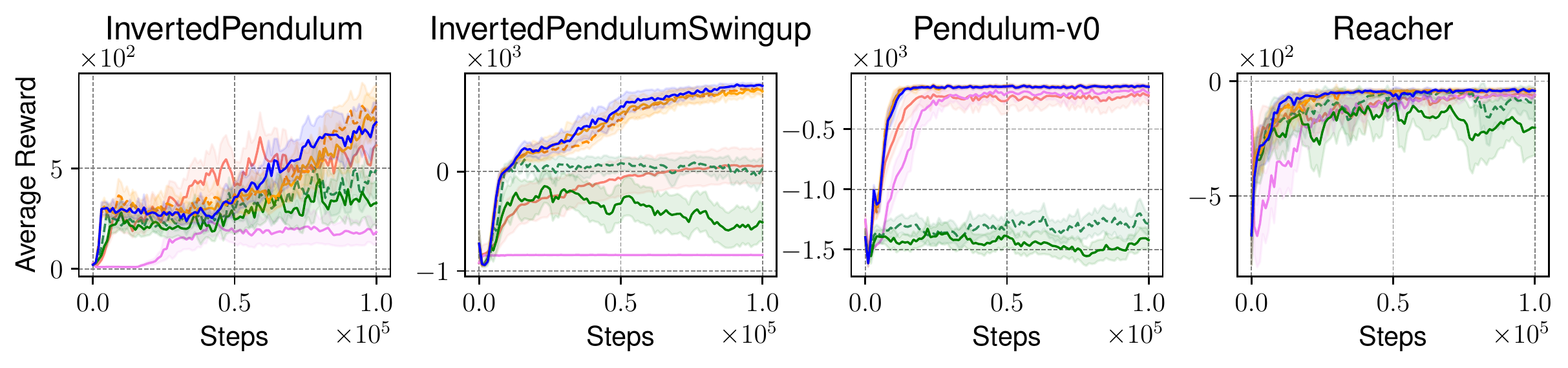}
    \\
  \includegraphics[width=1.0\textwidth]{figures/legend_stepbased.pdf}
  \caption{Policy evaluation results per samples collected during training on different tasks in deep \gls{rl}. Depicted are the average reward and the $95\%$ confidence interval of 25 random seeds.}
  \label{fig:step-based-experiments-low}
\end{figure*}

Off-policy experiments from Fig.~\ref{fig:step-based-experiments} use the environments from the PyBullet simulator~\cite{coumans2019bullet}. Fig.~\ref{fig:step-based-experiments-low} shows additional results in simpler tasks. Table~\ref{table:step-based-configs} contain the used hyperparameters. The neural network architectures are from the original papers.

\begin{table}[htb]
 \small
 \centering
 \begin{tabular}{ l l l l } 
     & Pendulum-v0 & InvertedPendulum-v0 & Ant-v0, HalfCheetah-v0 \\
     & & InvertedPendulumSwingup-v0  & Walker2d-v0, Hopper-v0 \\
     & & ReacherEnv-v0 & \\
     \hline \\
     \multicolumn{4}{l}{SAC variants} \\
     \cline{1-1} \\
     horizon & 200 & 1000 & 1000 \\
     $\gamma$ & 0.99 & 0.99 & 0.99 \\
     epochs & 50 & 50 & 100  \\
     steps/epoch & 1000 & 1000 & 10000 \\
     episodes evaluation & 10 & 10 & 10  \\
     batch size & 64 & 64 & 256  \\
     warmup transitions & 128 & 128 & 10000  \\
     max replay size & 500000 & 500000 & 500000 \\
     critic network & [64, 64] ReLU & [64, 64] ReLU & [256, 256] ReLU \\
     actor network & [64, 64] ReLU &  [64, 64] ReLU & [256, 256] ReLU \\
     optimizer & Adam & Adam & Adam \\
     lr actor & $1\times10^{-4}$ & $1\times10^{-4}$ & $1\times10^{-4}$  \\
     lr critic & $3\times10^{-4}$ & $3\times10^{-4}$ & $3\times10^{-4}$ \\
     \hline\\
     \multicolumn{4}{l}{DDPG and TD3} \\
     \cline{1-1} \\
     batch size & 64 & 64 & 256  \\
     warmup transitions & 128 & 128 & 10000  \\
     max replay size & 1000000 & 1000000 & 1000000 \\
     critic network & [64, 64] ReLU & [64, 64] ReLU & [400, 300] ReLU \\
     actor network & [64, 64] ReLU &  [64, 64] ReLU & [400, 300] ReLU \\
     optimizer & Adam & Adam & Adam \\
     lr actor & $1\times10^{-4}$ & $1\times10^{-4}$ & $1\times10^{-4}$  \\
     lr critic & $1\times10^{-3}$ & $1\times10^{-3}$ & $1\times10^{-3}$ \\
     \hline
 \end{tabular}
 \caption{Hyperparameters for the off-policy experiments.}
 \label{table:step-based-configs}
\end{table}

\clearpage
\section{On-Policy Experiments}
\label{appendix:on-policy}

For \gls{trpo} and \gls{ppo} the policy is a Gaussian distribution with diagonal covariance, where the mean is the output of a neural network, and the log-standard deviation is a state independent learnable parameter, as in the original papers. \gls{treemvdpg} optimizes the same policy but applies a $\tanh$ operator to the sampled actions, as done in \gls{sac}. Applying this operator to \glspl{mvd} is straight forward. 

Tables \ref{table:on-policy-configs-treemvd} and \ref{table:on-policy-configs-ppo} contain the hyperparameters used in the experiments. The neural network architectures are taken from the original works.

\begin{table}[htb]
 \small
 \centering
 \begin{tabular}{ l r r r r } 
     & Pendulum-v0 & LunarLanderContinuous-v2 & Room & Corridor \\
     \hline \\
     horizon & 200 & 1000 & 300 & 300 \\
     $\gamma$ & 0.99 & 0.99 & 0.99 \\
     epochs & 100 & 100 & 60 & 60  \\
     steps/epoch & 3000 & 3000 & 2000 & 2000 \\
     episodes evaluation & 10 & 10 & 10 & 10 \\
     iters bellman equation & 100 & 50 & 25 & 25 \\
     tree estimators & 100 & 50 & 25 & 25 \\
     min samples split node & 2 & 2 & 8 & 8 \\
     min samples leaf node & 1 & 1 & 4 & 4 \\
     max replay size & 500000 & 500000 & 500000 & 500000 \\
     replay batch size & 25000 & 25000 & 10000 & 10000 \\
     actor update epochs & 4 & 4 & 4 & 4 \\
     actor batch size & 256 & 128 & 128 & 128  \\
     actor network & [32, 32] ReLU &  [32, 32] ReLU & [32, 32] ReLU & [32, 32] ReLU \\     
     optimizer & Adam & Adam & Adam & Adam \\
     actor learning rate & $3\times10^{-4}$ & $3\times10^{-4}$ & $1\times10^{-4}$  & $1\times10^{-4}$ \\
     initial $\sigma$ & 1 & 1 & 1 & 1 \\
     \hline\\
 \end{tabular}
 \caption{Hyperparameters for the on-policy experiments with \gls{treemvdpg}.}
 \label{table:on-policy-configs-treemvd}
\end{table}

\begin{table}[htb]
 \small
 \centering
 \begin{tabular}{ l r r r r } 
     & Pendulum-v0 & LunarLanderContinuous-v2 & Room & Corridor \\
     \hline \\
     horizon & 200 & 1000 & 300 & 300 \\
     $\gamma$ & 0.99 & 0.99 & 0.99 \\
     epochs & 100 & 100 & 60 & 60  \\
     steps/epoch & 3000 & 3000 & 2000 & 2000 \\
     episodes evaluation & 10 & 10 & 10 & 10 \\
     critic update epochs & 10 & 10 & 10 & 10 \\
     critic batch size & 64 & 64 & 64 & 64  \\
     critic network & [32, 32] ReLU &  [32, 32] ReLU & [128, 128] ReLU & [128, 128] ReLU \\     
     critic learning rate & $3\times10^{-4}$ & $3\times10^{-4}$ & $3\times10^{-4}$  & $3\times10^{-4}$ \\
     actor update epochs & 8 & 4 & 4 & 4 \\
     actor batch size & 256 & 256 & 128 & 128  \\
     actor network & [32, 32] ReLU &  [32, 32] ReLU & [32, 32] ReLU & [32, 32] ReLU \\     
     optimizer & Adam & Adam & Adam & Adam \\
     actor learning rate & $3\times10^{-4}$ & $3\times10^{-4}$ & $1\times10^{-4}$  & $1\times10^{-4}$ \\
     initial $\sigma$ & 1 & 1 & 1 & 1 \\
     \hline\\
     PPO & \multicolumn{4}{l}{$\epsilon = 0.2$, $\lambda(\mathrm{GAE}) = 0.95$} \\
     TRPO & \multicolumn{4}{l}{$\max\mathrm{KL} = 0.01$, $\lambda(\mathrm{GAE}) = 0.95$, epochs line search $= 10$, epochs conj gradient $= 100$} \\
     \\
     \hline
 \end{tabular}
 \caption{Hyperparameters for the on-policy experiments with \gls{ppo} and \gls{trpo}.}
 \label{table:on-policy-configs-ppo}
\end{table}

\end{document}

%% file: definitions.tex
\usepackage{array}  %
\usepackage{graphicx}  %
\usepackage{amsmath, bm}  %
\usepackage{amsthm}   %
\usepackage{amsfonts} %
\usepackage{amssymb} %
\usepackage{mathtools}%
\usepackage{glossaries}

\usepackage[ruled,vlined,linesnumbered]{algorithm2e}

\usepackage{subfig}
\usepackage[export]{adjustbox}

\usepackage[inline, shortlabels]{enumitem}

\usepackage{mathtools}

\usepackage{wrapfig}

\usepackage{hyperref}

\usepackage{dblfloatfix}

\renewcommand{\vec}[1]{\bm{#1}}
\newcommand{\mat}[1]{\mathbf{#1}}

\DeclareMathOperator*{\argmax}{arg\,max}
	
\newcommand{\policy}{\pi}

\newcommand{\statespace}{\mathcal{S}}
\newcommand{\actionspace}{\mathcal{A}}

\newcommand{\state}{\vec{s}}
\newcommand{\nextstate}{\vec{s}'}
\newcommand{\action}{\vec{a}}
\newcommand{\omegavec}{\vec{\omega}}
\newcommand{\thetavec}{\vec{\theta}}
\newcommand{\phivec}{\vec{\phi}}

\newcommand{\xvec}{\vec{x}}
\newcommand{\epsvec}{\vec{\varepsilon}}

\newcommand{\dpartialomegak}{\frac{\partial}{\partial \omega_k}}

\newcommand{\dpartialmu}{\frac{\partial}{\partial \mu}}
\newcommand{\dpartialstd}{\frac{\partial}{\partial \sigma}}
\newcommand*\de{\mathop{}\!\mathrm{d}}

\newcommand{\wrt}{w.r.t.}

\DeclarePairedDelimiterX{\infdivx}[2]{(}{)}{%
	#1\;\delimsize\|\;#2%
}

\newcommand{\bigO}[1]{\mathcal{O}\!\left(#1\right)}

\newcommand{\R}{\mathbb{R}}
\newcommand{\E}[2]{\mathbb{E}_{#1}\left[#2\right]}

\newcommand{\V}[2]{\mathbb{V}_{#1}\left[#2\right]}
\newcommand{\cov}[2]{\mathrm{Cov}_{#1}\left[#2\right]}

\newcommand{\transpose}{\intercal}

\newcommand{\muvec}{\vec{\mu}}
\newcommand{\covariance}{\vec{\Sigma}}

\newcommand{\gradmuvec}{\nabla_{\muvec}}

\newcommand{\Gaussian}[1]{\mathcal{N}\left(#1\right)}

\newcommand{\Ptrans}{\mathcal{P}}

\newcommand{\diag}[1]{\textrm{diag}\left(#1\right)}

\newcommand{\policyparams}{\thetavec}

\newcommand{\gradpolicyparams}{\nabla_{\thetavec}}
\newcommand{\policyparametrized}{\policy_{\policyparams}}
\newcommand{\distributionparams}{\omegavec}
\newcommand{\distributionparamk}{\omega_k}
\newcommand{\graddistributionparams}{\nabla_{\omegavec}}
\newcommand{\graddistributionparamk}{\nabla_{\omega_k}}

%% file: acronyms.tex
\glsdisablehyper
\newacronym{ml}{ML}{Machine Learning}
\newacronym{rl}{RL}{Reinforcement Learning}
\newacronym{mdp}{MDP}{Markov Decision Process}

\newacronym{reptrick}{Rep-trick}{Reparametrization trick}
\newacronym{mvd}{MVD}{Measure-Valued Derivative}
\newacronym{sf}{SF}{Score-function}

\newacronym{vae}{VAE}{Variational Autoencoder}
\newacronym{vi}{VI}{Variational Inference}
\newacronym{elbo}{ELBO}{Evidence Lower Bound}

\newacronym{mc}{MC}{Monte Carlo}

\newacronym{fd}{FD}{Finite-difference}

\newacronym{lqr}{LQR}{Linear-Quadratic Regulator}

\newacronym{pgpe}{PGPE}{Policy Gradients with Parameter based Exploration}
\newacronym{nes}{NES}{Natural Evolutionary Strategies}
\newacronym{rwr}{RWR}{Reward Weighted Regression}
\newacronym{reps}{REPS}{Episodic Relative Entropy Policy Search}
\newacronym{nac}{NAC}{Natural Actor Critic}
\newacronym{power}{PoWER}{Policy Learning by Weighting Exploration with the Returns}
\newacronym{expected-sarsa}{Expected SARSA}{Expected SARSA}

\newacronym{trpo}{TRPO}{Trust Region Policy Optimization }
\newacronym{ppo}{PPO}{Proximal Policy Optimization}
\newacronym{ddpg}{DDPG}{Deep Deterministic Policy Gradient}
\newacronym{td3}{TD3}{Twin Delayed DDPG}
\newacronym{sac}{SAC}{Soft Actor-Critic}
\newacronym{gae}{GAE}{Generalized Advantage Estimation}

\newacronym{gpomdp}{}{GPOMPDP}
\newacronym{reinforce}{}{REINFORCE}

\newacronym{emvd}{E-MVD}{Episodic-MVD}
\newacronym{sacmvd}{SAC-MVD}{Soft Actor-Critic with MVD}
\newacronym{treemvdpg}{Tree-MVD}{Tree-MVD-Policy Gradient}

\newacronym{sacextrasamples}{SAC-extra-samples}{}

\newacronym{sacsf}{SAC-SF}{}
\newacronym{sacsfextrasamples}{SAC-SF-extra-samples}{}

\newacronym{spsa}{SPSA}{Simultaneous Perturbation Stochastic Approximation }

%% file: paper.bbl
\begin{thebibliography}{10}
\providecommand{\url}[1]{#1}
\csname url@samestyle\endcsname
\providecommand{\newblock}{\relax}
\providecommand{\bibinfo}[2]{#2}
\providecommand{\BIBentrySTDinterwordspacing}{\spaceskip=0pt\relax}
\providecommand{\BIBentryALTinterwordstretchfactor}{4}
\providecommand{\BIBentryALTinterwordspacing}{\spaceskip=\fontdimen2\font plus
\BIBentryALTinterwordstretchfactor\fontdimen3\font minus
  \fontdimen4\font\relax}
\providecommand{\BIBforeignlanguage}[2]{{%
\expandafter\ifx\csname l@#1\endcsname\relax
\typeout{** WARNING: IEEEtran.bst: No hyphenation pattern has been}%
\typeout{** loaded for the language `#1'. Using the pattern for}%
\typeout{** the default language instead.}%
\else
\language=\csname l@#1\endcsname
\fi
#2}}
\providecommand{\BIBdecl}{\relax}
\BIBdecl

\bibitem{kohl2004locomotion}
N.~{Kohl} and P.~{Stone}, ``Policy gradient reinforcement learning for fast
  quadrupedal locomotion,'' in \emph{IEEE International Conference on Robotics
  and Automation, 2004. Proceedings. ICRA '04. 2004}, vol.~3, 2004, pp.
  2619--2624 Vol.3.

\bibitem{deisenroth2013}
M.~P. Deisenroth, G.~Neumann, and J.~Peters, ``A survey on policy search for
  robotics.'' \emph{Found. Trends Robotics}, vol.~2, no. 1-2, pp. 1--142, 2013.

\bibitem{williamsReinforce1992}
R.~J. Williams, ``Simple statistical gradient-following algorithms for
  connectionist reinforcement learning,'' \emph{Mach. Learn.}, vol.~8, no.
  3–4, p. 229–256, 1992.

\bibitem{sutton1999PG}
R.~S. Sutton, D.~A. McAllester, S.~P. Singh, and Y.~Mansour, ``Policy gradient
  methods for reinforcement learning with function approximation,'' in
  \emph{Advances in Neural Information Processing Systems (NIPS)}, 1999, pp.
  1057--1063.

\bibitem{mohamed2019monte}
S.~Mohamed, M.~Rosca, M.~Figurnov, and A.~Mnih, ``Monte carlo gradient
  estimation in machine learning,'' \emph{Journal of Machine Learning
  Research}, vol.~21, no. 132, pp. 1--62, 2020.

\bibitem{bhatt2019pgweak}
S.~{Bhatt}, A.~{Koppel}, and V.~{Krishnamurthy}, ``Policy gradient using weak
  derivatives for reinforcement learning,'' in \emph{2019 IEEE 58th Conference
  on Decision and Control (CDC)}, 2019, pp. 5531--5537.

\bibitem{baxter2001GPOMDP}
J.~Baxter and P.~L. Bartlett, ``Infinite-horizon policy-gradient estimation,''
  \emph{J. Artif. Int. Res.}, vol.~15, no.~1, p. 319–350, Nov. 2001.

\bibitem{Peters:2006:PGRobotics}
J.~Peters and S.~Schaal, ``Policy gradient methods for robotics,'' in
  \emph{Proceedings of the IEEE/RSJ International Conference on Intelligent
  Robots and Systems (IROS)}, Beijing, China, 2006.

\bibitem{kingma2014autoencoding}
D.~P. Kingma and M.~Welling, ``{Auto-Encoding Variational Bayes},'' in
  \emph{2nd International Conference on Learning Representations, {ICLR} 2014,
  Banff, AB, Canada, April 14-16, 2014, Conference Track Proceedings}, 2014.

\bibitem{JanGuPoo17}
E.~Jang, S.~Gu, and B.~Poole, ``Categorical reparametrization with
  gumbel-softmax,'' in \emph{Proceedings International Conference on Learning
  Representations 2017}.\hskip 1em plus 0.5em minus 0.4em\relax
  OpenReviews.net, Apr. 2017.

\bibitem{Geurts2006ExtraTrees}
P.~Geurts, D.~Ernst, and L.~Wehenkel, ``Extremely randomized trees,''
  \emph{Mach. Learn.}, vol.~63, no.~1, p. 3–42, Apr. 2006.

\bibitem{glasserman2004monte}
P.~Glasserman, \emph{Monte Carlo methods in financial engineering}.\hskip 1em
  plus 0.5em minus 0.4em\relax New York: Springer, 2004.

\bibitem{Haarnoja2018SAC}
T.~Haarnoja, A.~Zhou, P.~Abbeel, and S.~Levine, ``Soft actor-critic: Off-policy
  maximum entropy deep reinforcement learning with a stochastic actor,'' in
  \emph{Proceedings of International Conference on Machine Learning ({ICML})},
  vol.~80, 2018, pp. 1856--1865.

\bibitem{krishnamurthy2011realtime}
V.~Krishnamurthy and F.~V. Abad, ``Real-time reinforcement learning of
  constrained markov decision processes with weak derivatives,'' 2011.

\bibitem{Schulman2015TRPO}
J.~Schulman, S.~Levine, P.~Moritz, M.~Jordan, and P.~Abbeel, ``Trust region
  policy optimization,'' in \emph{International Conference on International
  Conference on Machine Learning ({ICML})}, 2015, p. 1889–1897.

\bibitem{Lillicrap2016DDPG}
T.~P. Lillicrap, J.~J. Hunt, A.~Pritzel, N.~Heess, T.~Erez, Y.~Tassa,
  D.~Silver, and D.~Wierstra, ``Continuous control with deep reinforcement
  learning,'' in \emph{International Conference on Learning Representations,
  ({ICLR})}, 2016.

\bibitem{fujimoto2018td3}
S.~Fujimoto, H.~van Hoof, and D.~Meger, ``Addressing function approximation
  error in actor-critic methods,'' in \emph{Proceedings of the 35th
  International Conference on Machine Learning}, vol.~80, 2018, pp. 1587--1596.

\bibitem{Silver2014DPG}
D.~Silver, G.~Lever, N.~Heess, T.~Degris, D.~Wierstra, and M.~Riedmiller,
  ``Deterministic policy gradient algorithms,'' in \emph{Proceedings of the
  31st International Conference on International Conference on Machine Learning
  (ICML)}, 2014, p. 387–395.

\bibitem{krishnamurthy2003implementation}
V.~Krishnamurthy, K.~Martin, and F.~V. Abad, ``Implementation of gradient
  estimation to a constrained markov decision problem,'' in \emph{IEEE
  International Conference on Decision and Control}, vol.~5.\hskip 1em plus
  0.5em minus 0.4em\relax IEEE, 2003, pp. 4841--4846.

\bibitem{Degris2012OffPAC}
T.~Degris, M.~White, and R.~S. Sutton, ``Off-policy actor-critic,'' in
  \emph{Proceedings of the 29th International Coference on International
  Conference on Machine Learning}, ser. ICML'12.\hskip 1em plus 0.5em minus
  0.4em\relax Madison, WI, USA: Omnipress, 2012, p. 179–186.

\bibitem{Schulman2015GradEstimation}
J.~Schulman, N.~Heess, T.~Weber, and P.~Abbeel, ``Gradient estimation using
  stochastic computation graphs,'' in \emph{Proceedings of the International
  Conference on Neural Information Processing Systems (NIPS)}, 2015, p.
  3528–3536.

\bibitem{Glynn1987LikelihoodRatio}
P.~W. Glynn, ``Likelilood ratio gradient estimation: An overview,'' in
  \emph{Proceedings of the 19th Conference on Winter Simulation}, 1987, p.
  366–375.

\bibitem{PetersNAC2008}
J.~Peters and S.~Schaal, ``Natural actor-critic,'' \emph{Neurocomput.},
  vol.~71, no. 7–9, p. 1180–1190, Mar. 2008.

\bibitem{Pflug89}
G.~C. Pflug, ``Sampling derivatives of probabilities,'' \emph{Computing},
  vol.~42, no.~4, pp. 315--328, 1989.

\bibitem{heidergott2003mvds}
\BIBentryALTinterwordspacing
B.~Heidergott, G.~Pflug, and F.~J. Vazquez-Abad, ``Measure-valued
  differentiation for stochastic systems: from simple distributions to markov
  chains,'' 2003. [Online]. Available:
  \url{https://personal.vu.nl/b.f.heidergott/mvd.pdf}
\BIBentrySTDinterwordspacing

\bibitem{heidergott2000MVDfinitehorizon}
B.~Heidergott, F.~Vázquez-Abad, and M.~Gerad, ``Measure valued differentiation
  for stochastic processes: The finite horizon case,'' \emph{Technology
  Analysis and Strategic Management}, 2000.

\bibitem{rosca:2019:mvds}
M.~Rosca, M.~Figurnov, S.~Mohamed, and A.~Mnih, ``Measure-valued derivatives
  for approximate bayesian inference,'' in \emph{4th workshop on Bayesian Deep
  Learning}, 2019.

\bibitem{Pflug1996OptimizationStochasticModels}
G.~C. Pflug, \emph{Optimization of Stochastic Models}.\hskip 1em plus 0.5em
  minus 0.4em\relax Springer, 1996.

\bibitem{sehnke2010pgpe}
F.~Sehnke, C.~Osendorfer, T.~R{\"u}ckstiess, A.~Graves, J.~Peters, and
  J.~Schmidhuber, ``Parameter-exploring policy gradients,'' \emph{Neural
  Networks}, vol.~21, no.~4, pp. 551--559, May 2010.

\bibitem{Schulmanetal_ICLR2016_gae}
J.~Schulman, P.~Moritz, S.~Levine, M.~Jordan, and P.~Abbeel, ``High-dimensional
  continuous control using generalized advantage estimation,'' in
  \emph{Proceedings of the International Conference on Learning Representations
  (ICLR)}, 2016.

\bibitem{coumans2019bullet}
E.~Coumans and Y.~Bai, ``Pybullet, a python module for physics simulation for
  games, robotics and machine learning,'' \url{http://pybullet.org},
  2016--2019.

\bibitem{mnih2015humanlevel}
V.~Mnih, K.~Kavukcuoglu, D.~Silver, A.~A. Rusu, J.~Veness, M.~G. Bellemare,
  A.~Graves, M.~Riedmiller, A.~K. Fidjeland, G.~Ostrovski, S.~Petersen,
  C.~Beattie, A.~Sadik, I.~Antonoglou, H.~King, D.~Kumaran, D.~Wierstra,
  S.~Legg, and D.~Hassabis, ``Human-level control through deep reinforcement
  learning,'' \emph{Nature}, vol. 518, no. 7540, pp. 529--533, Feb. 2015.

\bibitem{Sutton2018IntroRL}
R.~S. Sutton and A.~G. Barto, \emph{Reinforcement Learning: An
  Introduction}.\hskip 1em plus 0.5em minus 0.4em\relax Cambridge, MA, USA: A
  Bradford Book, 2018.

\bibitem{brockman2016openai}
G.~Brockman, V.~Cheung, L.~Pettersson, J.~Schneider, J.~Schulman, J.~Tang, and
  W.~Zaremba, ``Openai gym,'' 2016, cite arxiv:1606.01540.

\bibitem{schulman2017ppo}
J.~Schulman, F.~Wolski, P.~Dhariwal, A.~Radford, and O.~Klimov, ``Proximal
  policy optimization algorithms.'' \emph{CoRR}, vol. abs/1707.06347, 2017.

\end{thebibliography}
